%% file: main.tex
\documentclass[review,onefignum,onetabnum]{siamonline220329}
\usepackage[top=1in,bottom=1in,left=1in,right=1in]{geometry}
\usepackage{amssymb}
\usepackage{amsmath}
\usepackage{amsbsy}
\usepackage{mathtools}
\usepackage{xcolor}
\usepackage{subfigure}
\usepackage{hyperref}
\usepackage{graphicx}
\usepackage{xcolor}
\usepackage{geometry}
\usepackage{caption}
\usepackage{subcaption}
\usepackage{float}
\usepackage{multirow}
\usepackage{adjustbox}
\usepackage{tikz}
\usepackage{relsize}
\usepackage{enumitem}

\newtheorem{rem}{Remark}[section]

\newcommand{\R}{\mathbb{R}}

\def\by {\boldsymbol{y}}

\def\bz {\boldsymbol{z}}

\def\R {\mathbb{R}}

\def\E {\mathbb{E}}

\def\Rn {{\R^n}}

\def\Cov {\mathrm{Cov}}

\newcommand{\Sxx}{P} 
\newcommand{\Sx}{\mathbf{q}} 
\newcommand{\Sc}{r} 

\def\weight{\theta}
\def\weightvec{{\boldsymbol{\weight}}}

\def\HJx{\mathbf{x}}
\def\HJp{\weightvec}
\def\HJt{t}

\def\MLbasismat{\Phi}

\def\MLy{\by}
\def\MLx{\bz}
\def\Svis{S_\epsilon}
\def\MLbasismatone{\Phi_1}
\def\noise{\xi}
\def\HJprior{g}
\def\Sigmainprior{\Lambda}

\definecolor{myCyan}{HTML}{08fcfc} 

\tikzstyle{box} = [rectangle, rounded corners, minimum width=0cm, minimum height=1cm,text centered, draw=black, fill=none]
\tikzstyle{smallbox} = [rectangle, rounded corners, minimum width=0cm, minimum height=0cm,text centered, draw=black, fill=none]
\tikzstyle{nobox} = [rectangle, rounded corners, minimum width=0cm, minimum height=1cm,text centered, draw=none, fill=none]
\tikzstyle{doublearrow} = [thick,<->,>=stealth]
\tikzstyle{dottedarrow} = [thick,<->,>=stealth,dotted]

\input{ex_shared}

\ifpdf
\hypersetup{
  pdftitle={Viscous HJ PDE for UQ in SciML},
  pdfauthor={P. Chen, T. Meng, Z. Zou, J. Darbon and G. E. Karniadakis}
}
\fi

\begin{document}

\nolinenumbers

\maketitle

\begin{abstract}
Uncertainty quantification (UQ) in scientific machine learning (SciML) combines the powerful predictive power of SciML with methods for quantifying the reliability of the learned models. However, two major challenges remain: limited interpretability and expensive training procedures. We provide a new interpretation for UQ problems  by establishing a new theoretical connection between some Bayesian inference problems arising in SciML and viscous Hamilton-Jacobi partial differential equations (HJ PDEs). Namely, we show that the posterior mean and covariance can be recovered from the spatial gradient and Hessian of the solution to a viscous HJ PDE. As a first exploration of this connection, we specialize to Bayesian inference problems with linear models, Gaussian likelihoods, and Gaussian priors. In this case, the associated viscous HJ PDEs can be solved using Riccati ODEs, and we develop a new Riccati-based methodology that provides computational advantages when continuously updating the model predictions. 
Specifically, our Riccati-based approach can efficiently add or remove data points to the training set invariant to the order of the data and continuously tune hyperparameters. Moreover, neither update requires retraining on or access to previously incorporated data. We provide several examples from SciML involving noisy data and \textit{epistemic uncertainty} to illustrate the potential advantages of our approach. In particular, this approach's amenability to data streaming applications demonstrates its potential for real-time inferences, which, in turn, allows for applications in which the predicted uncertainty is used to dynamically alter the learning process.
\end{abstract}

\begin{keywords}
Multi-time Hamilton-Jacobi PDEs, scientific machine learning, uncertainty quantification, Bayesian inference, Riccati equation
\end{keywords}

\begin{MSCcodes}
35F21, 62F15, 65L99, 65N99, 68T05, 35B37
\end{MSCcodes}

\input{sec_1}

\input{sec_2}

\input{sec_3}

\input{sec_4}

\input{sec_5}

\section*{Acknowledgement}
P.C. is funded by the Office of Naval Research (ONR) In-House Laboratory Independent Research Program (ILIR) at NAWCWD managed by Dr. Claresta Dennis and a Department of Defense (DoD) SMART Scholarship for Service SEED Grant. (Distribution Statement A. Approved for Public Release; distribution is unlimited. PR 24-0084.)
J.D., G.E.K., and Z.Z. are supported by the MURI/AFOSR FA9550-20-1-0358 project. We also
acknowledge the support of the DOE-MMICS SEA-CROGS DE-SC0023191 award.

\bibliographystyle{siamplain}
\bibliography{references}

\appendix
\input{sec_appendix}

\end{document}


\maketitle

\section{A detailed example}

Here we include some equations and theorem-like environments to show
how these are labeled in a supplement and can be referenced from the
main text.
Consider the following equation:
\begin{equation}
  \label{eq:suppa}
  a^2 + b^2 = c^2.
\end{equation}
You can also reference equations such as \cref{eq:matrices,eq:bb} 
from the main article in this supplement.

\lipsum[100-101]

\begin{theorem}
An example theorem.
\end{theorem}

\lipsum[102]
 
\begin{lemma}
An example lemma.
\end{lemma}

\lipsum[103-105]

Here is an example citation: \cite{KoMa14}.

\section[Proof of Thm]{Proof of \cref{thm:bigthm}}
\label{sec:proof}

\lipsum[106-112]

\section{Additional experimental results}
\Cref{tab:foo} shows additional
supporting evidence. 

\begin{table}[htbp]
\footnotesize
  \caption{Example table.}  \label{tab:smfoo}
\begin{center}
  \begin{tabular}{|c|c|c|} \hline
   Species & \bf Mean & \bf Std.~Dev. \\ \hline
    1 & 3.4 & 1.2 \\
    2 & 5.4 & 0.6 \\ \hline
  \end{tabular}
\end{center}
\end{table}

\bibliographystyle{siamplain}
\bibliography{references}

%% file: ex_shared.tex
\usepackage{lipsum}
\usepackage{amsfonts}
\usepackage{graphicx}
\usepackage{epstopdf}
\usepackage{algorithmic}
\usepackage{subfigure}

\ifpdf
  \DeclareGraphicsExtensions{.eps,.pdf,.png,.jpg}
\else
  \DeclareGraphicsExtensions{.eps}
\fi

\usepackage{enumitem}
\setlist[enumerate]{leftmargin=.5in}
\setlist[itemize]{leftmargin=.5in}

\newsiamremark{remark}{Remark}
\newsiamremark{hypothesis}{Hypothesis}
\crefname{hypothesis}{Hypothesis}{Hypotheses}
\newsiamthm{claim}{Claim}

\headers{Viscous HJ PDE for UQ in SciML}{Z. Zou, T. Meng, P. Chen, J. Darbon and G. E. Karniadakis}

\title{Leveraging viscous Hamilton-Jacobi PDEs for uncertainty quantification in scientific machine learning\thanks{Zongren Zou, Tingwei Meng, and Paula Chen contributed equally to this work.}}

\author{
Zongren Zou\thanks{Division of Applied Mathematics, Brown University, Providence, RI 02912 USA (\email{zongren\_zou@brown.edu}).}
\and
Tingwei Meng\thanks{Department of Mathematics, UCLA, Los Angeles, CA 90025 USA (\email{tingwei@math.ucla.edu}).}
\and
Paula Chen\thanks{Naval Air Warfare Center Weapons Division, China Lake, CA 93555 USA (\email{paula.x.chen.civ@us.navy.mil}).}
\and
J\'er\^ome Darbon\thanks{Corresponding author. Division of Applied Mathematics, Brown University, Providence, RI 02912 USA (\email{jerome\_darbon@brown.edu}).}
\and
George Em Karniadakis\thanks{Division of Applied Mathematics and School of Engineering, Brown University, Providence, RI 02912 USA, and Pacific Northwest National Laboratory, Richland, WA 99354 USA (\email{george\_karniadakis@brown.edu}).}
}

\usepackage{amsopn}


%% file: sec_1.tex
\section{Introduction}

\begin{figure}[htbp]
\centering
\begin{adjustbox}{width=\textwidth}
\begin{tikzpicture}[node distance=2cm]
\node (model) [nobox] {Bayesian model,};
\node (weight) [nobox, right of=model, xshift=1.2cm] {model parameters $\sim$};
\node (posterior) [box, right of=weight, draw=red!60, fill=red!5, xshift=0.75cm] {posterior};
\node (propto) [nobox, right of=posterior, xshift=-0.85cm] {$\propto$};
\node (prior) [box, right of=propto, draw=blue!60, fill=blue!5, xshift=-1.15cm] {prior};
\node (times) [nobox, right of=prior, xshift=-1.2cm] {$\times$};
\node (likelihood) [box, right of=times, draw=green!60, fill=green!5, xshift=-0.75cm] {likelihood};
\node (noise) [nobox, right of=likelihood, xshift=-0.3cm] {, noise $\sim$ };
\node (var) [box, right of=noise, draw=orange!60, fill=orange!5, xshift=-0.18cm] {Gaussian};
\node (index) [nobox, right of=var, xshift=0.25cm] {, $N$ data points};

\node (empty) [nobox, below of=model] { };

\node (S) [box, below of=empty, draw=red!60, fill=red!5, text width=3cm] {solution $\Svis$ to multi-time viscous HJ PDE};
\node (int) [nobox, right of=S, xshift=0.528cm] {$=\epsilon\log\mathlarger{\int}$};
\node (IC) [box, right of=int, draw=blue!60, fill=blue!5, xshift=0.55cm, text width=3cm] {term in the initial condition};
\node(times2) [nobox, right of=IC, xshift=-0.1cm] {$\times$};
\node (H) [box, right of=IC, draw=green!60, fill=green!5, xshift=4.19cm] {$\exp \left(-\sum_{i=1}^N\quad \,\,\,\,\,\quad\quad\text{quadratic Hamiltonian}\right)$};
\node (HJt) [smallbox, right of=IC, draw=orange!60, fill=orange!5, xshift=3.05cm] {$\frac{\text{time}}{\epsilon}$};
\node (times3) [nobox, right of=HJt, xshift=-1.27cm] {$\times$};

\draw [doublearrow, draw=red!60] (posterior) -- (S) node[midway, left, xshift=1cm, yshift=0.1cm, text width=8cm] {$\textrm{spatial gradient of }\Svis=\textrm{posterior mean}$, $\textrm{Hessian of }\Svis=\frac{1}{\epsilon}(\textrm{posterior covariance})$};
\draw [doublearrow, draw=blue!60] (prior) -- (IC) node[midway, right, xshift=-0.0cm] {$\propto$};
\draw [doublearrow, draw=green!60] (likelihood) -- (H)  node[midway, right, xshift=-0.07cm] {$\propto$};
\draw [doublearrow, draw=orange!60] (var) -- (HJt) node[midway, right, xshift=0.17cm, yshift=0.1cm] {$\text{variance}=\frac{\epsilon}{\text{time}}$};
\end{tikzpicture}
\end{adjustbox}

\caption{(See Section~\ref{sec:general_connection}) Illustration of a connection between a Bayesian inference problem in scientific machine learning (\textbf{top}) and the solution to a multi-time viscous HJ PDE (\textbf{bottom}). The colors indicate the associated quantities between problems. This color scheme is reused in the subsequent illustrations of our connection. The arrow labels describe how the boxed quantities are related. For example, the posterior mean in the learning problem is equivalent to the spatial gradient of the solution to the multi-time viscous HJ PDE (\textcolor{red}{\textbf{red}}). }
\label{fig:intro_connection_in_words}
\end{figure}
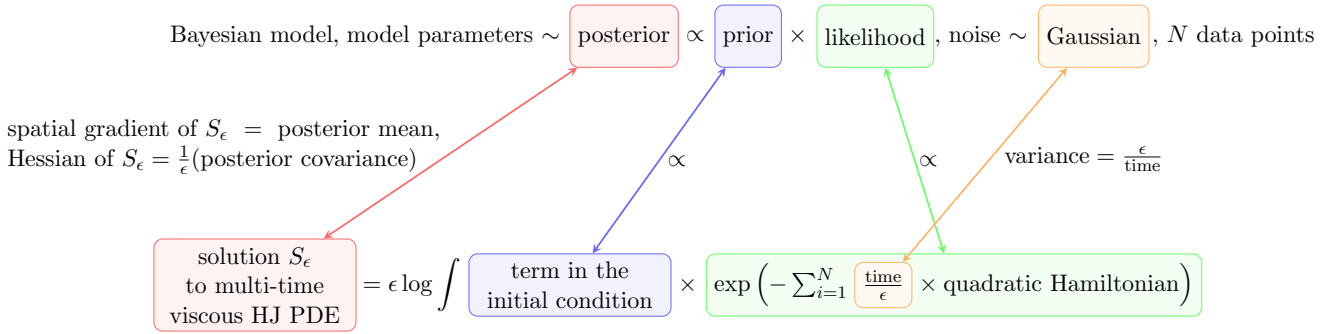

Scientific machine learning (SciML) is a recent and evolving field consisting of data-driven techniques for learning physics-based models \cite{raissi2019physics, karniadakis2021physics, williams2006gaussian, cuomo2022scientific, pang2019fpinns, raissi2017machine, raissi2018numerical, zou2023hydra, chen2021solving, zou2023correcting, yang2021b, meng2022learning, zou2023uncertainty} that finds applications in diverse areas \cite{mao2020physics, cai2021heat, cai2021fluid, sahli2020physics, misyris2020physics, linka2022bayesian, yin2023generative, zhang2024discovering}. 
While deterministic SciML models demonstrate impressive predictive power, real-life applications often require quantitative metrics for the trustworthiness of the learned models. 
To this end, uncertainty quantification (UQ) \cite{hullermeier2021aleatoric, psaros2023uncertainty, zou2022neuraluq} has been integrated with SciML to create new learning approaches that can  produce quantifiably high-confidence models in the absence of a ground truth and use these confidence metrics to help inform the learning process.

In this paper, we reinterpret certain Bayesian inference problems arising from SciML and UQ using the framework of viscous Hamilton-Jacobi partial differential equations (HJ PDEs)~\cite{evans2010pdes, yong2012stochastic, fabbri2017stochastic}. 
Specifically, we develop a new theoretical connection between these Bayesian inference problems and multi-time viscous HJ PDEs that shows that the posterior mean and covariance of the Bayesian model can be recovered from the spatial gradient and Hessian of the solution to an associated viscous HJ PDE (Section~\ref{sec:general_connection}). 
This connection is summarized in Figure~\ref{fig:intro_connection_in_words}. 
Our previous work in~\cite{chen2023leveraging, chen2023leveraging2} establishes similar theoretical connections between deterministic SciML problems and inviscid HJ PDEs. As such, the methodology in~\cite{chen2023leveraging, chen2023leveraging2} does not automatically extend to UQ. Meanwhile, the work in~\cite{darbon2021bayesian} establishes a similar link between Bayesian inference and viscous HJ PDEs but develops this connection via the Cole-Hopf transformation~\cite{evans2010pdes} and only considers the single-time case. In contrast, we do not rely on the Cole-Hopf transformation and consider multi-time HJ PDEs to enable our connection to SciML.

As a first exploration of this connection, we specialize to Bayesian inference problems with linear models, Gaussian likelihoods, and Gaussian priors (Section~\ref{sec:riccatimethod}). 
In this case, the corresponding viscous HJ PDE (and hence, the Bayesian inference problem) can be solved using Riccati ODEs~\cite{mceneaney2006max}. The resulting Riccati-based methodology can then be leveraged to efficiently update the model predictions (Section~\ref{sec:3_2}) and continuously tune hyperparameters (Section~\ref{sec:3_3}). 
To illustrate the potential advantages of this approach, we apply this Riccati-based methodology to several UQ-based examples from SciML (Section~\ref{sec:examples}). In particular, we focus on noisy data cases and quantify \textit{epistemic uncertainty} \cite{hullermeier2021aleatoric, psaros2023uncertainty, zou2022neuraluq}, which refers to the uncertainty associated with the model parameters.
In Section~\ref{sec:4_2}, we demonstrate the potential computational advantages of our approach over more standard SciML/UQ techniques when incrementally updating the model predictions. Namely, we show how our Riccati-based approach naturally coincides with continual learning 
while inherently avoiding catastrophic forgetting despite updating the model predictions without accessing the historical data (Section~\ref{sec:example_1a}). We also show how solving the Riccati ODEs provides a continuous flow of solutions with respect to the hyperparameters of the model, which allows the hyperparameters to be tuned continuously (Section~\ref{sec:example_1b}). In Section~\ref{sec:example2}, we demonstrate the versatility of our Riccati-based approach in handling different learning scenarios, e.g., big data (Section~\ref{sec:4_2_1}) and active learning (Section~\ref{sec:4_2_2}) settings. In particular, these examples illustrate how uncertainty metrics can be leveraged to inform how we learn. In Section~\ref{sec:example_3}, we highlight the invariance of our Riccati-based approach to the order of the data, which provides flexibility in how the data is sampled. 

The contributions of this work can be summarized as follows:
\begin{itemize}
    \item new mathematical theory connecting certain Bayesian inference problems arising from SciML to viscous HJ PDEs,
    \item new Riccati-based methodology for solving Bayesian inference problems with linear models, Gaussian likelihoods, and  Gaussian priors, and
    \item detailed experimental demonstrations of the potential computational advantages of this Riccati-based approach across a variety of learning applications.
\end{itemize}
This work presents exciting opportunities to advance the theoretical foundations of SciML and thus to create new interpretable SciML methods. While we demonstrate promising results, further research is needed to extend the work presented here to more general learning settings.
We discuss some possible future research directions of this work in Section~\ref{sec:summary}. Some additional technical details and numerical results are provided in the Appendix.

%% file: sec_2.tex
\section{Connecting viscous HJ PDEs to Bayesian inference in machine learning}\label{sec:general_connection}

In this section, we establish a new theoretical connection between viscous HJ PDEs and Bayesian inference in machine learning (ML). 
Specifically, we formulate a general regression task in the Bayesian framework and then connect it to a viscous HJ PDE with quadratic Hamiltonian.

\subsection{Bayesian inference in machine learning}\label{sec:learning_prob}

Bayesian inference is often adopted in ML  due to its effectiveness in integrating prior information, handling noisy data, providing uncertainty quantification and, hence, in making quantifiably trustworthy predictions. Consider the following model \cite{williams2006gaussian}: 
\begin{equation}\label{eq:noisy_model}
    \MLy = u_\weightvec(\MLx) + \noise,
\end{equation}
where $\MLx\in\R^{\ell}$ is the input, $\MLy\in\R^m$ is the observed output, $u_\weightvec:\R^{\ell}\to\R^m$ is the  model parameterized by $\weightvec\in\Rn$, and $\noise\in\R^m$ is additive noise. The goal is to learn $u_\weightvec$ from data $\mathcal{D} = \{(\MLx_i, \MLy_i)\}_{i=1}^N$, where 
$\MLx_i, \MLy_i$ denote measurements of $\MLx, \MLy$, respectively. 
Specifically, we learn the model $u_\weightvec$ by estimating $\weightvec$ from the posterior distribution $p(\weightvec|\mathcal{D})$ (e.g., $\weightvec$ could be the posterior mean estimator $\E_{\weightvec\sim p(\cdot | \mathcal{D})}[\weightvec]$ or the maximum a posteriori (MAP) estimator $\arg\max_\weightvec p(\weightvec|\mathcal{D})$). 
By Bayes' theorem, $p(\weightvec|\mathcal{D})$ can be computed as
\begin{equation}\label{eq:posterior}
    p(\weightvec|\mathcal{D}) = \frac{p(\mathcal{D}|\weightvec) p(\weightvec)}{p(\mathcal{D})},
\end{equation}
where $p(\mathcal{D}|\weightvec)$ is the likelihood, $p(\weightvec)$ is the prior, and $p(\mathcal{D}) = \int_{\Rn} p(\mathcal{D}|\weightvec) p(\weightvec) d\weightvec$ is the marginal likelihood, which is independent of $\weightvec$ and often treated as a normalizing constant. 
From these distributions, we obtain quantities that can be used to statistically analyze and quantify the uncertainty in the learned model arising from the noisy data and/or the prior on $\weightvec$.

The posterior $p(\weightvec|\mathcal{D})$ often becomes analytically intractable \cite{psaros2023uncertainty, zou2022neuraluq}, particularly when either the likelihood or the prior is complicated. In these scenarios, estimation techniques, such as Markov Chain Monte Carlo (MCMC) \cite{neal2011mcmc}, are typically employed. Here, we create a new theoretical connection between HJ PDEs and Bayesian inference in ML that yields opportunities for new HJ PDE-based methods for estimating $\weightvec$ and $p(\weightvec|\mathcal{D})$. 
As a first exploration of this connection, we consider the case where the likelihood $p(\mathcal{D} | \weightvec)$ follows a Gaussian distribution and the model $u_\weightvec$  depends linearly on $\weightvec$. In other words, we consider the case where the data is perturbed by additive white noise (i.e., $\noise$ is Gaussian), and we learn the model $u_\weightvec(\MLx) = \Phi(\MLx)\weightvec$, where $\Phi:\R^{\ell}\to\R^{m\times n}$. 
With some assumptions on the prior, this scenario yields a connection to a viscous HJ PDE with a quadratic Hamiltonian. 



\subsection{Connection to viscous HJ PDEs}\label{sec:connection}

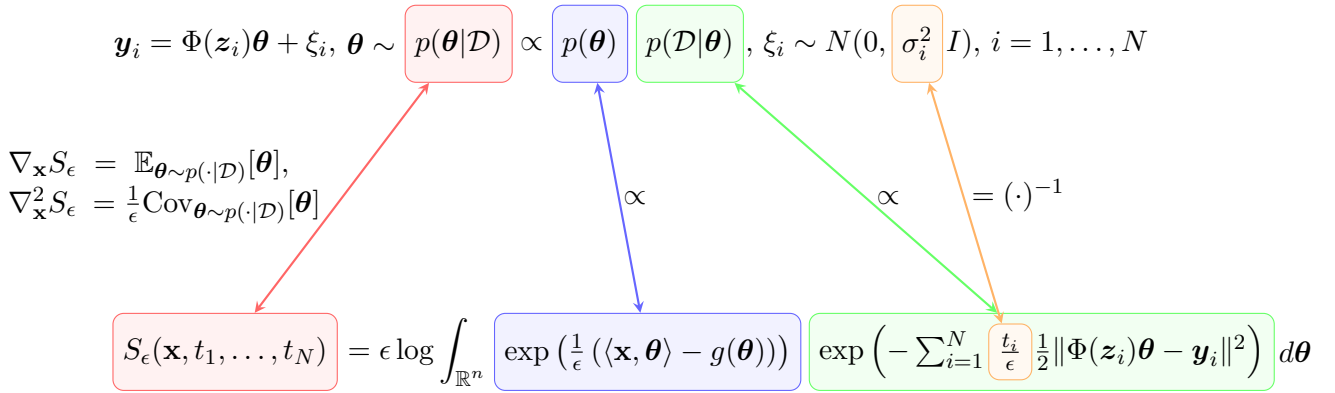
\begin{figure}[htbp]
\centering
\begin{adjustbox}{width=\textwidth}
\begin{tikzpicture}[node distance=2cm]
\node (model) [nobox] {$\MLy_i = \MLbasismat(\MLx_i)\weightvec + \noise_i$,};
\node (weight) [nobox, right of=model, xshift=-0.1cm] {$\weightvec\sim$};
\node (posterior) [box, right of=weight, draw=red!60, fill=red!5, xshift=-0.86cm] {$p(\weightvec|\mathcal{D})$};
\node (propto) [nobox, right of=posterior, xshift=-1.05cm] {$\propto$};
\node (prior) [box, right of=propto, draw=blue!60, fill=blue!5, xshift=-1.24cm] {$p(\weightvec)$};
\node (likelihood) [box, right of=prior, draw=green!60, fill=green!5, xshift=-0.7cm] {$p(\mathcal{D|}\weightvec)$};
\node (noise) [nobox, right of=likelihood, xshift=-0.36cm] {, $\noise_i\sim N(0,$};
\node (var) [box, right of=noise, draw=orange!60, fill=orange!5, xshift=-0.68cm] {$\sigma_i^2$};
\node (index) [nobox, right of=var, xshift=-0.31cm] {$I)$, $i = 1, \dots, N$};

\node (empty) [nobox, below of=model] { };

\node (S) [box, below of=empty, draw=red!60, fill=red!5] {$\Svis(\HJx, \HJt_1, \dots, \HJt_N)$};
\node (int) [nobox, right of=S, xshift=0.5cm] {$=\epsilon\log\mathlarger{\int}_\Rn$};
\node (IC) [box, right of=int, draw=blue!60, fill=blue!5, xshift=1cm] {$\exp\left(\frac{1}{\epsilon}\left(\langle\HJx, \weightvec\rangle - \HJprior(\weightvec)\right)\right)$};
\node (H) [box, right of=IC, draw=green!60, fill=green!5, xshift=3.13cm] {$\exp \left(-\sum_{i=1}^N\quad \,\,\,\frac{1}{2}\|\MLbasismat(\MLx_i)\weightvec-\MLy_i\|^2\right)$};
\node (HJt) [smallbox, right of=IC, draw=orange!60, fill=orange!5, xshift=2.72cm] {$\frac{\HJt_i}{\epsilon}$};
\node (dtheta) [nobox, right of=H, xshift=1.3cm] {$d\weightvec$};

\draw [doublearrow, draw=red!60] (posterior) -- (S) node[midway, left, xshift=0.3cm, yshift=0.15cm, text width=4.5cm] {$\nabla_\HJx\Svis=\E_{\weightvec\sim p(\cdot|\mathcal{D})}[\weightvec]$, $\nabla_\HJx^2\Svis \hspace{0.1cm}= \frac{1}{\epsilon}\Cov_{\weightvec\sim p(\cdot|\mathcal{D})}[\weightvec]$};
\draw [doublearrow, draw=blue!60] (prior) -- (IC) node[midway, right, xshift=-0.08cm] {$\propto$};
\draw [doublearrow, draw=green!60] (likelihood) -- (H)  node[midway, right] {$\propto$};
\draw [doublearrow, draw=orange!60] (var) -- (HJt) node[midway, right, xshift=-0.06cm, yshift=0.1cm] {$=(\cdot)^{-1}$};
\end{tikzpicture}
\end{adjustbox}

\caption{(See Section~\ref{sec:general_connection}) Mathematical formulation of the connection between a Bayesian inference problem with linear model and Gaussian likelihood (\textbf{top}) and the solution to a multi-time viscous HJ PDE with quadratic Hamiltonian (\textbf{bottom}). The content of this illustration matches that of Figure~\ref{fig:intro_connection_in_words} by replacing each term in Figure~\ref{fig:intro_connection_in_words} with its corresponding mathematical expression. The colors indicate the associated quantities. The arrow labels describe how the boxed quantities are related.}
\label{fig:connection_in_math}
\end{figure}

In this section, we detail the connection between certain Bayesian inference problems with Gaussian likelihood and viscous HJ PDEs with quadratic Hamiltonian. 
Specifically, we consider the case where we learn the model $u_\weightvec(\MLx) = \Phi(\MLx)\weightvec$, where $\Phi:\R^{\ell}\to\R^{m\times n}$ and the posterior of the parameter $\weightvec$ is estimated using the Bayesian framework described in Section~\ref{sec:learning_prob}. We assume that the likelihood is Gaussian given by $p(\mathcal{D}|\HJp) \propto \prod_{i=1}^N\exp\left(-\frac{1}{2\sigma_i^2}\|\Phi(\MLx_i) \HJp - \MLy_i\|^2\right)$, where $\sigma_i^2$ is the variance of the $i$-th data point $(\MLx_i, \MLy_i)$. 
We assume that the prior distribution is given by $p(\HJp)\propto \exp(-\frac{1}{\epsilon}(-\langle \HJx, \HJp\rangle + \HJprior(\HJp)))$, where $\epsilon>0$ is a hyperparameter and $\HJprior:\Rn\to\R$ is such that $\HJp\mapsto \exp(-\frac{1}{\epsilon}(-\langle \HJx, \HJp\rangle + \HJprior(\HJp)))$ is integrable for every $\HJx\in\Rn$. Hence, if $\HJprior$ is quadratic, then the prior distribution is also Gaussian. 
Note that we use $p(\weightvec)\propto f(\weightvec)$ to denote that the probability density function $p(\weightvec)$ is proportional to $f(\weightvec)$, or, in other words, $p(\weightvec) = \frac{f(\weightvec)}{\int_{\Rn} f(w)dw}$. Similarly, for conditional probabilities,  $p(\weightvec|\mathcal{D})\propto f(\weightvec,\mathcal{D})$ means  that $p(\weightvec|\mathcal{D}) = \frac{f(\weightvec,\mathcal{D})}{\int_{\Rn} f(w,\mathcal{D})dw}$.

\begin{rem}
The particular form of the likelihood that we consider corresponds to the data $\mathcal{D}$ being produced by $\MLy_i = u_{\weightvec}(\MLx_i) + \noise_i$, where $\noise_1,\dots, \noise_N$ are independent, each $\noise_i$ represents Gaussian noise with zero mean and covariance matrix $\sigma_i^2I$ ($I$ is the $n\times n$ identity matrix), and $u_{\weightvec}(\MLx) = \Phi(\MLx)\weightvec$. 
The results in this paper have a straightforward generalization to the case where $\noise_1,\dots, \noise_N$ are independent, each $\noise_i$ is a general Gaussian, and $u_{\weightvec}(\MLx) = \Phi(\MLx)\weightvec + c(\MLx)$, where $c:\R^\ell\to\R^m$ is a function independent of $\weightvec$. However, for simplicity, we only present the case where $c(\MLx) = 0$. 
\end{rem}


First consider the case where we have $N=1$ data point $(\MLx_1, \MLy_1)$. Denote $\Phi_1:=\Phi(\MLx_1)$. 
Define a Hamiltonian $H\colon \Rn\to \R$ by $H(\HJp) = \frac{1}{2}\|\Phi_1 \HJp-\MLy_1\|^2$ and a function $\Svis\colon \Rn\times[0,\infty)\to \R$ by
\begin{equation}
\Svis(\HJx,t) = \epsilon\log\int_{\Rn}  \exp\left(\frac{1}{\epsilon}(\langle \HJx, \HJp\rangle - tH(\HJp) - \HJprior(\HJp))\right) d\HJp.
\end{equation}
Then, $S_\epsilon$ satisfies the following viscous HJ PDE:
\begin{equation}\label{eqt:viscousHJ_dual}
\begin{dcases}
    \partial_t \Svis(\HJx,\HJt) + \frac{1}{2}\|\MLbasismat_1 \nabla_\HJx \Svis(\HJx,\HJt) - \MLy_1\|^2 + \frac{\epsilon}{2}\nabla_\HJx \cdot(\MLbasismat_1^T\MLbasismat_1\nabla_\HJx \Svis(\HJx,\HJt)) = 0 & \HJx\in\Rn, \HJt > 0, \\
    \Svis(\HJx,0) = J(\HJx) & \HJx\in \Rn,
\end{dcases}
\end{equation}
where the initial condition $J$ is defined by
\begin{equation}\label{eqt:def_J_Laplace_transform}
J(\HJx) = \epsilon\log \int_{\Rn} \exp\left(\frac{1}{\epsilon}(\langle \HJx, \HJp\rangle - \HJprior(\HJp))\right) d\HJp.
\end{equation}
Using this HJ PDE framework, the posterior distribution $p(\weightvec | \mathcal{D})$ can be characterized by $p(\weightvec | \mathcal{D}) \propto \exp\big(\frac{1}{\epsilon}(\langle \HJx, \HJp\rangle - tH(\HJp) - \HJprior(\HJp))\big)$ by setting the variance of the noise to $\sigma_1^2= \frac{\epsilon}{t}$.
By straightforward computation, the derivatives of $\Svis$ are given by
\begin{equation}
\begin{split}
\partial_t \Svis(\HJx,t) &= -\frac{\int_{\Rn} H(\HJp) \exp\left(\frac{1}{\epsilon}(\langle \HJx, \HJp\rangle - tH(\HJp) - \HJprior(\HJp))\right) d\HJp}{\int_{\Rn} \exp(\frac{1}{\epsilon}\left(\langle \HJx, \HJp\rangle - tH(\HJp) - \HJprior(\HJp))\right) d\HJp} = -\E_{\HJp\sim p(\cdot | \mathcal{D})}[H(\HJp)],\\
\nabla_\HJx \Svis(\HJx,t) &= \frac{\int_{\Rn} \HJp \exp\left(\frac{1}{\epsilon}(\langle \HJx, \HJp\rangle - tH(\HJp) - \HJprior(\HJp))\right) d\HJp}{\int_{\Rn} \exp\left(\frac{1}{\epsilon}(\langle \HJx, \HJp\rangle - tH(\HJp) - \HJprior(\HJp))\right) d\HJp} = \E_{\HJp\sim p(\cdot | \mathcal{D})}[\HJp],\\
\nabla_\HJx^2 \Svis(\HJx,t) &= \frac{1}{\epsilon}\left(\E_{\HJp\sim p(\cdot | \mathcal{D})}\left[\HJp\HJp^T\right] - \E_{\HJp\sim p(\cdot | \mathcal{D})}[\HJp]\E_{\HJp\sim p(\cdot | \mathcal{D})}[\HJp]^T\right) = \frac{1}{\epsilon}\Cov_{\HJp\sim p(\cdot | \mathcal{D})}[\HJp],
\end{split}
\end{equation}
all of which are related to the posterior expectation and posterior covariance. Namely, computing the first and second derivatives of the solution to the viscous HJ PDE~\eqref{eqt:viscousHJ_dual} is equivalent (up to a multiplicative constant) to computing the first and second moments of the posterior distribution, respectively, and the posterior mean estimator for $\weightvec$ can be computed as $\nabla_\HJx S_\epsilon(\HJx,t)$.


These connections can be easily generalized to the case with multiple data points ($N> 1$). 
Denote $\Phi_i = \Phi(\MLx_i)$, and set $\frac{\epsilon}{t_i} = \sigma_i^2$ for each $i=1,\dots, N$. The posterior distribution then becomes $p(\weightvec | \mathcal{D}) \propto \exp\left(\frac{1}{\epsilon}(\langle \HJx, \HJp\rangle - \sum_{i=1}^N \frac{t_i}{2}\|\Phi_i\HJp-\MLy_i\|^2 - \HJprior(\HJp))\right)$.
This learning problem is related to a multi-time viscous HJ PDE as follows. Define the $i$-th Hamiltonian $H_i:\Rn\to\R$ by $H_i(\HJp) = \frac{1}{2}\|\Phi_i \HJp-\MLy_i\|^2$ and the function $\Svis\colon \Rn\times [0,\infty)^N\to\R$ by
\begin{equation}\label{eq:hj_solution}
\Svis(\HJx,t_1, \dots, t_N) = \epsilon\log \int_{\Rn} \exp\left(\frac{1}{\epsilon}\left(\langle \HJx, \HJp\rangle - \sum_{i=1}^N\frac{t_i}{2}\|\MLbasismat_i \HJp - \MLy_i\|^2 - \HJprior(\HJp)\right)\right) d\HJp.
\end{equation}
Then, $\Svis$ satisfies the following multi-time viscous HJ PDE:
\begin{equation}\label{eqt:viscousHJ_multitime}
\begin{adjustbox}{width=0.99\textwidth}
$\begin{dcases}
\partial_{t_i} \Svis(\HJx,t_1, \dots, t_N) + \frac{1}{2}\|\MLbasismat_i \nabla_\HJx \Svis(\HJx,t_1, \dots, t_N) - \MLy_i\|^2 + \frac{\epsilon}{2}\nabla_\HJx \cdot(\MLbasismat_i^T\MLbasismat_i\nabla_\HJx \Svis(\HJx,t_1, \dots, t_N)) = 0 & \HJx\in\Rn, t_i>0, i = 1,\dots, N, \\
\Svis(\HJx,0,\dots, 0) = J(\HJx) & \HJx\in\Rn,
\end{dcases}$
\end{adjustbox}
\end{equation}
where the initial condition $J$ is defined in~\eqref{eqt:def_J_Laplace_transform}. 
Using the notation of the Hamiltonians $H_i$, the posterior distribution is also given by $p(\weightvec | \mathcal{D}) \propto \exp\left(\frac{1}{\epsilon}(\langle \HJx, \HJp\rangle - \sum_{i=1}^N t_iH_i(\HJp) - \HJprior(\HJp))\right)$.
With straightforward computation, the derivatives of $\Svis$ are given by
\begin{equation}
\begin{split}
\partial_{t_i} \Svis(\HJx,t_1, \dots, t_N) &= -\frac{\int_{\Rn} H_i(\HJp) \exp\left(\frac{1}{\epsilon}(\langle \HJx, \HJp\rangle - \sum_{i=1}^N t_iH_i(\HJp) - \HJprior(\HJp))\right) d\HJp}{\int_{\Rn} \exp\left(\frac{1}{\epsilon}(\langle \HJx, \HJp\rangle - \sum_{i=1}^N t_iH_i(\HJp) - \HJprior(\HJp))\right) d\HJp} = -\E_{\HJp\sim p(\cdot|\mathcal{D})}[H_i(\HJp)],\\
\nabla_\HJx \Svis(\HJx,t_1, \dots, t_N) &= \frac{\int_{\Rn} \HJp \exp\left(\frac{1}{\epsilon}(\langle \HJx, \HJp\rangle - \sum_{i=1}^N t_iH_i(\HJp) - \HJprior(\HJp))\right) d\HJp}{\int_{\Rn} \exp\left(\frac{1}{\epsilon}(\langle \HJx, \HJp\rangle - \sum_{i=1}^N t_iH_i(\HJp) - \HJprior(\HJp))\right) d\HJp} = \E_{\HJp\sim p(\cdot|\mathcal{D})}[\HJp],\\
\nabla_\HJx^2 \Svis(\HJx,t_1, \dots, t_N) &= \frac{1}{\epsilon}\left(\E_{\HJp\sim p(\cdot|\mathcal{D})}\left[\HJp\HJp^T\right] - \E_{\HJp\sim p(\cdot|\mathcal{D})}[\HJp]\E_{\HJp\sim p(\cdot|\mathcal{D})}[\HJp]^T\right) = \frac{1}{\epsilon}\Cov_{\HJp\sim p(\cdot|\mathcal{D})}[\HJp].
\end{split}
\end{equation}
As in the $N=1$ case, $\nabla_\HJx \Svis(\HJx,t_1,\dots, t_N)$ and $\epsilon\nabla_\HJx^2 \Svis(\HJx,t_1,\dots, t_N)$ are equivalent to the posterior expectation and posterior covariance matrix, respectively. These connections can then be leveraged to compute the posterior mean estimator and to estimate the uncertainty in the learned model.  Figure~\ref{fig:connection_in_math} summarizes the connections established above between Bayesian inference problems and multi-time viscous HJ PDEs.

%% file: sec_3.tex
\section{Riccati-based methodology}
\label{sec:riccatimethod}
In the previous section, we established a new theoretical connection between Bayesian inference problems with linear models and additive Gaussian noise and viscous HJ PDEs with quadratic Hamiltonians. In this section, we show how this connection can be leveraged to reuse existing HJ PDE solvers to create new efficient training methods for Bayesian inference. While the theoretical connection holds more generally, in this section, we focus on the case where the prior distribution is also Gaussian. In this case, the corresponding viscous HJ PDEs can be solved using Riccati ODEs, which in turn can be used to solve the original Bayesian inference problem of interest. 
Specifically, we develop a new Riccati-based approach for solving these Bayesian inference problems that
\begin{enumerate}
    \item can add and remove data points from the training dataset invariant to the order of the data,
    \item does not require retraining on or access to previously incorporated data to update the learned model, which may provide computational benefits, particularly when the posterior distribution must be re-computed many times, and
    \item yields a continuous flow of solutions, which allows the hyperparameters in the learning problem to be tuned continuously.
\end{enumerate}


\subsection{Connection using Gaussian prior to Riccati ODEs}\label{sec:quadIC}

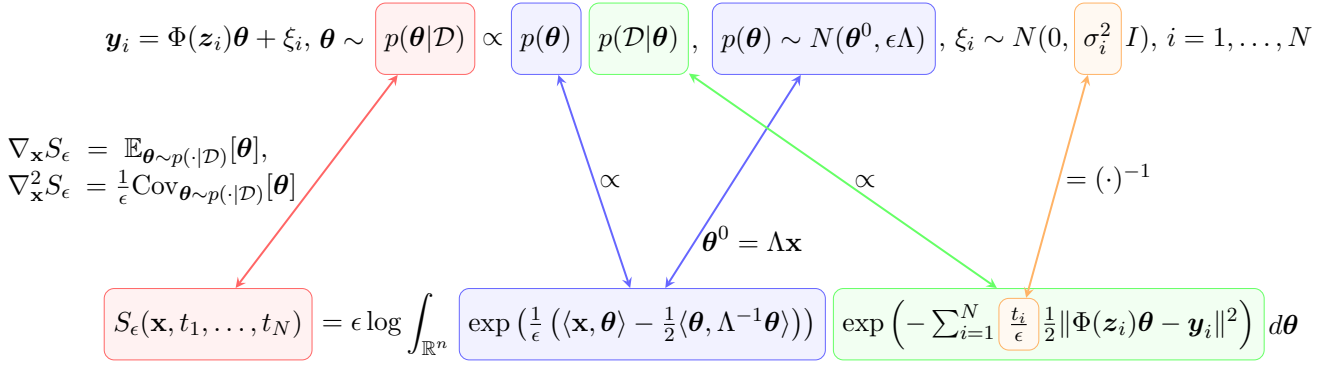
\begin{figure}[htbp]
\centering
\begin{adjustbox}{width=\textwidth}
\begin{tikzpicture}[node distance=2cm]
\node (model) [nobox] {$\MLy_i = \MLbasismat(\MLx_i)\weightvec + \noise_i$,};
\node (weight) [nobox, right of=model, xshift=-0.12cm] {$\weightvec\sim$};
\node (posterior) [box, right of=weight, draw=red!60, fill=red!5, xshift=-0.83cm] {$p(\weightvec|\mathcal{D})$};
\node (propto) [nobox, right of=posterior, xshift=-1.05cm] {$\propto$};
\node (prior) [box, right of=propto, draw=blue!60, fill=blue!5, xshift=-1.25cm] {$p(\weightvec)$};
\node (likelihood) [box, right of=prior, draw=green!60, fill=green!5, xshift=-0.7cm] {$p(\mathcal{D|}\weightvec)$};
\node (comma) [nobox, right of=likelihood, xshift=-1.21cm, yshift=-0.15cm] {$,$};
\node (priorN) [box, right of=likelihood, draw=blue!60, fill=blue!5, xshift=0.6cm] {$p(\weightvec) \sim N(\weightvec^0, \epsilon\Sigmainprior)$};
\node (noise) [nobox, right of=priorN, xshift=0.53cm] {, $\noise_i\sim N(0,$};
\node (var) [box, right of=noise, draw=orange!60, fill=orange!5, xshift=-0.68cm] {$\sigma_i^2$};
\node (index) [nobox, right of=var, xshift=-0.3cm] {$I)$, $i = 1, \dots, N$};

\node (empty) [nobox, below of=model] { };

\node (S) [box, below of=empty, draw=red!60, fill=red!5] {$\Svis(\HJx, \HJt_1, \dots, \HJt_N)$};
\node (int) [nobox, right of=S, xshift=0.5cm] {$=\epsilon\log\mathlarger{\int}_\Rn$};
\node (IC) [box, right of=int, draw=blue!60, fill=blue!5, xshift=1.6cm] {$\exp\left(\frac{1}{\epsilon}\left(\langle\HJx, \weightvec\rangle - \frac{1}{2}\langle \HJp, \Sigmainprior^{-1} \HJp\rangle\right)\right)$};
\node (H) [box, right of=IC, draw=green!60, fill=green!5, xshift=3.71cm] {$\exp \left(-\sum_{i=1}^N\quad \,\,\,\frac{1}{2}\|\MLbasismat(\MLx_i)\weightvec-\MLy_i\|^2\right)$};
\node (HJt) [smallbox, right of=IC, draw=orange!60, fill=orange!5, xshift=3.29cm] {$\frac{\HJt_i}{\epsilon}$};
\node (dtheta) [nobox, right of=H, xshift=1.3cm] {$d\weightvec$};

\draw [doublearrow, draw=red!60] (posterior) -- (S) node[midway, left, xshift=0.3cm, yshift=0.15cm, text width=4.5cm] {$\nabla_\HJx\Svis=\E_{\weightvec\sim p(\cdot|\mathcal{D})}[\weightvec]$, $\nabla_\HJx^2\Svis \hspace{0.1cm}= \frac{1}{\epsilon}\Cov_{\weightvec\sim p(\cdot|\mathcal{D})}[\weightvec]$};
\draw [doublearrow, draw=blue!60] (prior) -- (IC) node[midway, right, xshift=-0.07cm] {$\propto$};
\draw [doublearrow, draw=blue!60] (priorN) -- (IC) node[midway, right, xshift=-0.58cm, yshift=-0.8cm] {$\weightvec^0 = \Sigmainprior\HJx$};
\draw [doublearrow, draw=green!60] (likelihood) -- (H)  node[midway, right, xshift=0.02cm] {$\propto$};
\draw [doublearrow, draw=orange!60] (var) -- (HJt) node[midway, right, xshift=-0.02cm, yshift=0.1cm] {$=(\cdot)^{-1}$};
\end{tikzpicture}
\end{adjustbox}

\caption{(See Section~\ref{sec:quadIC}) Mathematical formulation of the connection between a Bayesian inference problem with linear model, Gaussian likelihood, and Gaussian prior (\textbf{top}) and the solution to a multi-time viscous HJ PDE with quadratic Hamiltonian and quadratic initial condition (\textbf{bottom}). The content of this illustration is a special case of the connection in Figure~\ref{fig:connection_in_math}, where the prior is Gaussian (set $g(\HJp) = \frac{1}{2}\langle \HJp, \Sigmainprior^{-1} \HJp\rangle$ in Figure~\ref{fig:connection_in_math}). The colors indicate the associated quantities. The arrow labels describe how the boxed quantities are related.}
\label{fig:connection_riccati}
\end{figure}

In this section, we discuss the special case where the prior distribution is Gaussian with mean $\weightvec^0 = \Sigmainprior\HJx$ and covariance matrix $\epsilon \Sigmainprior$, where $\Sigmainprior\in\R^{n\times n}$ is symmetric positive definite. Note that this choice of mean and covariance is not as restrictive as the notation suggests since we have freedom in picking $\HJx\in\Rn$ (picking a different $\HJx$ corresponds to evaluating the associated viscous HJ PDE at a different point in space); as long as $\Sigmainprior$ is invertible (i.e., the prior is a non-degenerate Gaussian), $\weightvec^0$ can be chosen to be any point in $\Rn$.
This particular prior corresponds to setting $g(\HJp) = \frac{1}{2}\langle \HJp, \Sigmainprior^{-1} \HJp\rangle$, and the initial condition~\eqref{eqt:def_J_Laplace_transform} of the associated viscous HJ PDE becomes $J(\HJx) = \frac{1}{2}\langle \HJx, \Sigmainprior \HJx\rangle +\frac{\epsilon}{2} \log \det(\Sigmainprior) +\frac{\epsilon n}{2} \log (2\pi \epsilon)$. 
Since the viscous HJ PDE has quadratic Hamiltonian and quadratic initial condition, its solution $S_\epsilon$ can be computed using Riccati ODEs as follows.

We first consider the single data point case ($N=1$).
In this case, the solution $\Svis$ to the viscous HJ PDE~\eqref{eqt:viscousHJ_dual} is given by $\Svis(\HJx, t)=\frac{1}{2}\HJx^T\Sxx(t)\HJx + \Sx(t)^T\HJx + \Sc(t)$, where $\Sxx\colon [0,+\infty)\to \R^{n\times n}$, $\Sx\colon [0,+\infty)\to\Rn$, and $\Sc\colon [0,+\infty)\to\R$ satisfy the following Riccati ODEs: 
\begin{equation}\label{eqt:Riccati_singletime}
\begin{dcases}
\dot \Sxx(s) + \Sxx(s)\MLbasismatone^T\MLbasismatone \Sxx(s) = 0,\\
\dot \Sx(s) + \Sxx(s)\MLbasismatone^T(\MLbasismatone \Sx(s) - \MLy_1) = 0,\\
\dot \Sc(s) + \frac{1}{2}\|\MLbasismatone \Sx(s) - \MLy_1\|^2 + \frac{\epsilon}{2} Tr(\MLbasismatone^T\MLbasismatone \Sxx(s)) = 0,
\end{dcases}
\quad\quad\quad\quad 
\begin{dcases}
\Sxx(0) = \Sigmainprior, \\
\Sx(0) = 0, \\
\Sc(0) = \frac{\epsilon}{2} \log \det(\Sigmainprior) + \frac{\epsilon n}{2} \log (2\pi \epsilon).
\end{dcases}
\end{equation}
The mean and covariance matrix of the posterior distribution are given by $\E_{\weightvec\sim p(\cdot|\mathcal{D})}[\weightvec] = \nabla_{\HJx} \Svis(\HJx,t) = \Sxx(t) \HJx + \Sx(t)$ and $\Cov_{\weightvec\sim p(\cdot|\mathcal{D})}[\weightvec] = \epsilon\nabla_{\HJx}^2 \Svis(\HJx, t) = \epsilon\Sxx(t)$, respectively. Moreover, the MAP estimator and posterior mean estimator are equivalent in this case (since the posterior is Gaussian) and can be computed as $\weightvec = \nabla_{\HJx} \Svis(\HJx,t)= \Sxx(t) \HJx + \Sx(t)$.
Note that none of these formulas involve $\Sc$, and hence, if we only care about solving the associated learning problem, the ODE for $\Sc$ may be ignored.
The ODE system~\eqref{eqt:Riccati_singletime} also has an analytical solution given by
$\Sxx(s) = (\Sigmainprior^{-1} + s\MLbasismatone^T\MLbasismatone)^{-1}$, 
$\Sx(s) = s\Sxx(s)\MLbasismatone^T\MLy_1$, which can be computed directly when the dimension $n$ is small.  However, when $n$ is large, inverting an $n\times n$ matrix becomes inefficient, and  using other numerical methods (e.g., Runge-Kutta method, recursive least squares) to solve the Riccati ODEs~\eqref{eqt:Riccati_singletime} may be preferable.

Now consider the case where we have $N>1$ data points. Then, the solution $\Svis$ to the multi-time viscous HJ PDE~\eqref{eqt:viscousHJ_multitime} satisfies
$\Svis(\HJx,t_1,\dots, t_N) = \frac{1}{2}\HJx^T\Sxx\left(\sum_{i=1}^N t_i\right)\HJx + \Sx\left(\sum_{i=1}^N t_i\right)^T\HJx + \Sc\left(\sum_{i=1}^N t_i\right)$.
Here, $\Sxx\colon [0,+\infty)\to \R^{n\times n}$, $\Sx\colon [0,+\infty)\to\Rn$, and $\Sc\colon [0,+\infty)\to\R$ are three continuous functions that satisfy the following ODE system over the interval $s\in \left[\sum_{i=1}^{j-1} t_i, \sum_{i=1}^{j} t_i\right]$ for each $j=1,\dots, N$:
\begin{equation}\label{eqt:RiccatiODE_k}
\begin{dcases}
\dot \Sxx(s) + \Sxx(s)\MLbasismat_j^T\MLbasismat_j P(s) = 0,\\
\dot \Sx(s) + \Sxx(s)\MLbasismat_j^T(\MLbasismat_j \Sx(s) - \MLy_j) = 0,\\
\dot \Sc(s) + \frac{1}{2}\|\MLbasismat_j \Sx(s) - \MLy_j\|^2 + \frac{\epsilon}{2} Tr(\MLbasismat_j^T\MLbasismat_j\Sxx(s)) = 0, 
\end{dcases}
\end{equation}
where the initial condition is the same as in~\eqref{eqt:Riccati_singletime}.
The posterior mean, posterior covariance, and MAP/posterior mean estimator are computed identically as in the $N=1$ case but with $t = \sum_{i=1}^{N} t_i$.
This ODE system also has an analytical solution: 
$\Sxx(s) = \left(\Sigmainprior^{-1} + \sum_{i=1}^{j-1} t_i\MLbasismat_i^T\MLbasismat_i + (s - \sum_{i=1}^{j-1} t_i)\MLbasismat_j^T\MLbasismat_j\right)^{-1}$,
$\Sx(s) = \Sxx(s)\left(\sum_{i=1}^{j-1} t_i\MLbasismat_i^T \MLy_i + (s- \sum_{i=1}^{j-1} t_i)\MLbasismat_j^T \MLy_j\right)$ for $s\in\left[\sum_{i=1}^{j-1} t_i, \sum_{i=1}^{j} t_i\right]$ and $j=1,\dots, N$. 
When $n$ is relatively small and there are no computational restrictions on accessing all $N$ data points at once, these analytical formulas can be used to compute the posterior mean and covariance. However, when the dimension $n$ is large or there are computational restrictions on the storage of or access to the entire dataset, using numerical methods to solve the Riccati ODEs~\eqref{eqt:RiccatiODE_k} may be more advantageous than using the analytical formulas. Detailed discussions of some potential computational advantages are presented in the remainder of Section~\ref{sec:riccatimethod}. The theoretical connection using a Gaussian prior is summarized in Figure~\ref{fig:connection_riccati}.

\subsection{Updating the likelihood}\label{sec:3_2}

In this section, we discuss how the likelihood can be updated efficiently using our Riccati-based methodology. Namely, we show how data points can be added or removed from the learned model without retraining on or accessing the entire previous training set and in a manner that is invariant to the order of the data points. We also show how this Riccati-based framework naturally yields a continuous flow of solutions that allows hyperparameters to be tuned continuously. 

\subsubsection{Adding and removing data points}\label{sec:3_2_1}
First, consider adding a data point $(\MLx_{N+1}, \MLy_{N+1})$ to the training set $\mathcal{D}$. Recalling that we assume each data point is independent, the posterior distribution after adding the $(N+1)$th data point becomes 
\begin{equation}\label{eq:new_posterior}
    p(\weightvec|\mathcal{D}, (\MLx_{N+1}, \MLy_{N+1}))\propto p(\weightvec|\mathcal{D})\exp\left(-\frac{ 1}{2\sigma_{N+1}^2}\|\MLbasismat_{N+1}\weightvec - \MLy_{N+1}\|^2\right),
\end{equation}
where $\MLbasismat_{N+1} = \MLbasismat(\MLx_{N+1})$.
In the HJ PDE framework, this corresponds to augmenting the multi-time system~\eqref{eqt:viscousHJ_multitime} with the following HJ PDE:
\begin{equation}
\begin{adjustbox}{width=0.99\textwidth}$
\partial_{t_{N+1}} \Svis(\HJx,t_1, \dots, t_{N+1}) + \frac{1}{2}\|\MLbasismat_{N+1} \nabla_\HJx \Svis(\HJx,t_1, \dots, t_{N+1}) - \MLy_{N+1}\|^2 + \frac{\epsilon}{2}\nabla_\HJx \cdot(\MLbasismat_{N+1}^T\MLbasismat_{N+1}\nabla_\HJx \Svis(\HJx,t_1, \dots, t_{N+1})) = 0.
$\end{adjustbox}
\end{equation}
Thus, adding an additional data point can be interpreted as evolving a new time variable $t_{N+1}$ from 0 to $\frac{\epsilon}{\sigma_{N+1}^2}$, where $\sigma_{N+1}^2$ denotes the variance of the $(N+1)$th data point. As such, the new posterior mean 
\begin{equation}
 \E_{\weightvec\sim p(\cdot|\mathcal{D}, (\MLx_{N+1}, \MLy_{N+1}))}[\weightvec]=\Sxx\left(\sum_{i=1}^{N+1} t_i\right)\HJx + \Sx\left(\sum_{i=1}^{N+1} t_i\right) 
\end{equation}
and posterior covariance
\begin{equation}
\Cov_{\weightvec\sim p(\cdot|\mathcal{D}, (\MLx_{N+1}, \MLy_{N+1}))}[\weightvec] =\epsilon\Sxx\left(\sum_{i=1}^{N+1} t_i\right)
\end{equation}
can be computed by solving the Riccati ODE~\eqref{eqt:RiccatiODE_k} (with index $j=N+1$) on $\left(\sum_{i=1}^{N} t_i, \sum_{i=1}^{N+1} t_i\right)$ with initial conditions $\Sxx\left(\sum_{i=1}^{N} t_i\right)=\frac{\Sigma_{N}}{\epsilon}$ and $\Sx\left(\sum_{i=1}^{N} t_i\right)=\mu_{N} - \frac{1}{\epsilon}\Sigma_N \HJx$, where $\mu_N := \E_{\weightvec\sim p(\cdot|\mathcal{D})}[\weightvec]$ and $\Sigma_N := \Cov_{\weightvec\sim p(\cdot|\mathcal{D})}[\weightvec]$ denote the original posterior mean and covariance, respectively.


Next, we discuss removing one data point $(\MLx_k, \MLy_k)$, $k\in\{1, \dots, N\}$ from $\mathcal{D}$. In the HJ PDE framework, removing the $k$th data point corresponds to evolving the time variable $t_k$ from $\frac{\epsilon}{\sigma_k^2}$ to 0. Thus, the new posterior mean and covariance resulting from deleting the $k$th data point can be computed as $\Sxx(\sum_{i=1}^{N} t_i-t_k)\HJx + \Sx(\sum_{i=1}^{N} t_i-t_k)$ and $\epsilon\Sxx(\sum_{i=1}^{N} t_i-t_k)$, respectively, where $\Sxx, \Sx$ are obtained by solving the Riccati ODE~\eqref{eqt:RiccatiODE_k} (with index $j=k$) backwards on $(\sum_{i=1}^{N} t_i-t_k, \sum_{i=1}^{N} t_i)$ with terminal condition $\Sxx(\sum_{i=1}^{N} t_i) = \frac{\Sigma_{N}}{\epsilon}$ and $\Sx(\sum_{i=1}^{N} t_i) = \mu_{N} - \frac{1}{\epsilon}\Sigma_N \HJx$, where we again denote by $\mu_N, \Sigma_N$ the original posterior mean and covariance using all $N$ data points.

Note that when adding or removing a data point, we do not require access to the entire previous dataset $\mathcal{D}$. Instead, we only require access to the previous posterior mean $\mu_N$, the previous posterior covariance $\Sigma_N$, and the data point to be added or removed; all of the information about the rest of the training data remains encoded in the solution to the HJ PDE. As such, our Riccati-based approach provides potential computational and memory advantages when adding or removing data points by allowing the posterior distribution to be updated without having to retrain on the entire dataset. Additionally, note that, due to our assumption of independent data, the posterior mean and covariance are invariant to the order of the data points. As a result, adding or removing data points using our Riccati-based approach is also invariant to the order of the data; i.e., the final learned model will be the same regardless of the order in which data points are added or removed.


\subsubsection{Tuning the variance}

Tuning the variance $\sigma_k^2$ of the $k$th data point can be done using a similar methodology as that for adding or removing a data point. 
If we decrease the hyperparameter from $\sigma_k^2$ to $\tilde\sigma_k^2$, the new posterior mean and covariance can be computed as $\Sxx(\frac{\epsilon}{\tilde\sigma_k^2} - \frac{\epsilon}{\sigma_k^2})\HJx + \Sx(\frac{\epsilon}{\tilde\sigma_k^2} - \frac{\epsilon}{\sigma_k^2})$ and $\epsilon\Sxx(\frac{\epsilon}{\tilde\sigma_k^2} - \frac{\epsilon}{\sigma_k^2})$, respectively, where $\Sxx, \Sx$ are obtained by solving the Riccati ODE~\eqref{eqt:RiccatiODE_k} (with index $j=k$) forward from $s=0$ to $s=\frac{\epsilon}{\tilde\sigma_k^2} - \frac{\epsilon}{\sigma_k^2}$ with initial condition $\Sxx(0)=\frac{\Sigma}{\epsilon}$ and $\Sx(0)=\mu-\frac{1}{\epsilon}\Sigma\HJx$, where we denote by $\mu, \Sigma$ the original posterior mean and covariance computed using the original value of $\sigma_k^2$. 
Note that every $s\in [0,\frac{\epsilon}{\tilde\sigma_k^2} - \frac{\epsilon}{\sigma_k^2}]$ can be written as $\frac{\epsilon}{\hat\sigma_k^2}-\frac{\epsilon}{\sigma_k^2}$ for some $\hat\sigma_k^2\in[\tilde\sigma_k^2, \sigma_k^2]$. Hence, as we evolve the Riccati ODE from $s=0$ to $s=\frac{\epsilon}{\tilde\sigma_k^2} - \frac{\epsilon}{\sigma_k^2}$, we simultaneously obtain a flow of solutions corresponding to each $\hat\sigma_k^2\in[\tilde\sigma_k^2, \sigma_k^2]$, or, in other words, we tune the variance continuously from $\sigma_k^2$ to $\tilde\sigma_k^2$.

Similarly, if we increase the hyperparameter from $\sigma_k^2$ to $\tilde\sigma_k^2$, the new posterior mean and covariance can be computed as $\Sxx(0)\HJx + \Sx(0)$ and $\epsilon\Sxx(0)$, respectively, where $\Sxx, \Sx$ are obtained by solving the Riccati ODE~\eqref{eqt:RiccatiODE_k} (with index $j=k$) backward from $s=\frac{\epsilon}{\sigma_k^2} - \frac{\epsilon}{\tilde\sigma_k^2}$ to $s=0$ with terminal condition $\Sxx(\frac{\epsilon}{\sigma_k^2} - \frac{\epsilon}{\tilde\sigma_k^2})=\frac{\Sigma}{\epsilon}$ and $\Sx(\frac{\epsilon}{\sigma_k^2} - \frac{\epsilon}{\tilde\sigma_k^2})=\mu-\frac{1}{\epsilon}\Sigma\HJx$. As in the previous case, evolving the Riccati ODE from $s=\frac{\epsilon}{\sigma_k^2} - \frac{\epsilon}{\tilde\sigma_k^2}$ to $s=0$ again corresponds to continuously tuning the variance from $\sigma_k^2$ to $\tilde\sigma_k^2$, and hence, every value of $\Sxx(s)\HJx + \Sx(s)$ and $\epsilon\Sxx(s)$, $s\in[0,\frac{\epsilon}{\sigma_k^2} - \frac{\epsilon}{\tilde\sigma_k^2}]$ corresponds to the posterior mean and covariance using a different variance for the $k$th data point.

\subsection{Updating the prior: tuning hyperparameters}\label{sec:3_3}

In this section, we discuss how our Riccati-based approach can be used to update the mean and covariance of the Gaussian prior. Tuning these hyperparameters may improve the inference accuracy of the learned model by updating the prior as new information becomes available. 


Changing the prior mean $\weightvec^0$ simply requires evaluating the solution to the corresponding HJ PDE at a different point  $\HJx = \Sigmainprior^{-1}\weightvec^0$ in space, which, in turn, only requires the posterior mean to be updated with the new $\HJx$-value using some matrix-vector multiplications and addition, while the posterior covariance remains unchanged. Hence, tuning the prior mean is relatively cheap computationally.


Consider changing the prior covariance from $\epsilon\Sigmainprior$ to $\epsilon\tilde \Sigmainprior$. Originally, the prior covariance appears in the initial condition of the sequence of Riccati ODEs~\eqref{eqt:RiccatiODE_k}. However, we do not have to re-solve the entire sequence of Riccati ODEs in order to update the prior covariance. Instead, we reinterpret the prior covariance as part of a Hamiltonian of the corresponding multi-time HJ PDE, which allows it to be tuned using only the results of training with the original prior. 
Denote the original posterior mean and covariance (computed using the the original prior) by $\mu_{pos}$ and $\Sigma_{pos}$, respectively. 
Then, the new posterior mean $\tilde\mu_{pos}$ and covariance $\tilde\Sigma_{pos}$ using the new prior covariance $\epsilon\tilde \Sigmainprior$ can be obtained using Riccati ODEs as follows. 
First, solve the Riccati ODE~\eqref{eqt:RiccatiODE_k} forward from $s=0$ to $s=1$ with $\MLbasismat_j = \tilde \Sigmainprior^{-1/2}$, $\MLy_j = 0$, and initial conditions $\Sxx(0) = \frac{\Sigma_{pos}}{\epsilon}$ and $\Sx(0) = \mu_{pos} - \frac{1}{\epsilon}\Sigma_{pos} \HJx$. This step incorporates information about the new prior covariance into the model and yields a solution $\Sxx(1)$ and $\Sx(1)$. Next, solve the Riccati ODE~\eqref{eqt:RiccatiODE_k} backward from $s=1$ to $s=0$ with $\MLbasismat_j = \Sigmainprior^{-1/2}$, $\MLy_j = 0$, and terminal conditions $\tilde \Sxx(1) = \Sxx(1)$ and $\tilde \Sx(1) = \Sx(1)$, where $\Sxx(1), \Sx(1)$ are the solutions from the previous step. This step removes information about the old prior covariance from the model and yields the solution $\tilde \Sxx(0)$ and $\tilde\Sx(0)$. Finally, the new posterior mean and covariance are computed as $\tilde\mu_{pos} = \tilde \Sxx(0) \HJx + \tilde \Sx(0)$ and $\tilde\Sigma_{pos} = \epsilon\tilde \Sxx(0)$, respectively.

When the new prior covariance $\epsilon\tilde \Sigmainprior$ is a scaling of the original prior covariance $\epsilon\Sigmainprior$ (i.e., $\tilde \Sigmainprior = \alpha \Sigmainprior$ for some $\alpha > 0$), the posterior can be updated by solving only one Riccati ODE instead of two. Again denote the original posterior mean and covariance computed using the original prior by $\mu_{pos}$ and $\Sigma_{pos}$. 
If $\alpha > 1$, then the new posterior mean and covariance can be computed as $\Sxx(\frac{1}{\alpha}) + \Sx(\frac{1}{\alpha})\HJx$ and $\epsilon\Sxx(\frac{1}{\alpha})$, respectively, where $\Sxx, \Sx$ are obtained by solving the Riccati ODE~\eqref{eqt:RiccatiODE_k} backward from $s = 1$ to $s = \frac{1}{\alpha}$ with $\MLbasismat_j = \Sigmainprior^{-1/2}$, $\MLy_j = 0$, and terminal conditions $\Sxx(1) = \frac{\Sigma_{pos}}{\epsilon}$ and $\Sx(1) = \mu_{pos} - \frac{1}{\epsilon}\Sigma_{pos} \HJx$. 
Similarly, if $\alpha \in (0,1)$, then the new posterior mean and covariance can be computed as $\Sxx(\frac{1}{\alpha}) + \Sx(\frac{1}{\alpha})\HJx$ and $\epsilon\Sxx(\frac{1}{\alpha})$, respectively, where $\Sxx, \Sx$ are obtained by solving the Riccati ODE~\eqref{eqt:RiccatiODE_k} forward from $s=1$ to $s=\frac{1}{\alpha}$ with $\MLbasismat_j = \Sigmainprior^{-1/2}$, $\MLy_j = 0$, and initial conditions $\Sxx(1) = \frac{\Sigma_{pos}}{\epsilon}$ and $\Sx(1) = \mu_{pos} - \frac{1}{\epsilon}\Sigma_{pos} \HJx$. In both cases, evolving the Riccati ODE yields a continuous flow of solutions corresponding to tuning the hyperparameter $\alpha$ continuously; namely, every $s$ between 1 and $1/\alpha$ corresponds to another solution with a different prior covariance $\frac{\epsilon}{s}\Sigmainprior$.

%% file: sec_4.tex
\section{Examples in scientific machine learning}\label{sec:examples}

In this section, we apply our methodology to three examples from SciML, in which ML tools are employed to solve ODEs or PDEs \cite{raissi2019physics, karniadakis2021physics}. 
In each example, we learn the solution $u:\R^\ell\to\R^m$ and the source term $f:\R^\ell\to\R^m$ of a differential equation
\begin{equation}
\begin{dcases}
\mathcal{F}[u] = f & \textrm{in } \Omega, \\ 
\mathcal{B}[u] = b & \textrm{on } \partial\Omega,
\end{dcases}
\end{equation}
using noisy data and quantify corresponding epistemic uncertainties \cite{hullermeier2021aleatoric, psaros2023uncertainty, zou2022neuraluq}. In the differential equation above, 
$\mathcal{F}$ and $\mathcal{B}$ represent general differential and boundary operators, respectively. Following the framework in Section~\ref{sec:learning_prob}, we estimate $u$ using the linear model $u_\weightvec = \Phi(\cdot)\weightvec$, where $\Phi:\R^\ell\to\R^{m\times n}$ is the matrix whose columns are the basis functions $\phi_i:\R^\ell\to\R^m$, $i=1, \dots, n$. 
As a result, we also estimate $f$ using a linear model $f_\weightvec = \mathcal{F}[u_\weightvec](\cdot) = \mathcal{F}[\Phi](\cdot)\weightvec$. 

We use the predicted mean associated with our learned models as our prediction of the quantities of interest; i.e., our prediction for $u$ is $\E[u_\weightvec|\mathcal{D}] = \Phi\E[\weightvec|\mathcal{D}]$ and our prediction for $f$ is $\E[f_\weightvec|\mathcal{D}] = \mathcal{F}[\Phi]\E[\weightvec|\mathcal{D}]$. Note that considering the predicted mean of our learned models is equivalent to taking $\weightvec$ to be the posterior mean estimate.
We compute the predicted uncertainty of our learned models as their covariances: $\Cov[u_\weightvec|\mathcal{D}] = \Phi\Cov[\weightvec|\mathcal{D}]\Phi^T$ and $\Cov[f_\weightvec|\mathcal{D}] = \mathcal{F}[\Phi]\Cov[\weightvec|\mathcal{D}]\mathcal{F}[\Phi]^T$. The predicted uncertainty can then be used to help inform the learning process (e.g., by providing a metric for determining when to stop learning or where more data may be needed). When $m=1$, the covariance is scalar-valued and we can take twice its square root (i.e., twice the posterior standard deviation) as the predicted uncertainty.
In each example, we assume a Gaussian likelihood and Gaussian prior and apply the Riccati-based methodology from Section~\ref{sec:riccatimethod} to compute the posterior mean and covariance. For demonstration purposes, we use the 4th-order Runge-Kutta (RK4) method to numerically solve the corresponding Riccati ODEs, although any other appropriate numerical method could be used instead. Details for the choice of hyperparameters in each example can be found in Appendix~\ref{sec:hyperparameters}. Code for all of the examples will be made publicly available at \url{https://github.com/ZongrenZou/HJPDE4UQSciML} after the paper is accepted.


\subsection{Solving a boundary-value ODE problem}\label{sec:4_2}

Consider the following boundary value problem:
\begin{equation}\label{eq:example_1}
    \begin{dcases}
        \kappa\frac{d^4u}{dt^4}(\tau) + \beta\frac{d^2 u}{dt^2}(\tau) + u(\tau) = f(\tau), \tau\in(0, 1), \\
        u(0) = u_0, u(T) = u_T, u^\prime(0) = u^\prime_0, u^\prime(T) = u^\prime_T,
    \end{dcases}
\end{equation}
where $\kappa=0.0001, \beta=0.01$ are known constants. 
In this example, we solve \eqref{eq:example_1} using noisy measurements of $u_0, u_T, u^\prime_0, u^\prime_T$ and noisy measurements of the source term $f$ at different times $\tau$, which we denote by $\{(\tau_i, f_i)\}_{i=1}^{N_f}$. To demonstrate the capabilities and potential computational advantages of our Riccati-based approach, we consider the following three learning scenarios:
\begin{enumerate}[label=\Alph*.]
    \item Continual learning: stream data of $f$ sequentially in $\tau$ (Section \ref{sec:example_1a}),
    \item Hyperparameter tuning: continuously tune the standard deviation of the prior  (Section \ref{sec:example_1b}),
    \item Outlier removal: remove excessively noisy data from the learned model (Appendix \ref{sec:example_1c}).
\end{enumerate}
In each scenario, we use the truncated Karhunen-Lo\`eve (KL) expansion of a Gaussian process as our linear model.
Specifically, we use the leading $n=30$ terms of the KL expansion: 
\begin{equation}
    u_\weightvec(\tau) = \sum_{k=1}^n \weight_k\sqrt{\alpha_k}\tilde{\phi}_k(\tau),
\end{equation}
where $\tilde{\phi}_k, \alpha_k$ are the eigenfunctions and eigenvalues of the kernel
\begin{equation}
    k(x_1, x_2) = \exp\left(-\frac{|x_1 - x_2|}{0.05}\right), \quad \forall x_1, x_2 \in [-10, 10].
\end{equation}
In other words, we use the basis functions $\{\phi_k(\tau) = \sqrt{\alpha_k}\tilde{\phi}_k(\tau), k=1,...,n\}$ \cite{yang2021b}. For more details, the analytic expansion can be found in \cite{ghanem2003stochastic, xiu2010numerical}. 
Unless otherwise specified, we also assume that the prior on $\weightvec$ is standard independent Gaussian, i.e., $\weightvec \sim N(0, I)$ where $I$ is the $n\times n$ identity matrix.
The exact solution is assumed to be $u(\tau) = \exp(-2\tau)\sin(15\tau)$ (but treated as unknown a priori), and we set $T=1$.
The data of $u_0, u_T$ ($u^\prime_0, u^\prime_T$, respectively) are corrupted by additive Gaussian noise with mean zero and standard deviation $0.01$ ($0.001$, respectively).

\subsubsection{Case A: continual learning}\label{sec:example_1a}

\begin{figure}[ht!]
    \centering
    \subfigure[Inferences of $u$ with UQ.]{
        \includegraphics[width=.3\textwidth]{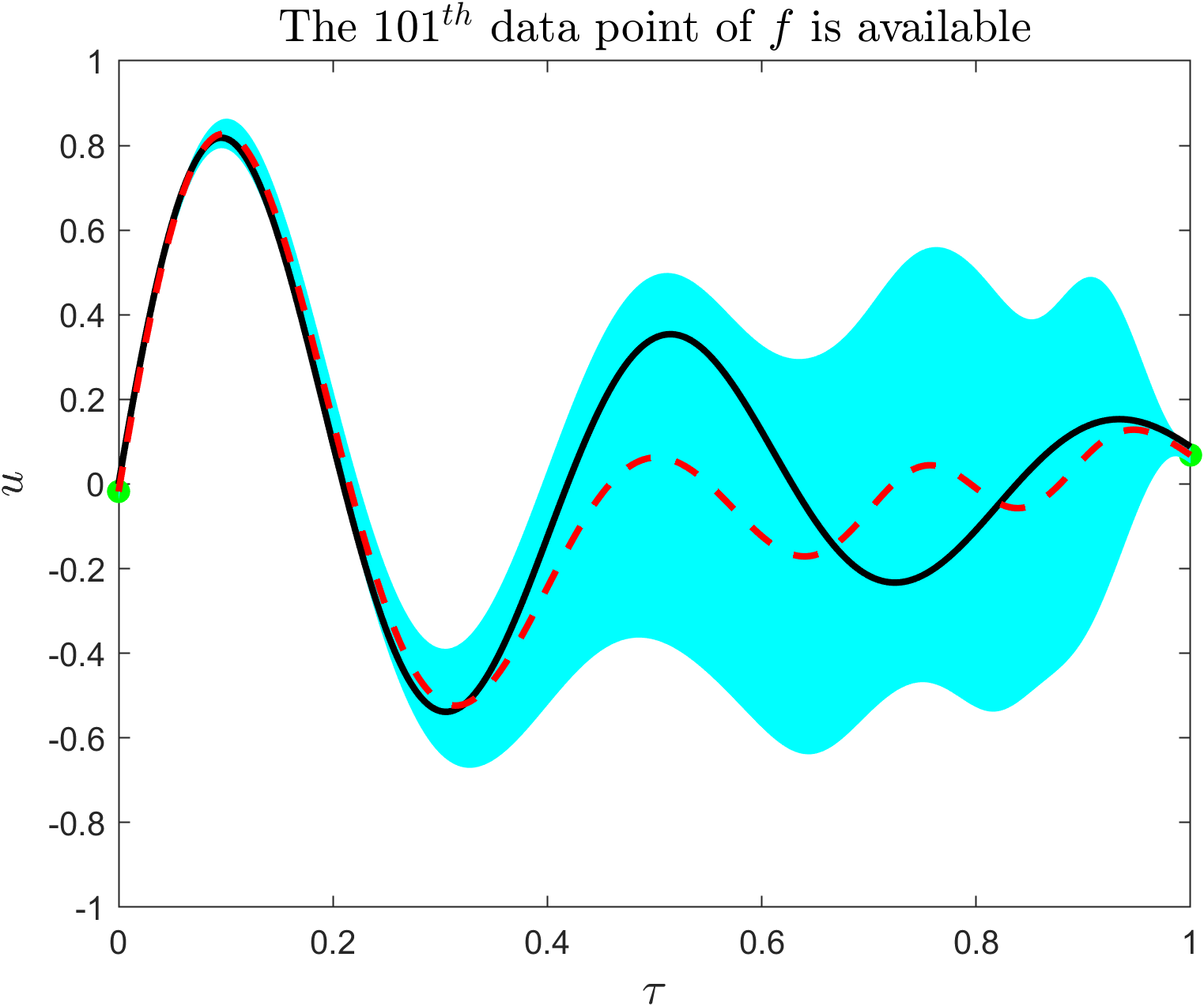}
        \includegraphics[width=.3\textwidth]{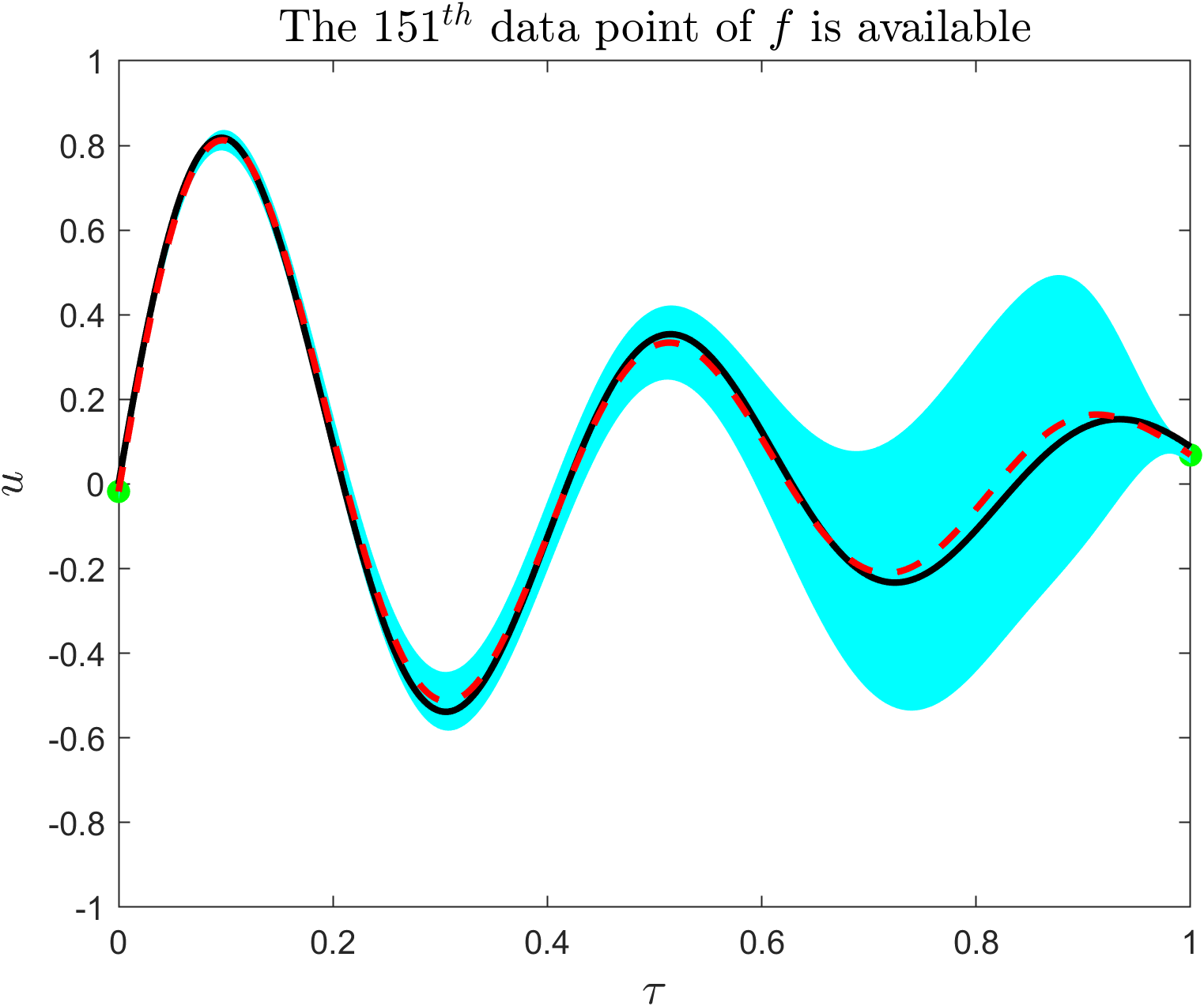}
        \includegraphics[width=.3\textwidth]{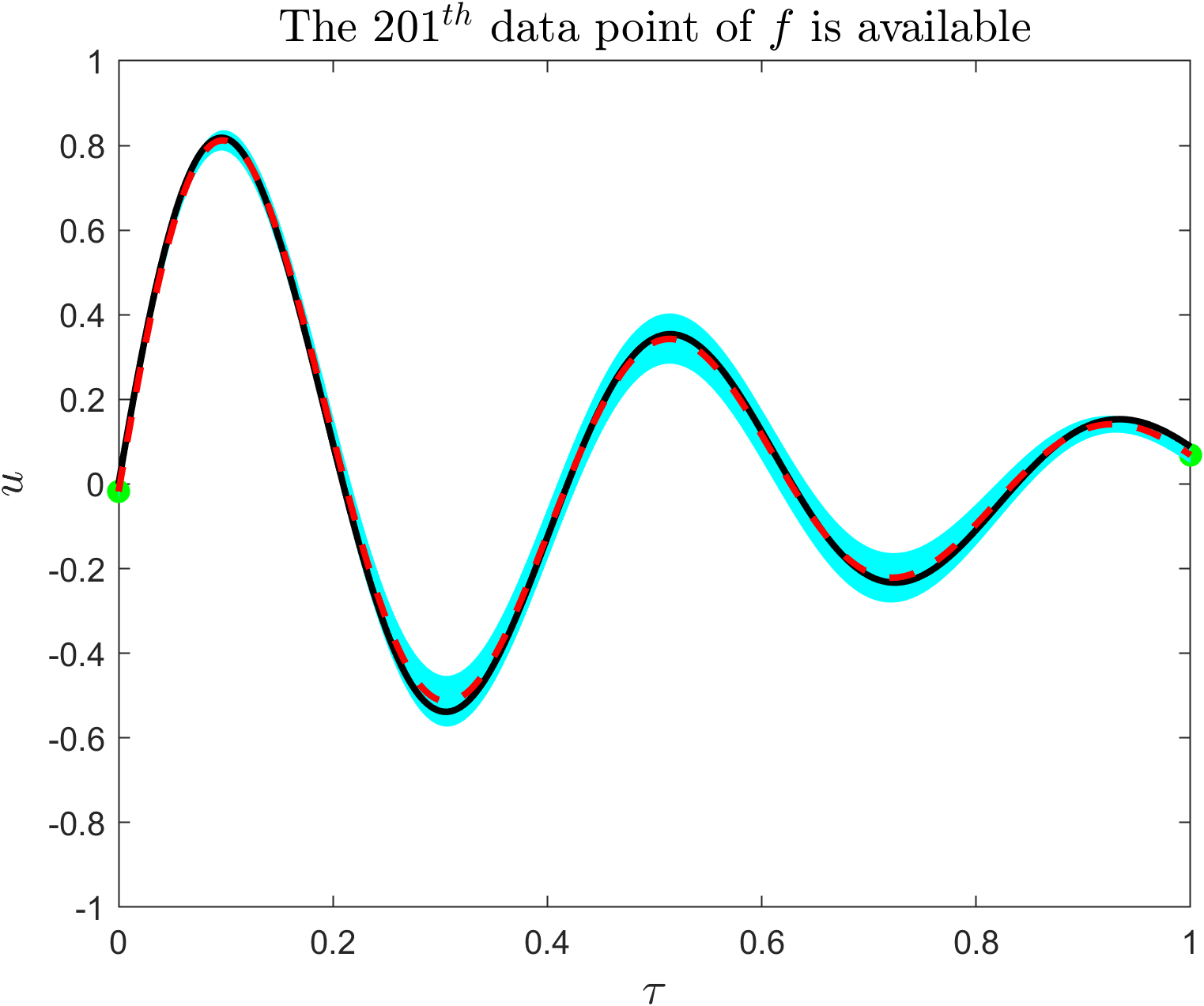}
    }
    \subfigure[Inferences/fitting of $f$ with UQ.]{
        \includegraphics[width=.3\textwidth]{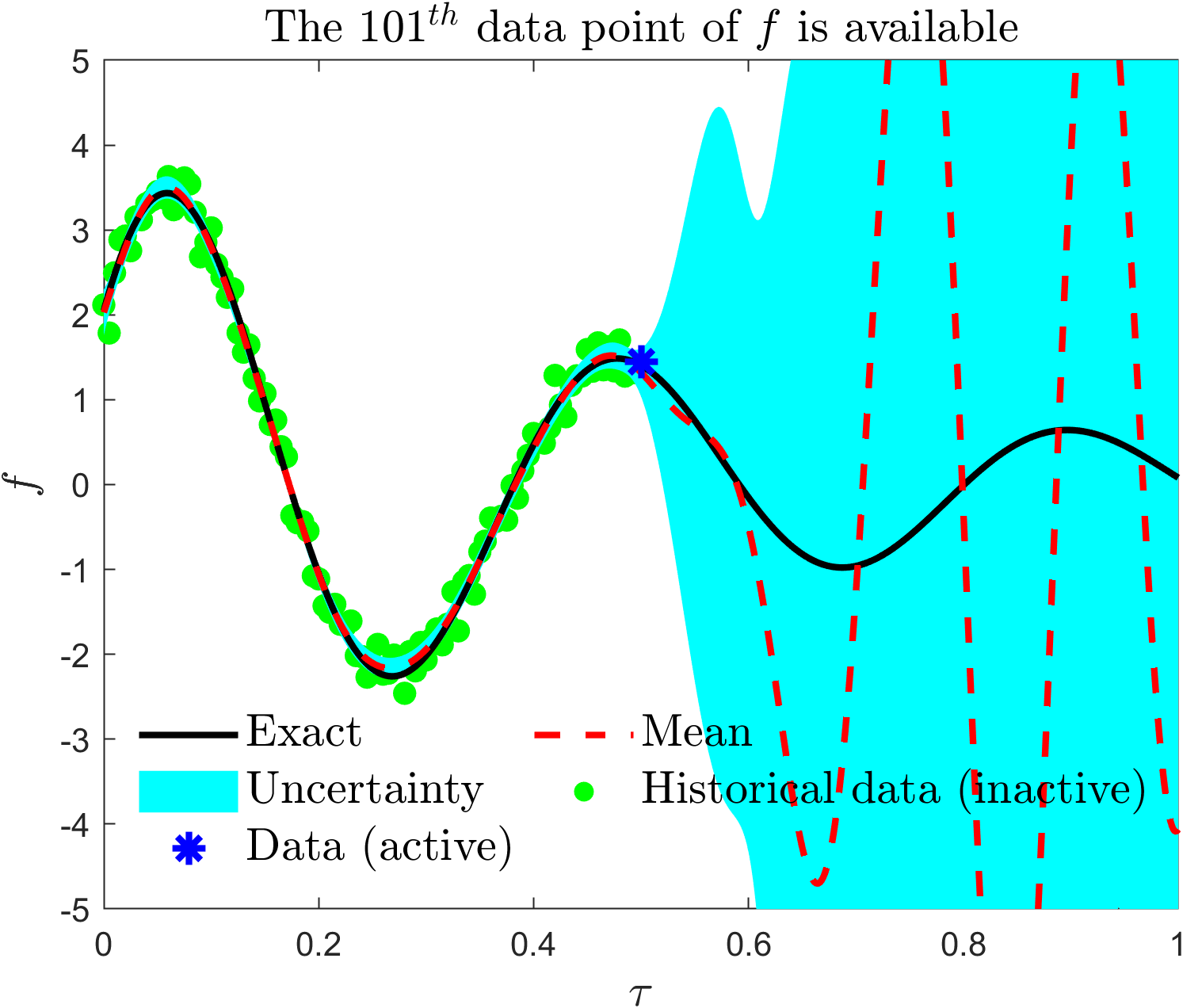}
        \includegraphics[width=.3\textwidth]{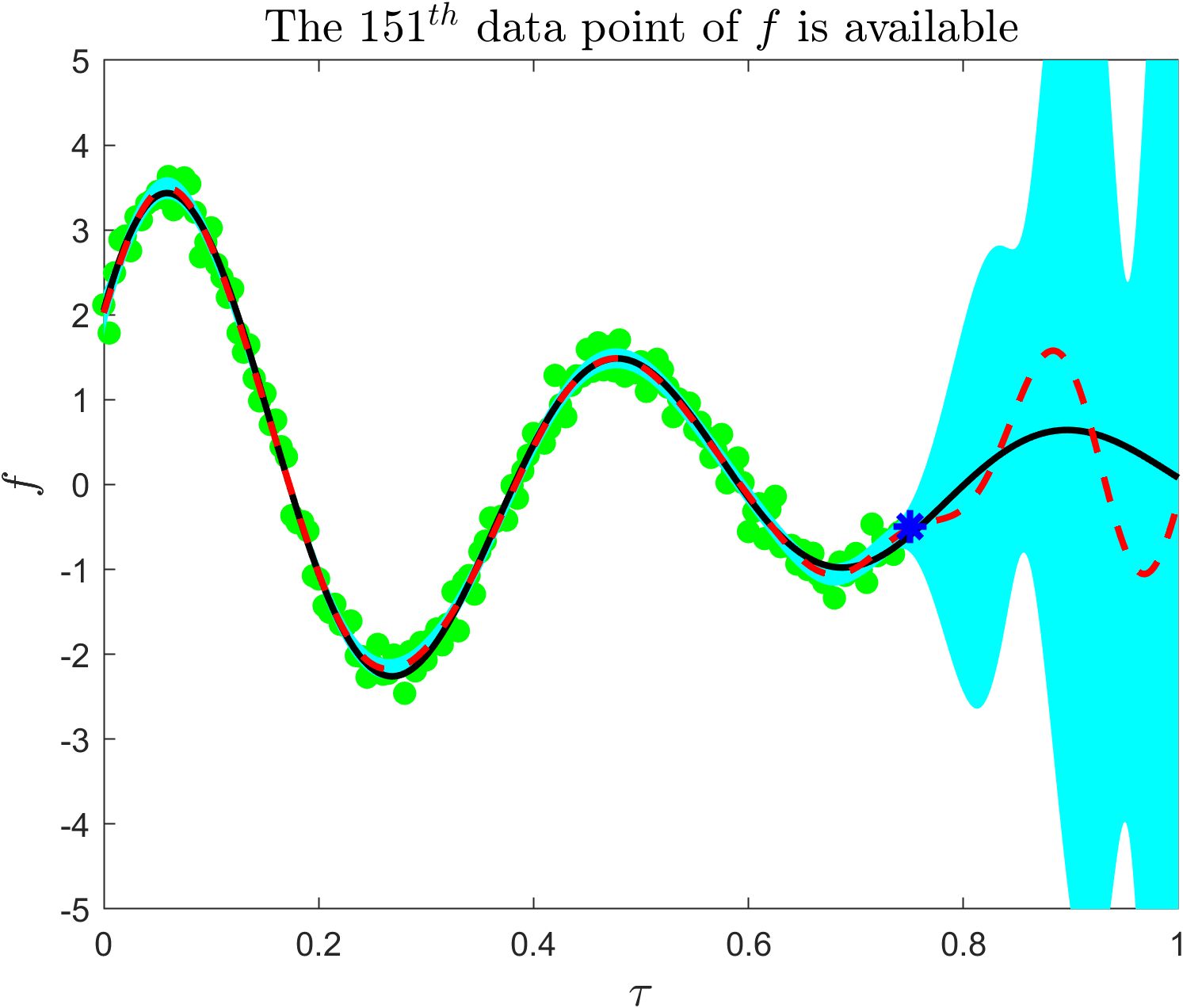}
        \includegraphics[width=.3\textwidth]{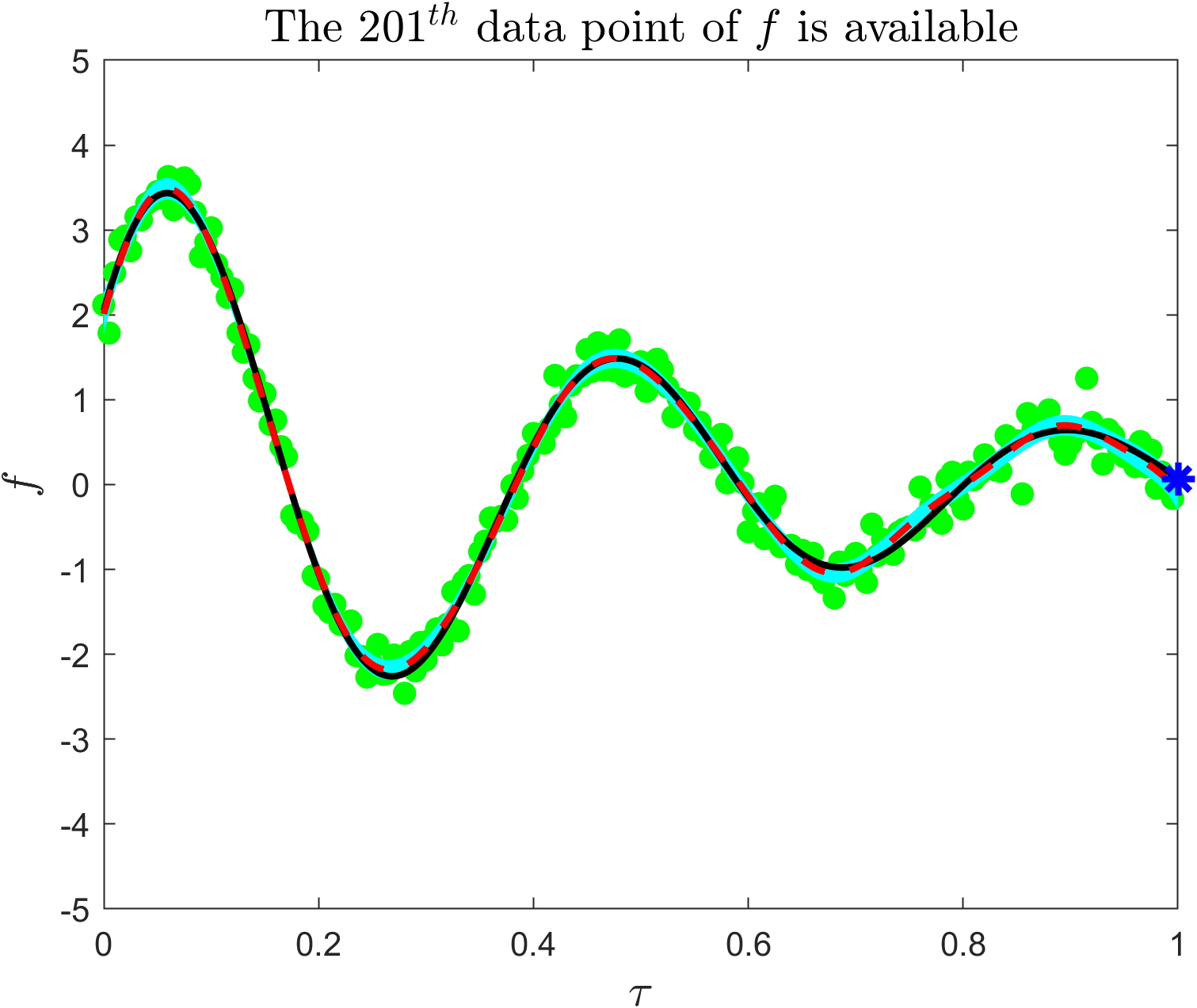}
    }
    \caption{Results of solving \eqref{eq:example_1} using continual learning and our Riccati-based approach. (a) and (b) show the predicted mean and uncertainty of $u$ and $f$, respectively, after the 101th, 151th, and 201th noisy data point of $f$ becomes available. Our Riccati-based approach naturally coincides with the continual learning framework while inherently avoiding catastrophic forgetting. Our approach allows us to incrementally update the learned models without accessing the historical data (\textcolor{green}{$\boldsymbol{\cdot}$}), while also providing a quantitative metric for our confidence in the learned models in the form of the predicted uncertainties (\textcolor{myCyan}{$\blacksquare$}). We observe that regions of low predicted uncertainty generally coincide with regions of high inference accuracy, which implies that this confidence metric is a good indicator of the reliability of the model.}
    \label{fig:example_1}
\end{figure}

\begin{table}[ht]
    \footnotesize
    \centering
    \begin{tabular}{c|c|c|c}
    \hline
    \hline
    & After the $101$th data point & After the $151$th data point & After the $201$th data point\\
    \hline 
       Error of $u$ & $41.77\%$  & $6.79\%$ & $3.88\%$ \\
       \hline
       Error of $f$ & $217.27\%$  & $24.48\%$ & $3.39\%$ \\
       \hline
       \hline
    \end{tabular}
    \caption{Relative $L_2$ errors of the predicted means of $u$ and $f$ when using continual learning and our Riccati-based approach to solve~\eqref{eq:example_1} after different amounts of noisy measurements of $f$ are incorporated into the learned models. We achieve high accuracy inferences after all the data is incorporated.} 
    \label{tab:example_1}
\end{table}

In this scenario, we learn $u$ and $f$ using continual learning \cite{parisi2019continual, van2019three} to demonstrate the potential of our Riccati-based approach for real-time inferences. Continual learning refers to learning scenarios in which data is accessed in a stream and learned models are updated incrementally as new data becomes available. In some cases, the historical data is also assumed to become inaccessible after it is incorporated into the learned model, which often leads to catastrophic forgetting (\cite{kirkpatrick2017overcoming, parisi2019continual}), i.e., the abrupt degradation in the performance of learned models on previous tasks upon training on new tasks. Here, we assume that measurements of $f$ are streamed sequentially in $\tau$ (assume that one measurement of $f$ becomes available every $\Delta\tau = \frac{T}{200}$) and that the previous data cannot be accessed again after a new measurement arrives. As such, we want to update our learned models as soon as new data becomes available. 
We also assume that the measurements of $f$ are corrupted by additive Gaussian noise with mean zero and standard deviation $0.2$, so that a Bayesian approach may be useful. We apply the Riccati-based methodology from Section~\ref{sec:3_2_1}, which is naturally amenable to data streaming. In particular, this Riccati-based approach provides computational advantages over more standard SciML and UQ techniques in continual learning settings by using only information of the newly available data to update the learned model instead of retraining on the entire dataset. These computational advantages become more significant in long-term integration problems, in which retraining on and storage of the historical data becomes considerably more expensive. Furthermore, although our Riccati-based approach does not rely on historical data, it does not suffer from catastrophic forgetting as all of the information from the previous data remains intrinsically encoded in the solution to the corresponding HJ PDE.  


Figure \ref{fig:example_1} displays the predicted means and uncertainties of  $u$ and  $f$ after the $101$th, $151$th, and $201$th noisy measurement of $f$ becomes available. We see that as more data points of $f$ are incorporated into the learned models, the predicted means of both $u$ and $f$ more closely match their exact values and the predicted uncertainties shrink. Moreover, we see that regions of high predicted uncertainty generally coincide with low inference accuracy, which shows that our confidence in the learned models is correlated with their reliability.
We also observe that the predicted uncertainty of $u$ does not develop in the same way as that of $f$. While the predicted uncertainty of $f$ is generally uniformly small on $(0, \tau^*)$ and large on $[\tau^*, T)$, where $\tau^*$ denotes the $\tau$-coordinate of the currently available data point, $u$ can often still have relatively large predicted uncertainty on $(0, \tau^*)$ since we do not learn $u$ directly. Note that $u$ always has low predicted uncertainty on the boundaries as the boundary data were incorporated into the learned models at the beginning of the training.
Table \ref{tab:example_1} presents the relative $L_2$ errors of the predicted means, which show that we achieve high inference accuracy after all the data has been incorporated. 
We compute the $L_2$ errors using trapezoidal rule with a uniform grid of size 1001 over the whole domain $[0, T]$.

\subsubsection{Case B: tuning the standard deviation of the prior}\label{sec:example_1b}

\begin{figure}[ht!]
    \centering
    \subfigure[Validation error.]{\includegraphics[width=.3\textwidth]{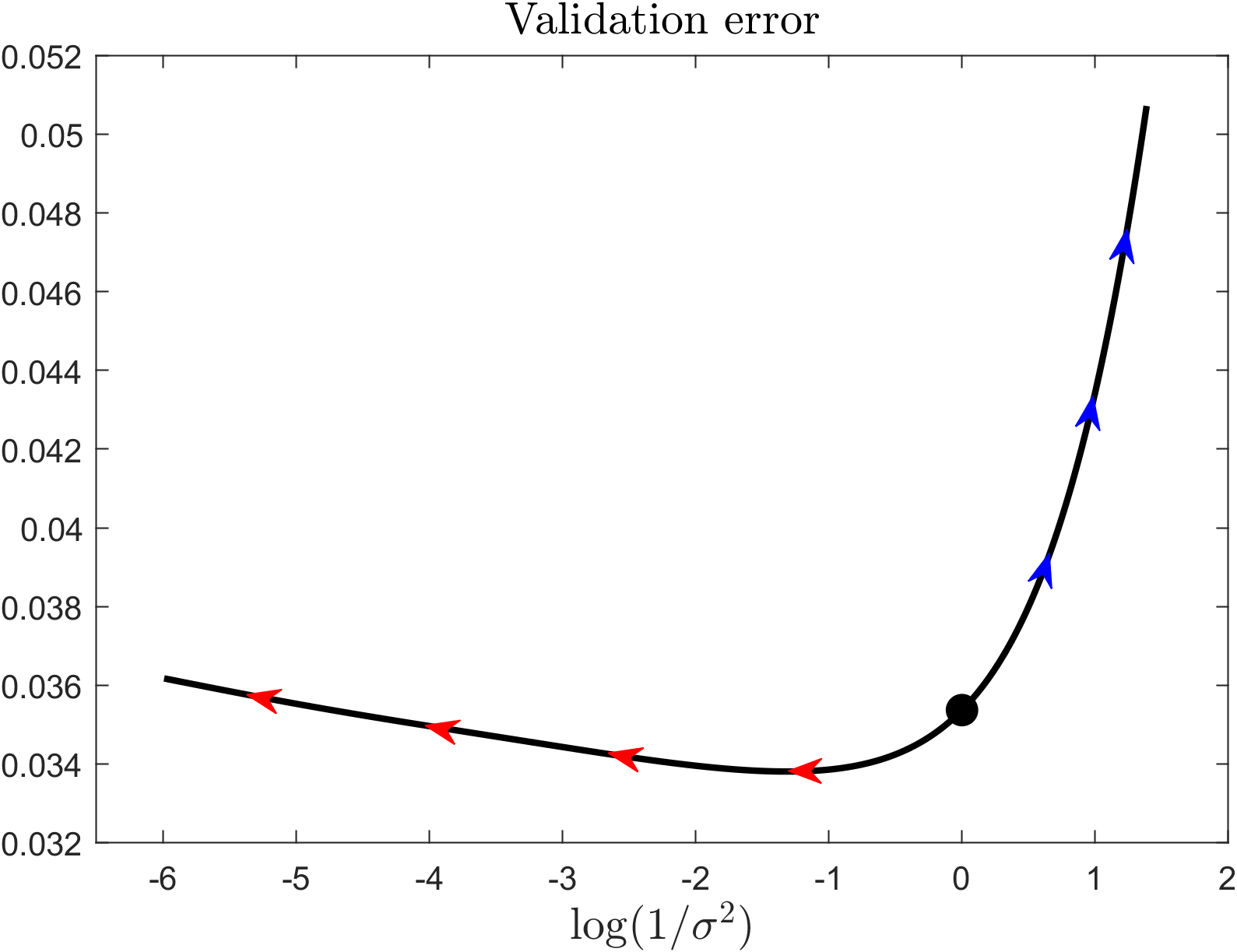}}
    \subfigure[Predicted mean of $u$ at different $\tau$ as the function of $\sigma$.]{
    \includegraphics[width=.2\textwidth]{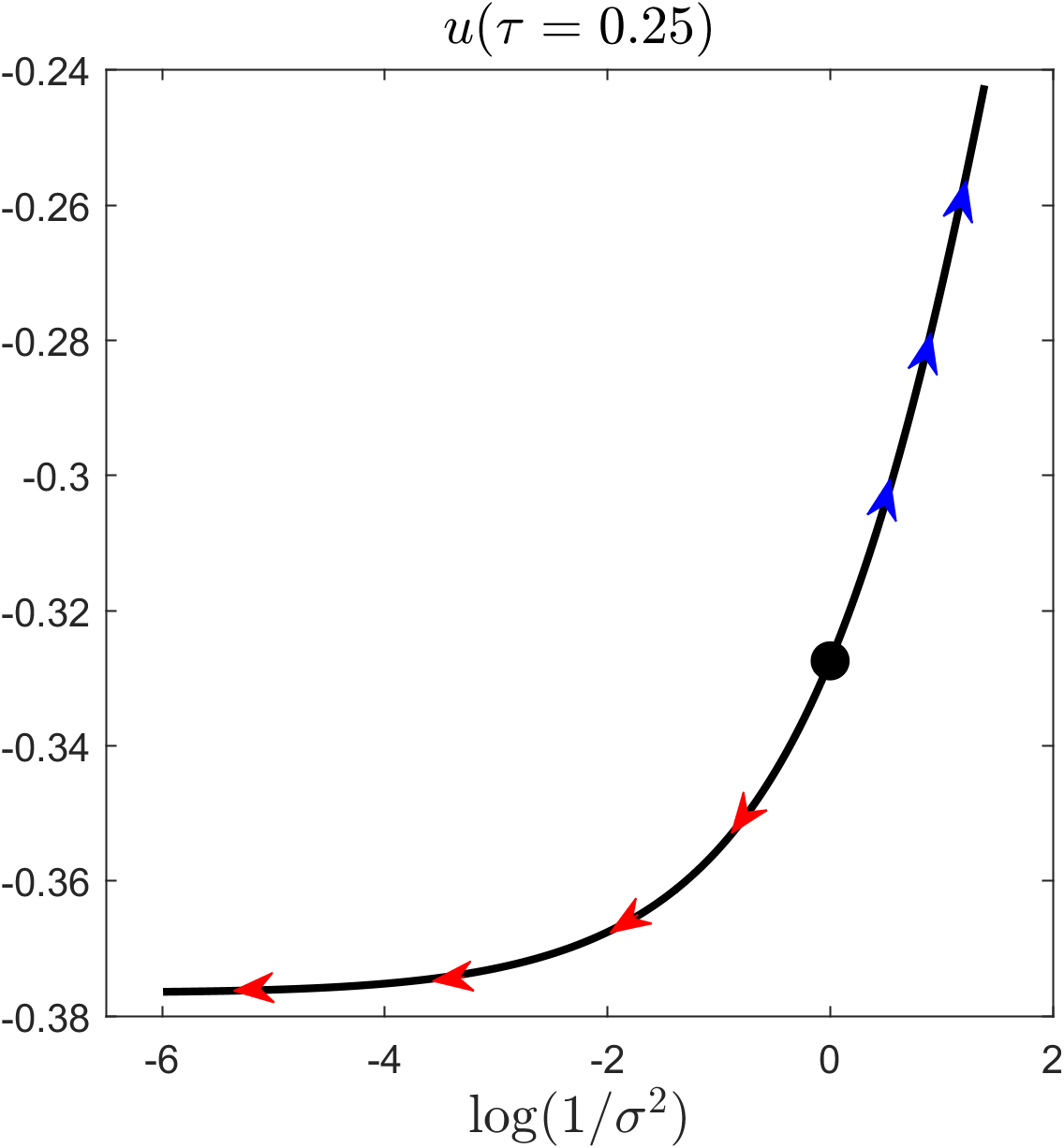}
    \includegraphics[width=.2\textwidth]{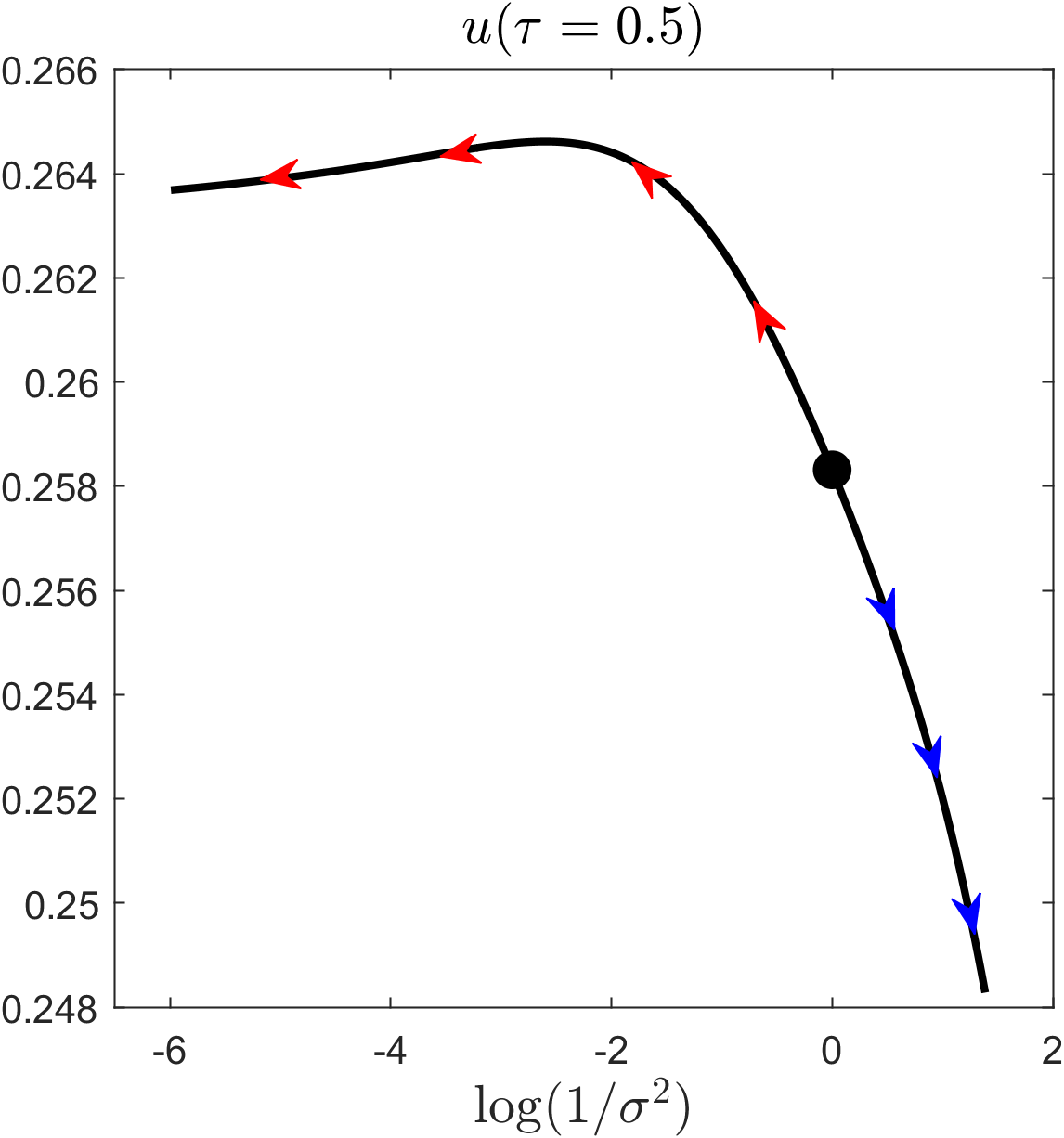}
    \includegraphics[width=.205\textwidth]{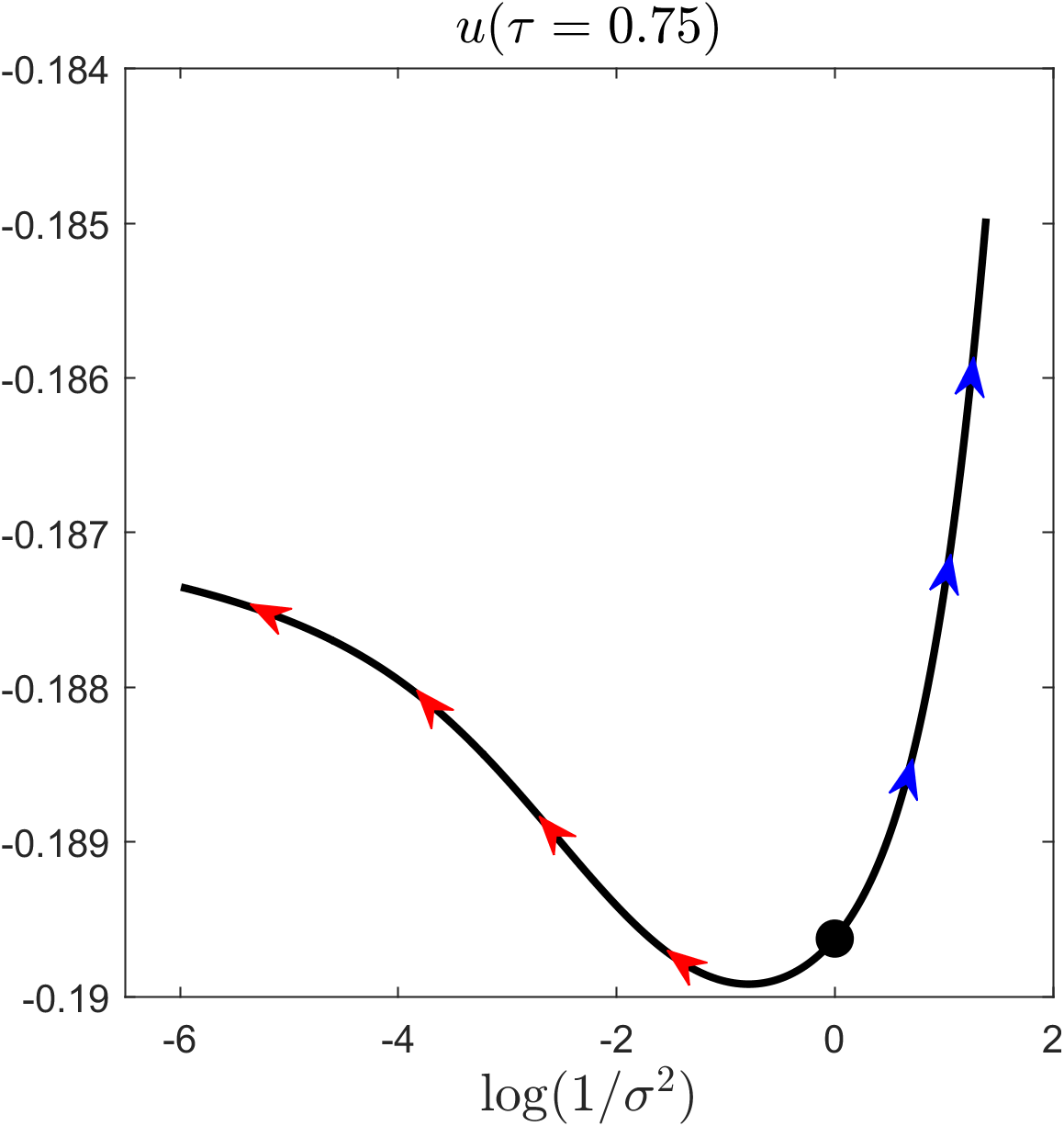}
    }
    \caption{Results of tuning the standard deviation $\sigma$  of the prior using our Riccati-based approach to solve \eqref{eq:example_1}. 
    (a) shows how the validation error changes as $\sigma$ is tuned. (b) shows the flow with respect to $\sigma$ of the predicted mean of $u$ at different times ($\tau=0.25, 0.5, 0.75$). The black dot represents the value using the original choice of $\sigma=1$, and the arrows represent the direction of the flow of solutions as we increase $\sigma$ from 1 ({\color{red}\textbf{red}}; i.e., we evolve the corresponding Riccati ODE backward) or decrease $\sigma$ from 1 ({\color{blue}\textbf{blue}}; i.e., we evolve the corresponding Riccati ODE forward). In contrast to conventional ML approaches, our Riccati-based approach allows the hyperparameter to be tuned continuously without retraining on or access to the training data.}
    \label{fig:example_1_1}
\end{figure}

In the previous case, the prior distribution was set to be independent standard Gaussian, i.e., $\weightvec\sim N(0, \sigma^2 I)$, which is the usual choice for KL expansions of Gaussian processes \cite{williams2006gaussian}.
However, this choice may not be appropriate for every problem, and tuning the hyperparameters of the prior may yield better results.
In this section, we demonstrate how the Riccati-based approach from Section \ref{sec:3_3} can be used to tune the standard deviation $\sigma$ of the prior continuously. 
In this case, the training data consists of the same measurements of $u$ and $u^\prime$ used in Case A and 41 measurements of $f$ sampled equidistantly on $[0, T]$. 
The validation data consists of 10 measurements of $f$ sampled randomly on $[0, T]$ following a uniform distribution. Both the training and validation data are corrupted by additive Gaussian noise with mean zero and standard deviation $0.2$.

Figure \ref{fig:example_1_1} shows how the validation error and predicted mean of $u$ develop as $\sigma$ is tuned.  Figure \ref{fig:example_1_1}(b) shows that the predicted mean of $u$ varies as $\sigma$ changes, and Figure \ref{fig:example_1_1}(a) shows that the lowest validation error is not achieved with the original choice of $\sigma=1$, which highlights the importance of hyperparameter tuning.
Recall that tuning $\sigma$ corresponds to evolving a Riccati ODE in time. Hence, tuning $\sigma$ can be done continuously and results in a continuous flow of solutions, where each solution corresponds to a different choice of $\sigma$. This flow of solutions is obtained numerically via the intermediary steps of RK4 and is represented by the arrows in Figure~\ref{fig:example_1_1}. In Figure \ref{fig:example_1_1}(a), tuning $\sigma$ results in a continuous flow of validation errors, and in Figure \ref{fig:example_1_1}(b), it results in a continuous flow of inferences of $u$. Our Riccati-based approach allows hyperparameter values to be explored flexibly and continuously and, in contrast to more standard ML techniques, does not require retraining on or access to the training dataset during the tuning process. 

 


\subsection{Solving the 1D steady-state advection-diffusion equation}\label{sec:example2}

Consider the following steady-state advection-diffusion equation with Dirichlet boundary conditions:
\begin{equation}\label{eq:example_2}
    \begin{dcases}
        D\frac{\partial^2 u}{\partial x^2}(x) + \kappa \frac{\partial u}{\partial x}(x) = f(x), x\in (0, 1),\\
        u(0) = u(1) = 0,
    \end{dcases}
\end{equation}
where $D=0.001$ and $\kappa=1$. We solve~\eqref{eq:example_2} using noisy measurements of $f$ at different $x$, which we denote by $\{(x_i, f_i)\}_{i=1}^{N}$. To demonstrate the versatility of our Riccati-based approach in handling different learning settings, we consider the following two cases: 
\begin{enumerate}[label=\Alph*.]
    \item The data of $f$ are corrupted by large-scale noise but are cheap to obtain (Section \ref{sec:4_2_1}),
    \item The data of $f$ are corrupted by small-scale noise but are expensive to obtain (Section \ref{sec:4_2_2}).
\end{enumerate}
Case A uses a large-scale problem with a huge amount of data to show that we can achieve high accuracy despite a large amount of noise, while Case B uses active learning \cite{ren2021survey} to search for where new measurements are needed.
In both cases, we use the truncated KL expansion of the Brownian bridge to model $u$. 
Specifically, we use the leading $n=50$ terms of the KL expansion \cite{ghanem2003stochastic}:
\begin{equation}
 u_\weightvec(x) = \sum_{k=1}^n \weight_k \frac{\sqrt{2} \sin(k\pi x)}{k\pi}.
\end{equation}
Note that this choice of basis functions automatically enforces the Dirichlet boundary conditions.
We take the prior to be independent standard Gaussian, i.e., $\weightvec\sim N(0,I)$.

\subsubsection{Case A: large-scale data with high noise level}\label{sec:4_2_1}

\begin{figure}[ht!]
    \centering
    \subfigure[Inferences of $u$ with UQ.]{
        \includegraphics[width=.3\textwidth]{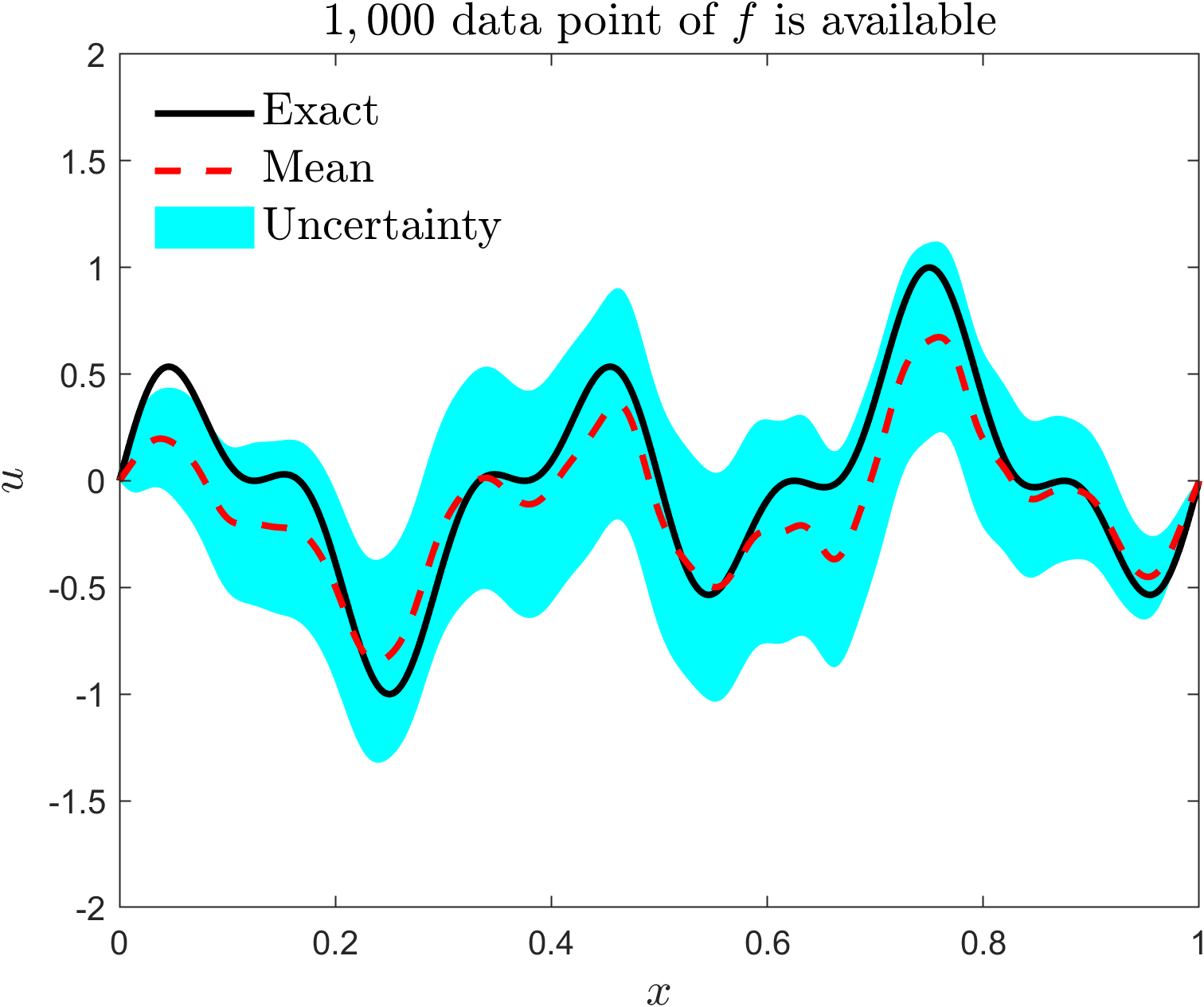}
        \includegraphics[width=.3\textwidth]{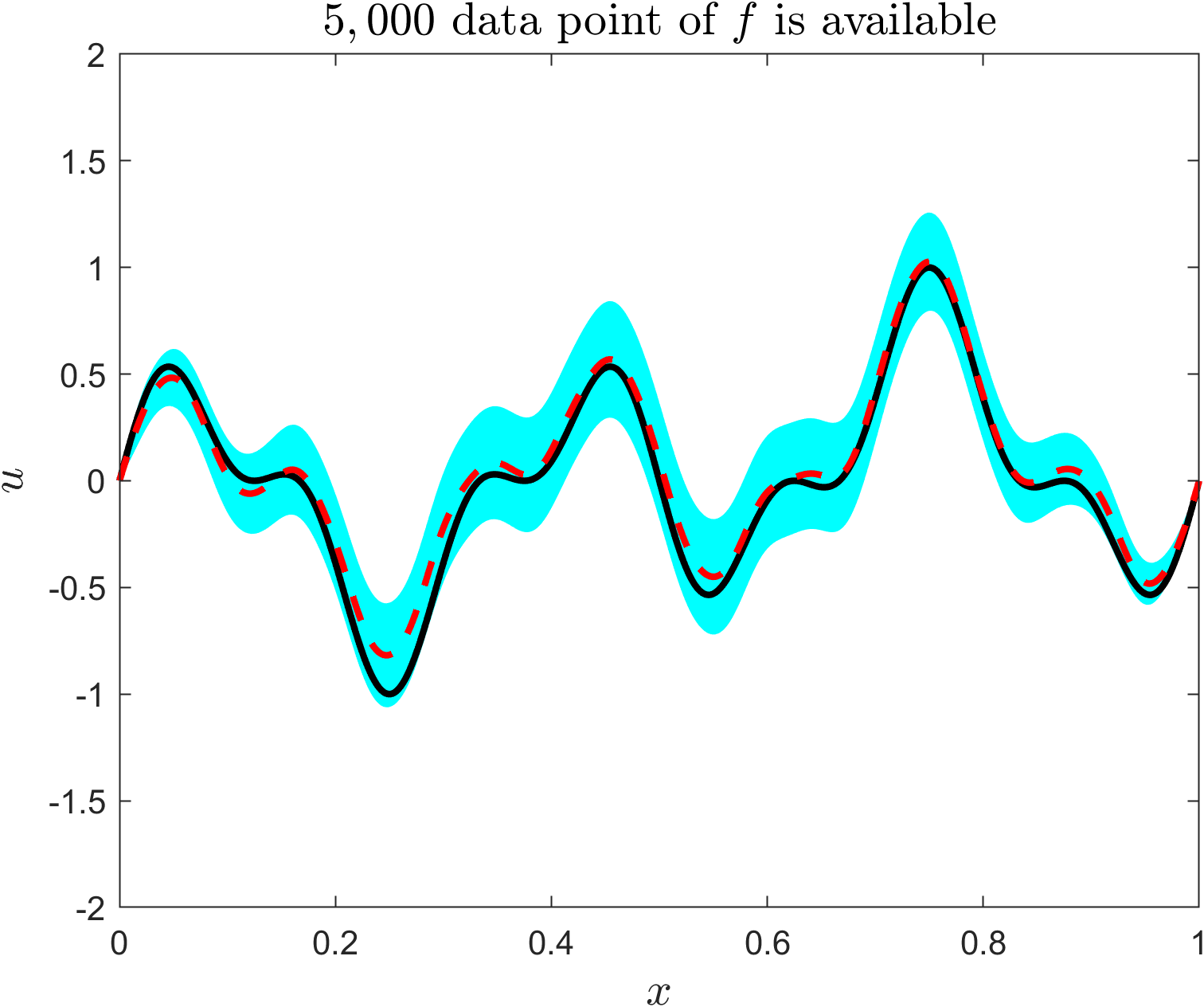}
        \includegraphics[width=.3\textwidth]{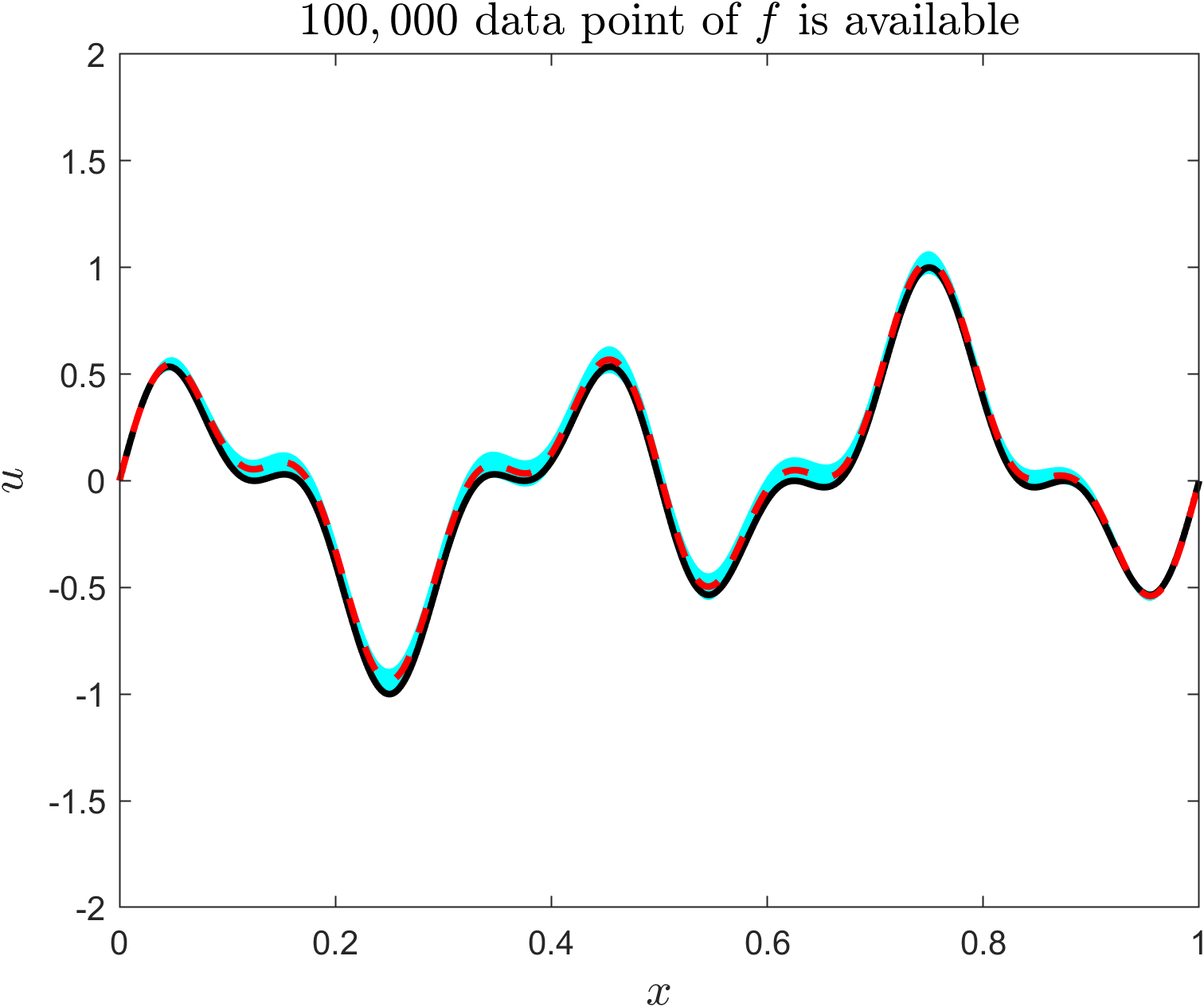}
    }
    \subfigure[Inferences/fitting of $f$.]{
        \includegraphics[width=.3\textwidth]{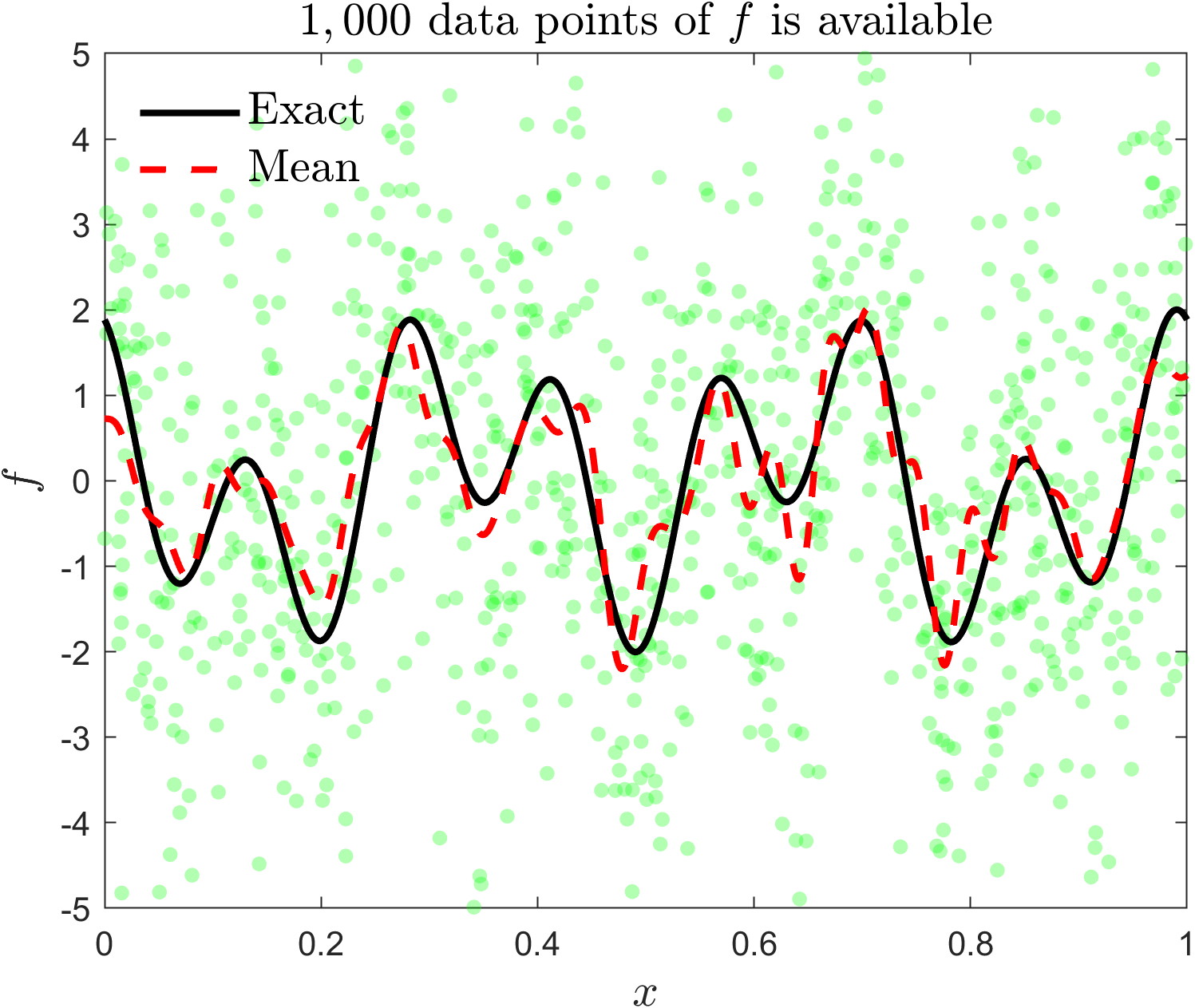}
        \includegraphics[width=.3\textwidth]{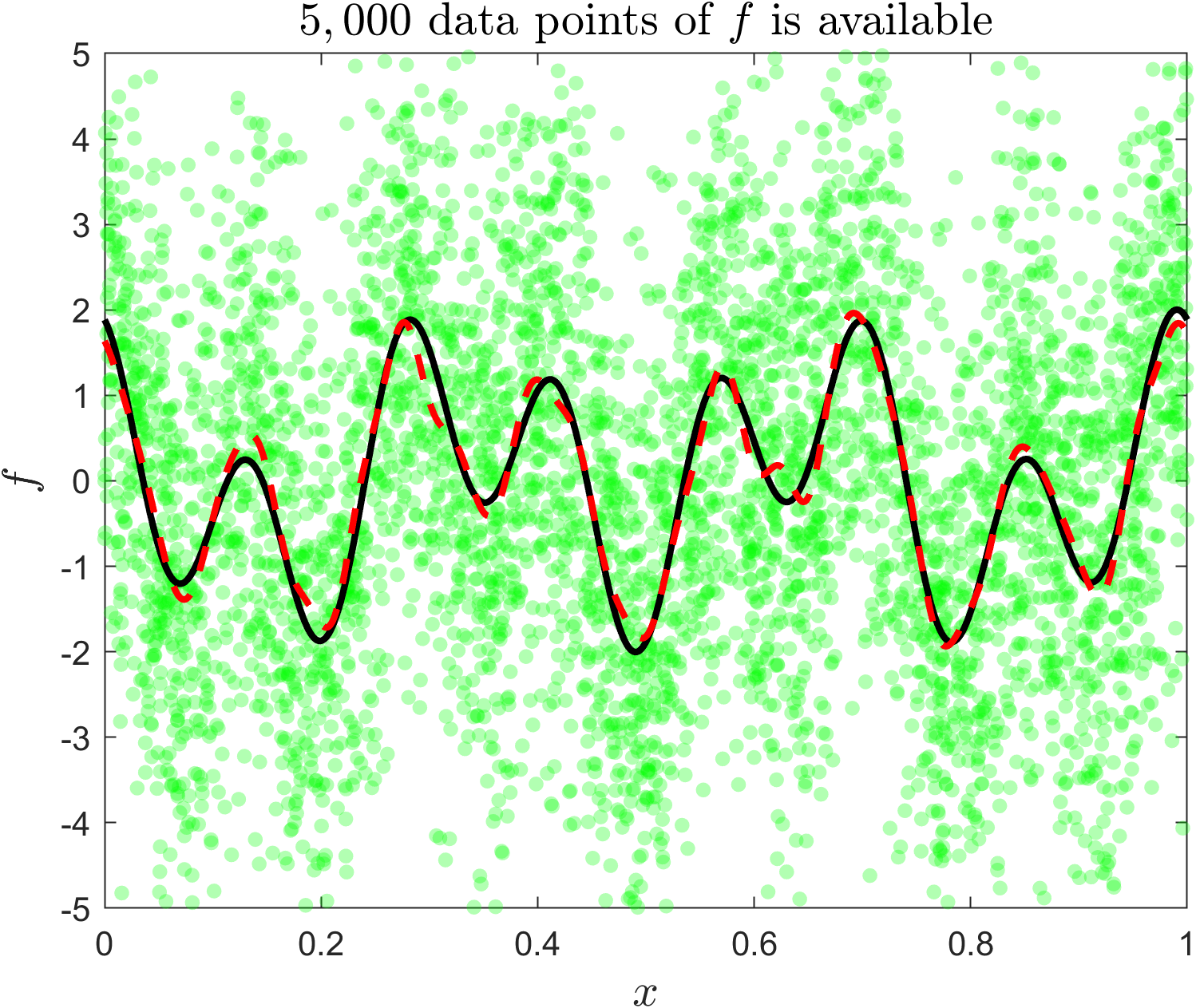}
        \includegraphics[width=.3\textwidth]{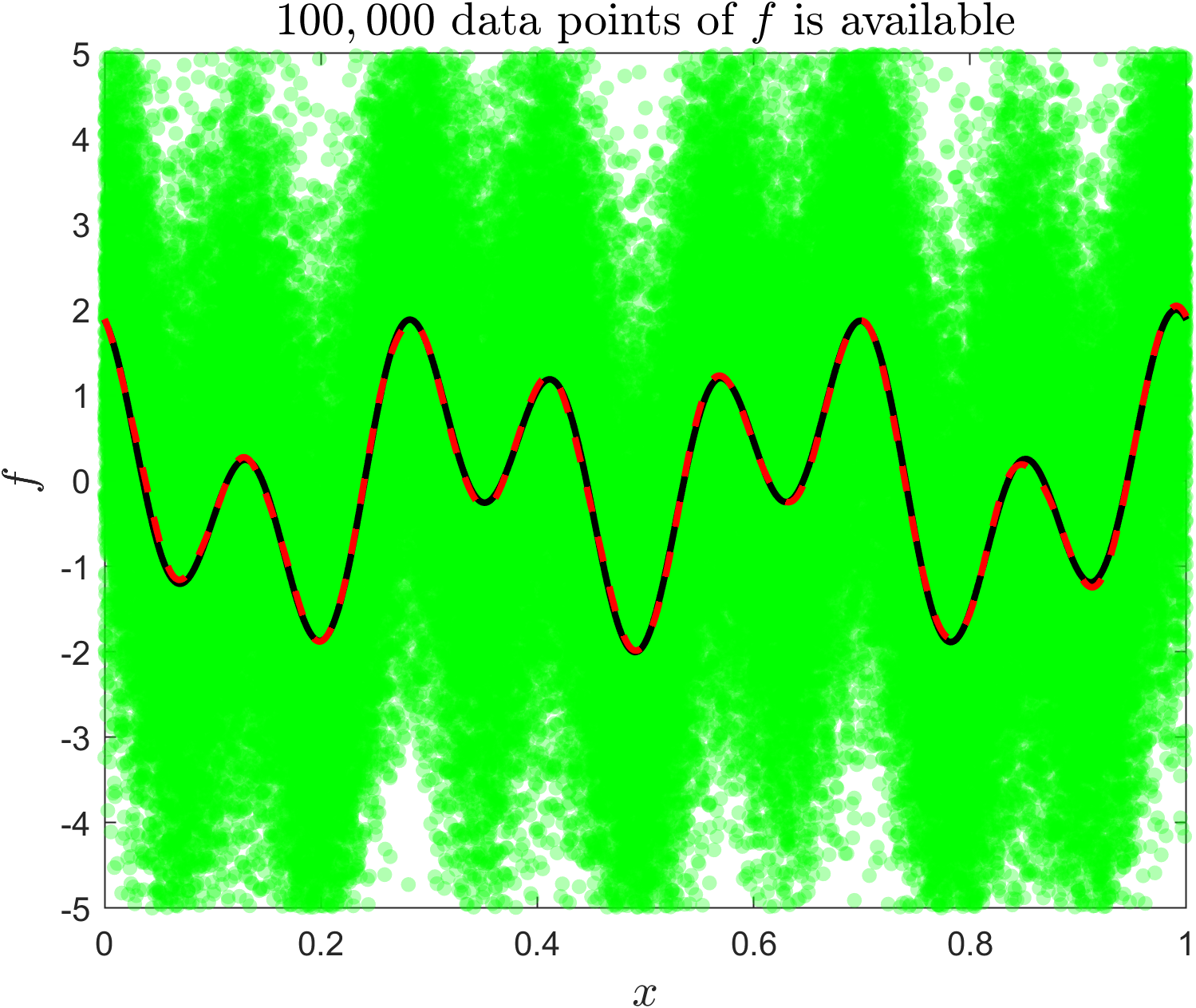}
    }
    \caption{Results of solving \eqref{eq:example_2} using large-scale data with high noise level and our Riccati-based approach. (a) shows the corresponding predicted mean and uncertainty (\textcolor{myCyan}{$\blacksquare$}) of $u$. (b) shows the inference/fitting of $f$ using $1,000$, $5,000$, and $100,000$ noisy data points. Note that we do not show the predicted uncertainty of $f$ for clarity of presentation.  Our Riccati-based approach updates the learned models incrementally using information from one data point (\textcolor{green}{$\boldsymbol{\cdot}$}) at a time, which provides significant computational and memory savings in  big data regimes.}
    \label{fig:example_2_1}
\end{figure}

\begin{table}[ht]
    \footnotesize
    \centering
    \begin{tabular}{c|c|c|c}
    \hline
    \hline
    & $1,000$ data points & $5,000$ data points & $100,000$ data points\\
    \hline
       Error of $u$ & $44.44\%$ & $17.19\%$ & $9.49\%$ \\
       \hline
       Error of $f$ & $41.89\%$ & $16.72\%$ & $3.81\%$ \\
       \hline
       \hline
    \end{tabular}
    \caption{Relative $L_2$ errors of the predicted means of $u$ and $f$ using different amounts of noisy measurements of $f$ to solve \eqref{eq:example_2}. Our Riccati-based approach updates the learned models without requiring retraining when new data becomes available or storing and processing large datasets in their entirety. }
    \label{tab:example_2}
\end{table}

In this case, we demonstrate the computational efficiency of our Riccati-based approach on a large-scale problem. Specifically, we assume that measurements of $f$ are corrupted by a large amount of noise but are cheap to acquire. Thus, to obtain accurate predictions and compensate for the high noise level, we need a lot of data. Working with such a large dataset in its entirety would be extremely expensive in terms of both computations and memory. Instead, we apply our Riccati-based approach from Section~\ref{sec:3_2_1} to continuously learn our model without have to train on or store the entire dataset.
In this case, the measurements of $f$ are sampled uniformly randomly on $[0, 1]$ and corrupted by additive Gaussian noise with mean zero and standard deviation $2$ (e.g., see in Figure \ref{fig:example_2_1}(b)). We keep sampling $f$ until a satisfactory result is obtained (i.e., until the predicted uncertainty is sufficiently low).

Figure \ref{fig:example_2_1} shows inferences of $u$ and $f$ after $1,000$, $5,000$, and $100,000$ data points of $f$ are incorporated into the learned models.
Note that the predicted uncertainty at the boundary points is always zero because our choice of basis functions automatically enforces the boundary conditions. 
Due to the high noise level, the inferences of $u$ and $f$ are inaccurate when only 1,000 data points are used. Additionally, the predicted uncertainty of $u$ is relatively large, which indicates that the learned models are low confidence. 
As we incorporate more measurements of \(f\) into the learning process, the inferences become more accurate and the predicted uncertainty of $u$ decreases.
Once the predicted uncertainty is small enough, we determine that we are confident in our learned model and can stop collecting data. For example, in Figure \ref{fig:example_2_1}(a), we see that the predicted uncertainty of $u$ is significantly and sufficiently reduced after using $100,000$ data points, and we stop learning. In Table~\ref{tab:example_2}, we observe that the relative $L_2$ errors of our inferences are also significantly reduced, confirming that our confidence in the model is a reasonable indicator of its accuracy. 
The $L_2$ errors are computed using trapezoidal rule with a uniform grid of size 1001 over the domain $[0, 1]$. 
We again note that by using the Riccati-based methodology from Section~\ref{sec:3_2_1}, we avoid having to retrain every time we acquire new data points and never have to work with all 100,000 data points at once. Instead, our Riccati-based approach only requires information about one data point at a time, which provides significant computational and memory advantages in these big data regimes.

\subsubsection{Case B: expensive data with low noise level}\label{sec:4_2_2}

\begin{figure}[ht]
    \centering
    \subfigure[Inferences/fitting of $f$ with UQ.]{
        \includegraphics[width=.22\textwidth]{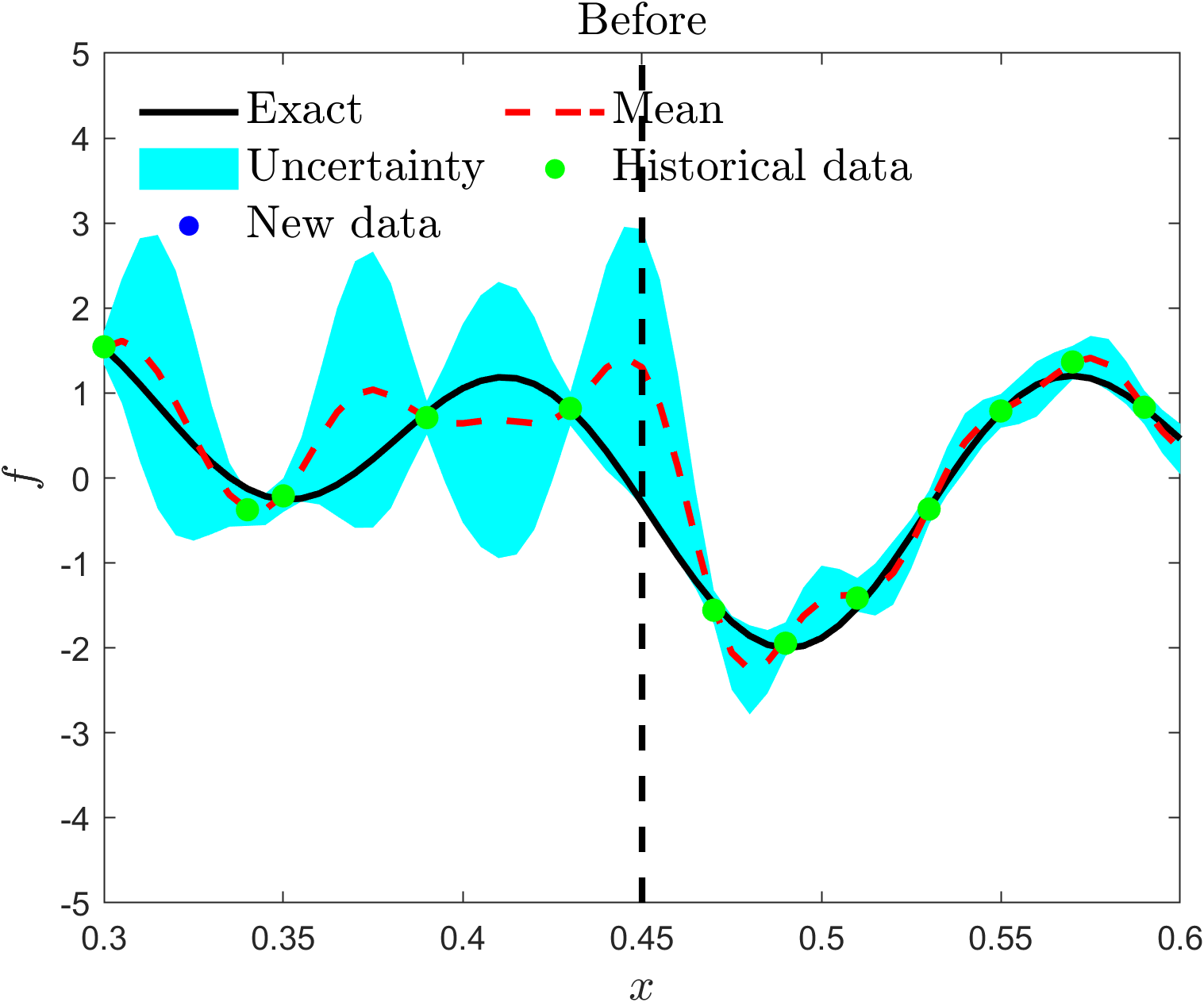}
        \includegraphics[width=.22\textwidth]{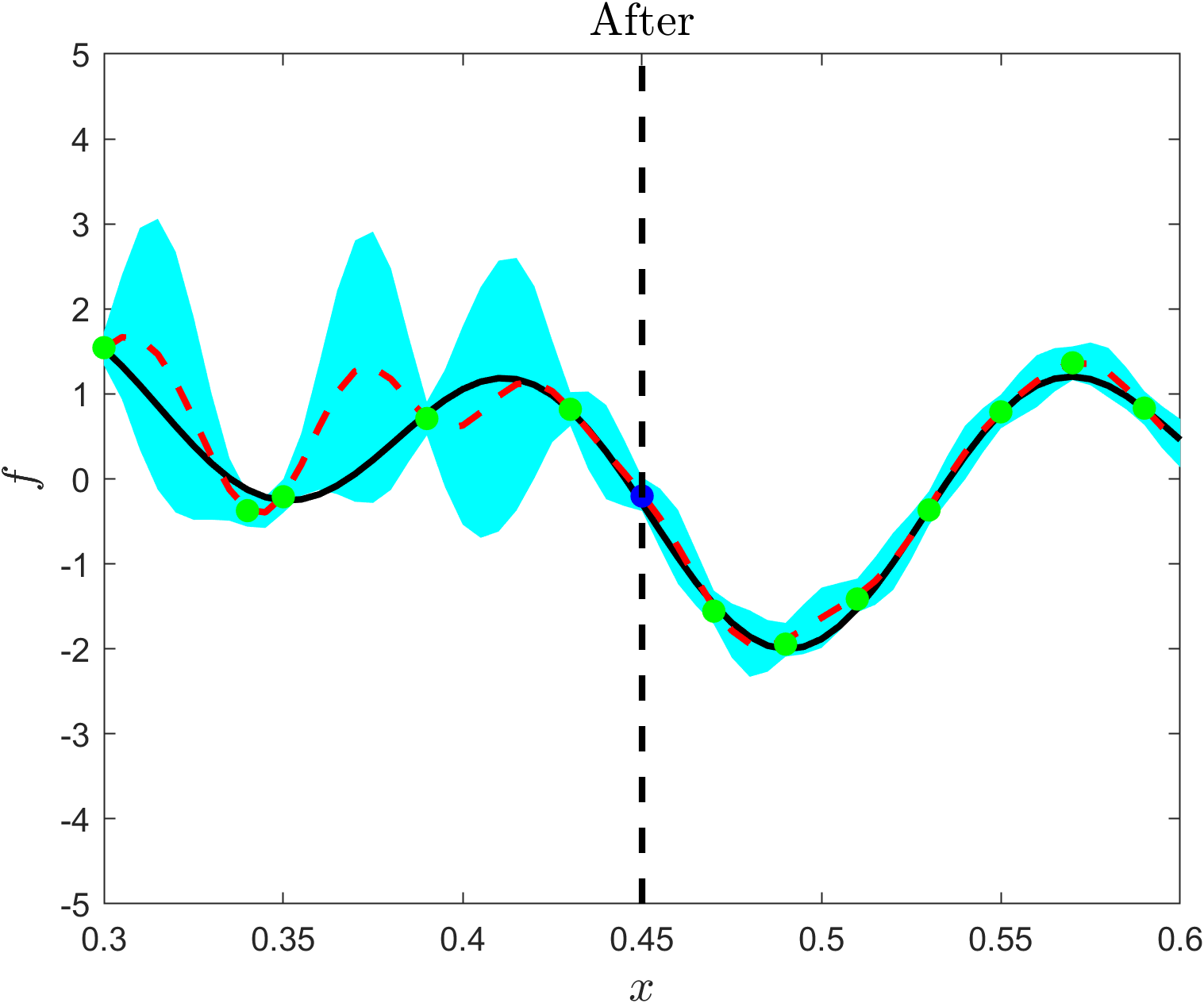}
    }
    \subfigure[Inferences of $u$ with UQ.]{
        \includegraphics[width=.22\textwidth]{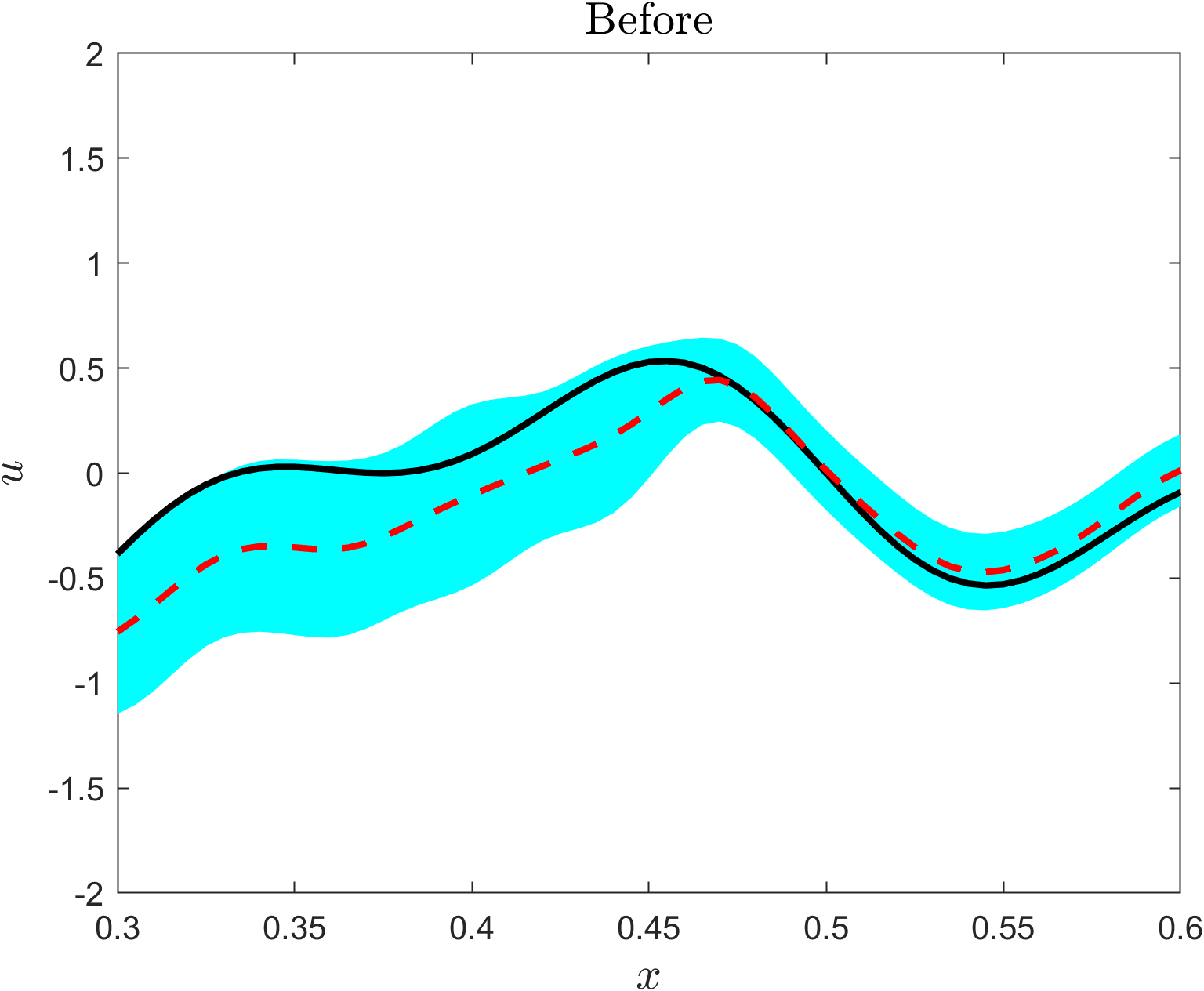}
        \includegraphics[width=.22\textwidth]{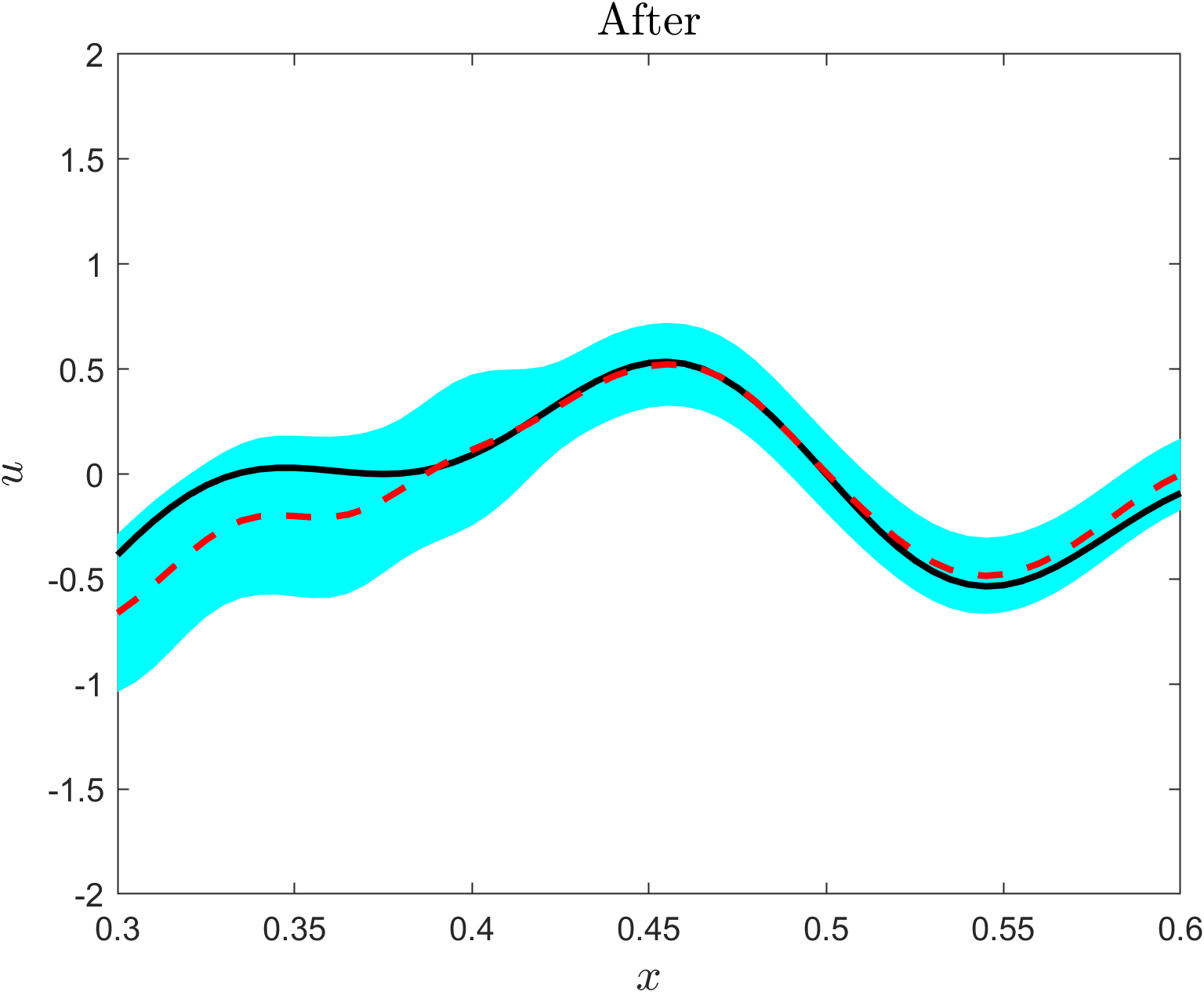}
    }
    \caption{Intermediate result for adding the $46$th data point when using active learning and our Riccati-based approach to solve \eqref{eq:example_2}. 
    The \textbf{left} side of (a) and (b) shows the inferences of $f$ and $u$, respectively, before the placement of the new sensor and incorporation of the 46th data point. The \textbf{right} side of (a) and (b) shows the updates to $f$ and $u$ after these actions.
    The vertical black dashed line represents the location where the $46$th sensor will be placed. Note that we zoom in on the region around the new sensor instead of showing the full domain for clarity of comparison. 
    Following an active learning framework, we repeat the following steps: compute the predicted mean and uncertainty of $u$ and $f$, place a new sensor at the location of the largest predicted uncertainty (\textcolor{myCyan}{$\blacksquare$}) of $f$, and obtain a new measurement of $f$ (\textcolor{blue}{\textbf{*}}) at the new sensor location. Our Riccati-based approach naturally complements this framework as it can efficiently perform repeated updates of the predicted mean and uncertainty.}
    \label{fig:example_2_3}
\end{figure}

In this case, we demonstrate the versatility of our Riccati-based methodology by showing how it naturally complements an active learning approach. 
Active learning refers to the ML paradigm where the training data are  sampled dynamically until a desired result is achieved (e.g., low generalization error) \cite{settles2009active, ren2021survey, mackay1992information}. In this section, we focus on uncertainty-based active learning \cite{mackay1992information, gramacy2009adaptive, pickering2022discovering}, in which we choose the new data point to be sampled from the location where the predicted uncertainty is largest.
Assume that we have access to high quality data with low-level noise (here, we use additive Gaussian noise with mean zero and standard deviation 0.1) but that the data are expensive to collect. For example, consider the scenario where high precision but expensive sensors are deployed to measure the value of $f$ at particular locations $x$. As a result, we have limited access to data and must carefully select where sensors should be placed in order to obtain reliable inferences. 
Assume the potential locations for sensors form a $101$-point uniform grid on $[0, 1]$.
We use the predicted uncertainty of $f$ to determine where the next sensor should be placed. Note that although we could have instead used the predicted uncertainty of $u$ to choose the sensor location, using the predicted uncertainty of $f$ is more consistent with the fact that we collect measurements of $f$. 
Following the active learning framework \cite{ren2021survey, learningdataset, pickering2022discovering}, we start with an empty dataset $\mathcal{D}$ and index $i=0$ and then do the following:
\begin{enumerate}
    \item Compute the posterior $p(\weightvec|\mathcal{D})$ based on the current dataset $\mathcal{D}$ as well as the predicted mean and uncertainty of $u$ and $f$ at all potential sensor locations. If the predicted uncertainty is sufficiently low, then terminate the procedure.
    \item Identify the location where the predicted uncertainty of $f$ is highest out of all remaining potential locations. Label this location as $x_i$, and place a sensor there to measure $f$ (denote these measurements by $f_i$).
    \item Update the current dataset $\mathcal{D}$ to $\mathcal{D} \cup \{(x_i, f_i)\}$ and the index $i$ to $i+1$. Go back to step 1.
\end{enumerate}
To perform steps 1 and 2, we employ the Riccati-based approach from Section~\ref{sec:3_2_1}, in which the models and their predicted uncertainty are updated incrementally as new data becomes available. Note that the Riccati-based approach performs these updates without using historical data. 

Figure~\ref{fig:example_2_3} displays a zoom in of the intermediate result when adding the $46$th measurement of $f$. 
The vertical black dashed line indicates where the predicted uncertainty of $f$ is the highest (i.e., where the new sensor is placed). 
While the addition of the sensor significantly reduces the predicted uncertainty of $f$ around that point, it only mildly reduces the predicted uncertainty of $u$ since we do not learn $u$ directly.
Figure~\ref{fig:example_2_4} in Appendix~\ref{sec:example_2c} displays  intermediate results when adding the $47$th sensor, which more dramatically reduces the predicted uncertainty of $u$. 
These results demonstrate how our Riccati-based approach can be seamlessly integrated into an active learning framework by allowing for continual updates of the models and uncertainty metrics that can be leveraged to interactively improve the learning process.


\subsection{Solving the 2D Helmholtz equation}\label{sec:example_3}

In this example, we solve the 2D Helmholtz equation with Dirichlet boundary conditions:
\begin{equation}\label{eq:example_3}
    \begin{dcases}
        (\kappa^2 - \Delta)u(x,y) = f(x,y), &x, y\in[0, 2\pi],\\
        u(x, 0) = u(x, 2\pi) = u(0, y) = u(2\pi, y) = 0, & x, y\in[0, 2\pi],
    \end{dcases}
\end{equation}
where $\kappa^2=1$ is the Helmholtz constant and $\Delta$ is the Laplacian operator.
We consider the scenario where we have large-scale, noisy data of $f$ but assume that computational limitations prevent us from being able to process or store all of the data at once. We use this example to show how our Riccati-based approach naturally overcomes these computational limitations.
Specifically, we decompose the domain into multiple smaller subdomains and update our learned model on each piece one-by-one using continual learning and the Riccati-based methodology from Section~\ref{sec:3_2}. Note that since the Riccati-based approach is invariant to the order of the data points, it does not require a specific method for decomposing the domain. To illustrate the flexibility of our approach, we decompose the domain $[0, 2\pi]^2$ uniformly into $7\times 7$ equal subdomains and consider the following two patterns for traversing the subdomains:
\begin{enumerate}[label=\Alph*.]
    \item A sequential pattern (Section~\ref{sec:seq_domain}; e.g., see Figure~\ref{fig:example_3_1}(a)),
    \item A multi-level order of traversal (Section~\ref{sec:multigrid_domain}; e.g., see Figure~\ref{fig:example_3_3}).
\end{enumerate}
In both cases, we employ a linear model $u_\weightvec(x,y) = \sum_{k=1}^n\weight_k\phi_k(x,y)$, where 
    $\{(x,y)\mapsto\phi_k(x,y)\}_{k=1}^n = 
    \left\{(x,y)\mapsto\frac{\sqrt{2L}\sin(\frac{j\pi x}{L})}{j\pi}\frac{\sqrt{2L}\sin(\frac{k\pi y}{L})}{k\pi}\right\}_{j, k=1}^{\sqrt{n}},$
$n=225$, and $L=2\pi$. Note that these basis functions automatically enforce the boundary conditions. We learn the model using measurements of $f$ that are corrupted by additive Gaussian noise with mean 0 and standard deviation $0.5$. We define $f$ by
\begin{equation}
    f(x, y) = \sin(6x)\sin(4y) - 0.8 \sin(5x)\sin(7y),
\end{equation}
and the solution to \eqref{eq:example_3} can be analytically derived accordingly. To compute the accuracy of learned models, we evaluate our inferences of $u$ and $f$ on a uniform grid of size $450\times450$ over the domain. To learn $\weightvec$, we take noisy measurements of $f$ at each of the $200,704$ interior points of that grid. Hence, each subdomain contains $4,096$ measurements of $f$.

\subsubsection{Case A: sequential domain decomposition}\label{sec:seq_domain}

\begin{figure}[ht]
    \centering
    \subfigure[Domain.]{
    \includegraphics[width=.25\textwidth]{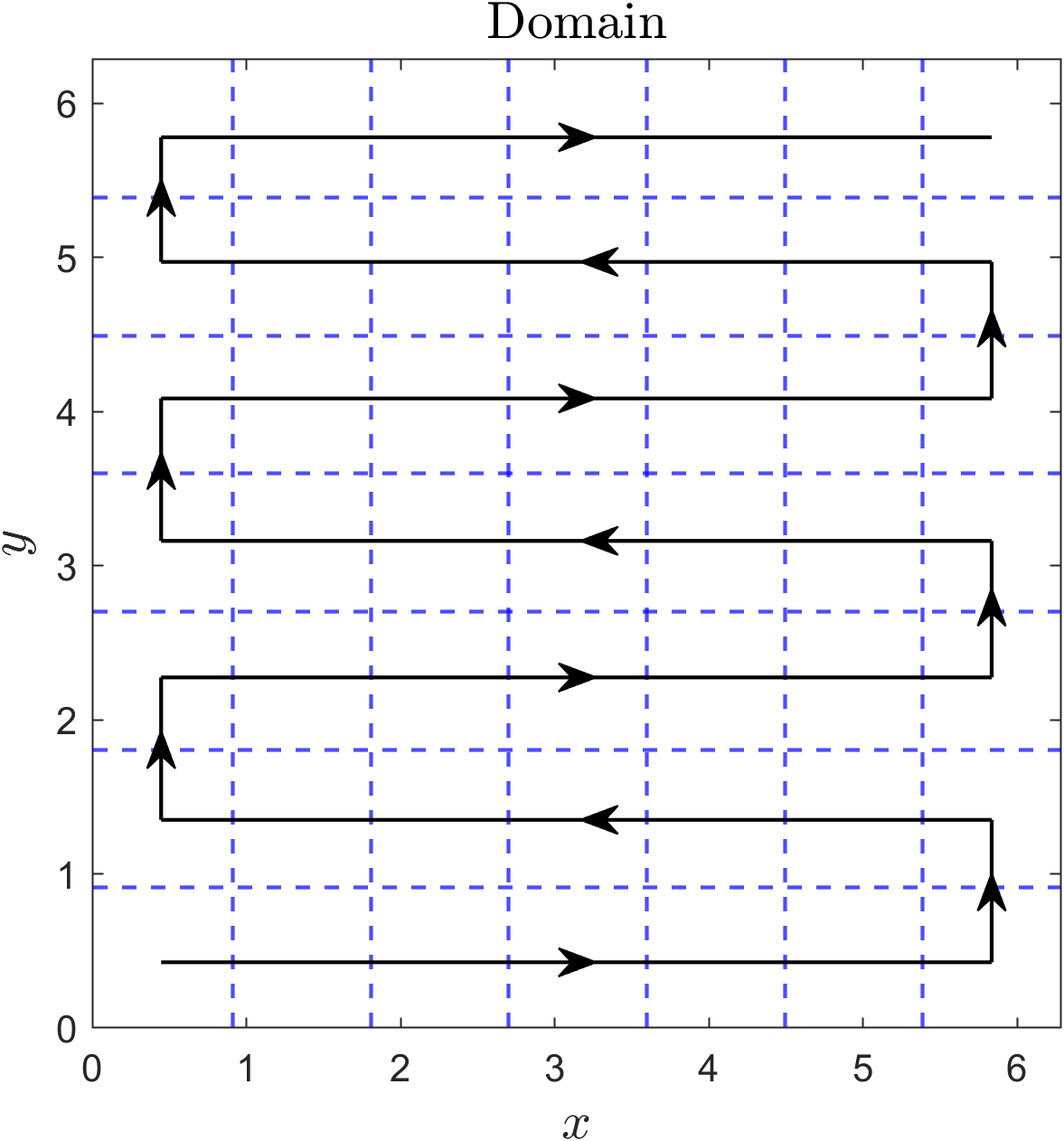}
    }
    \subfigure[Absolute error of the inference.]{
    \includegraphics[width=.25\textwidth]{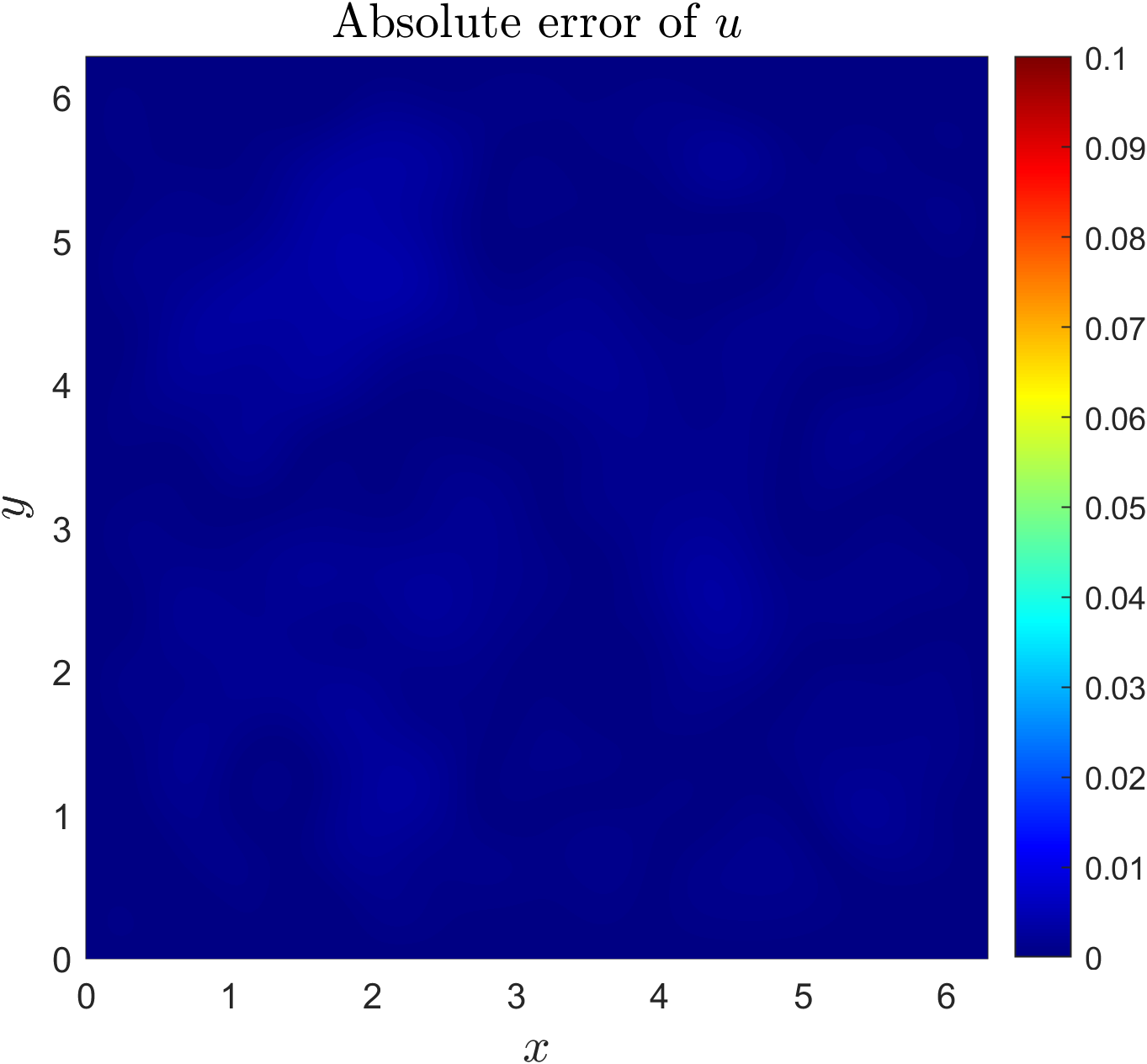}
    \includegraphics[width=.25\textwidth]{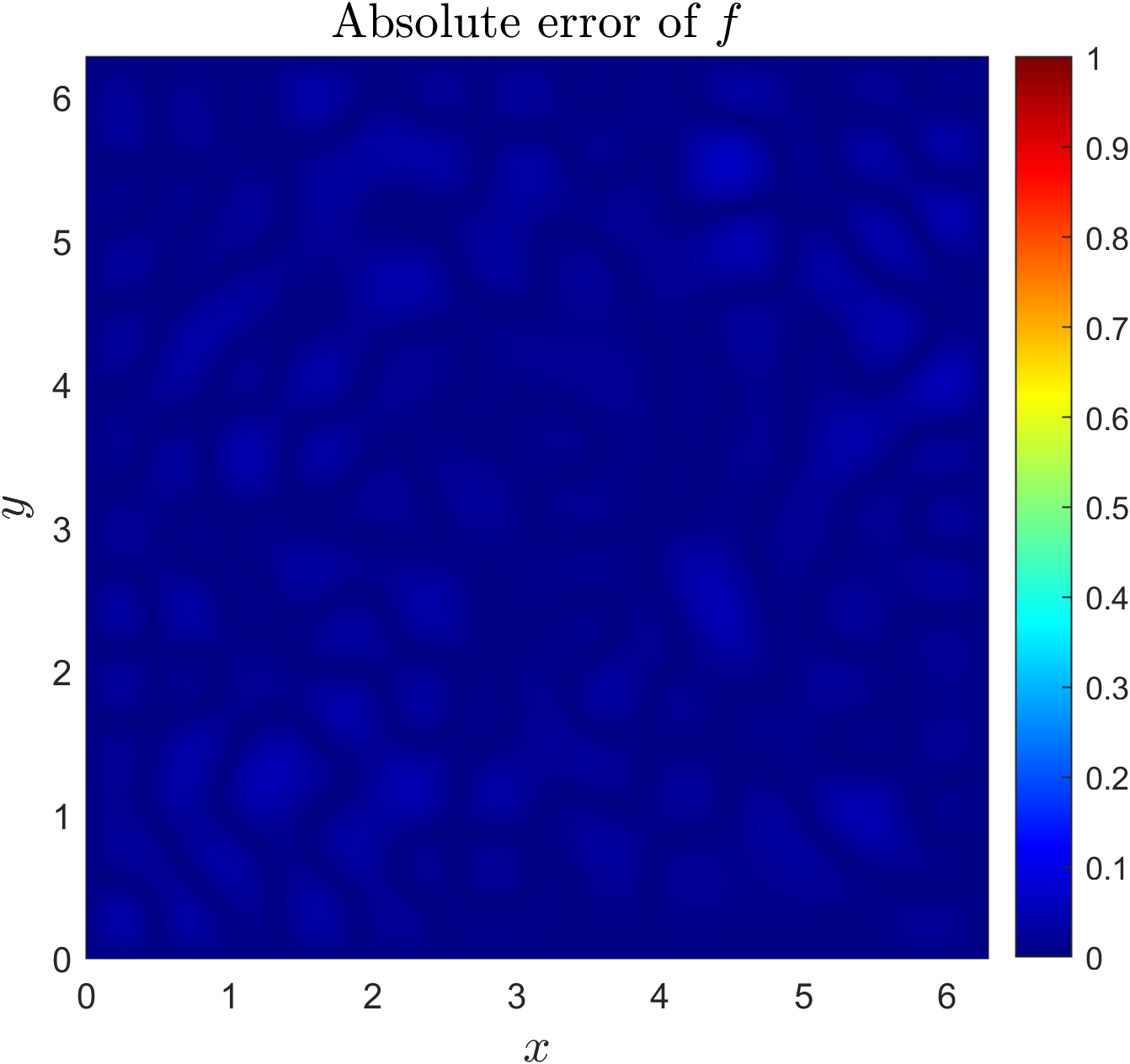}
    }
    \subfigure[Slices of the inference of $u$ with UQ.]{
        \includegraphics[scale=.3]{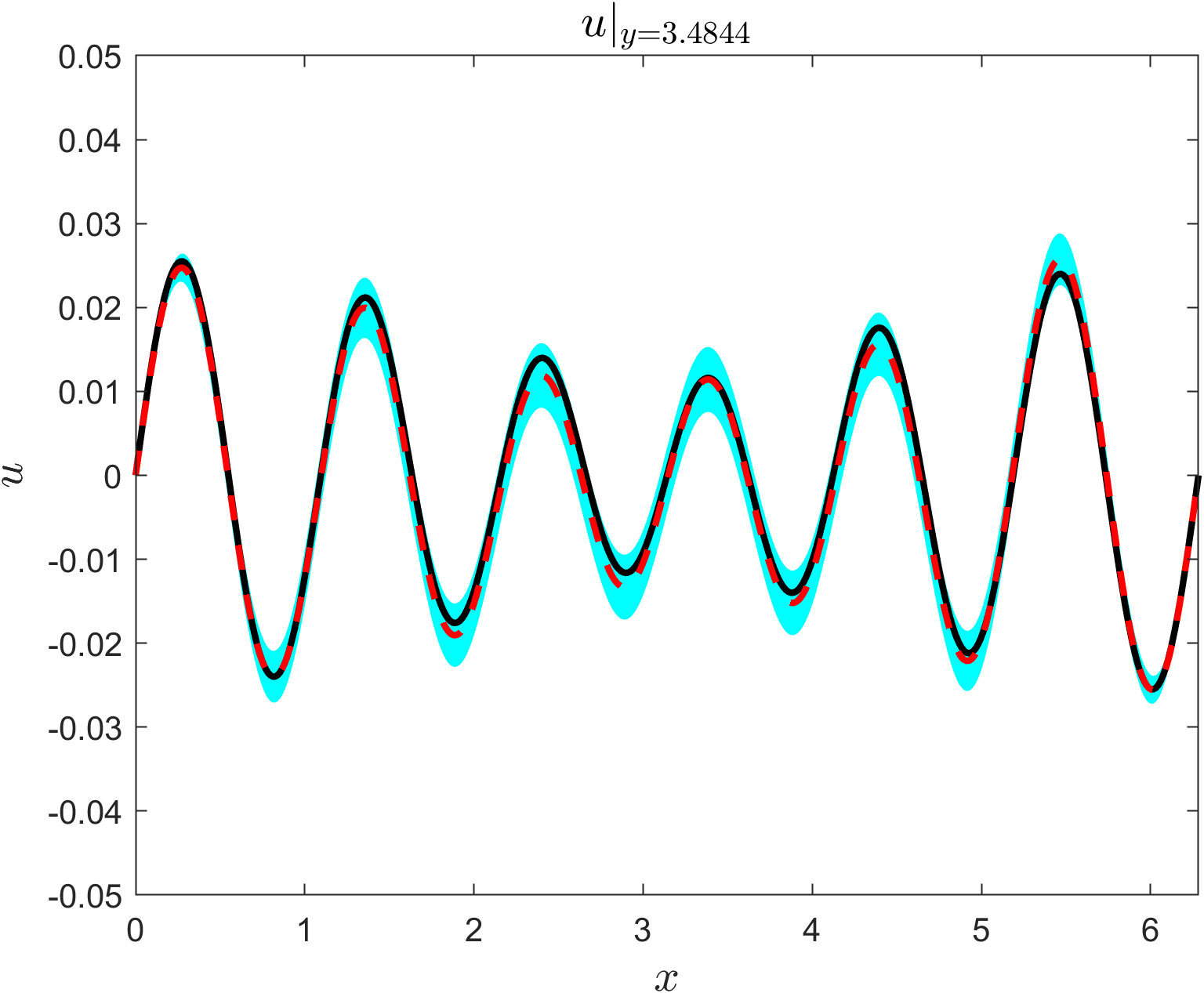}
        \includegraphics[scale=.3]{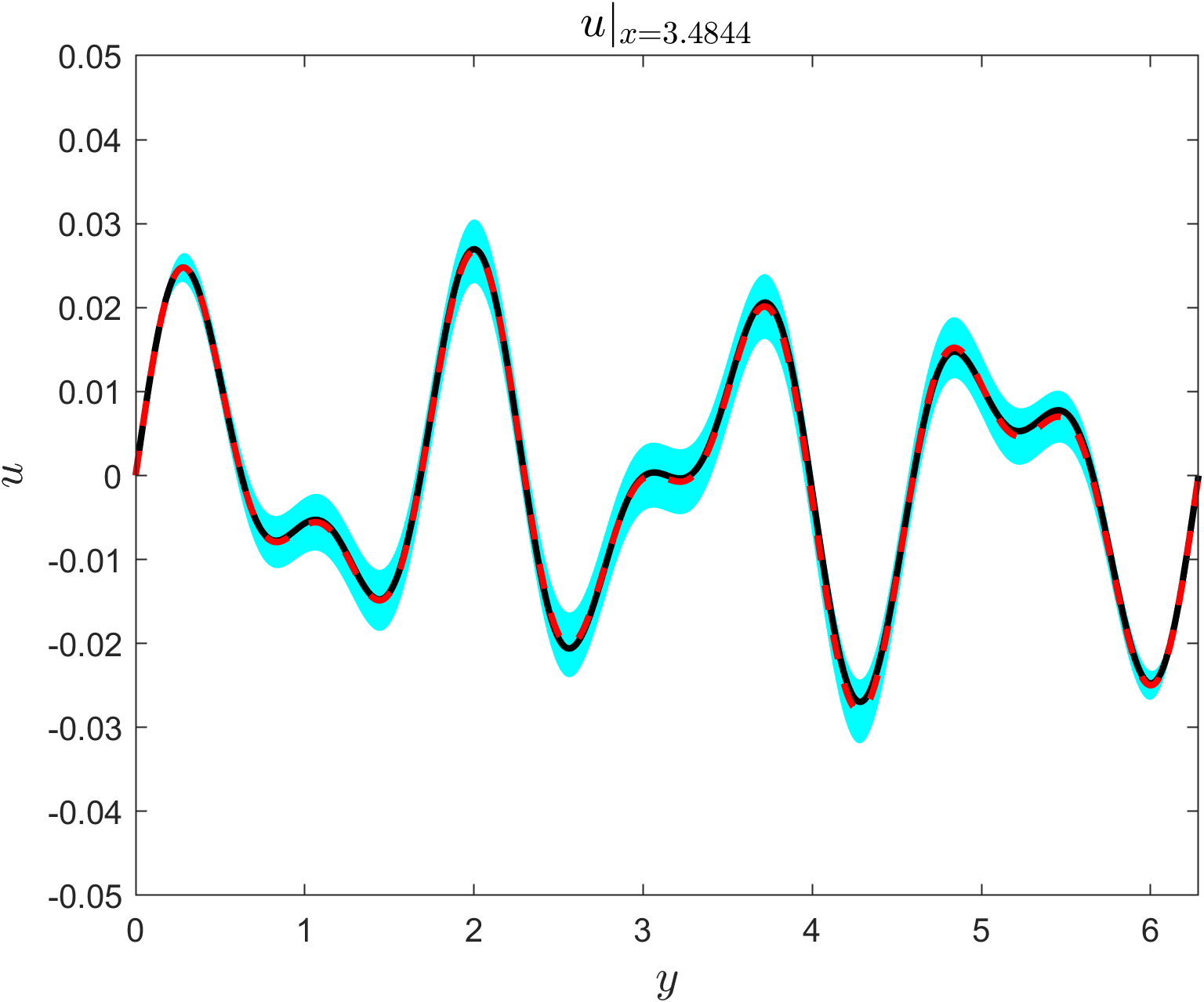}
    }
    \subfigure[Slices of the inference/fitting of $f$ with UQ.]{
        \includegraphics[scale=.3]{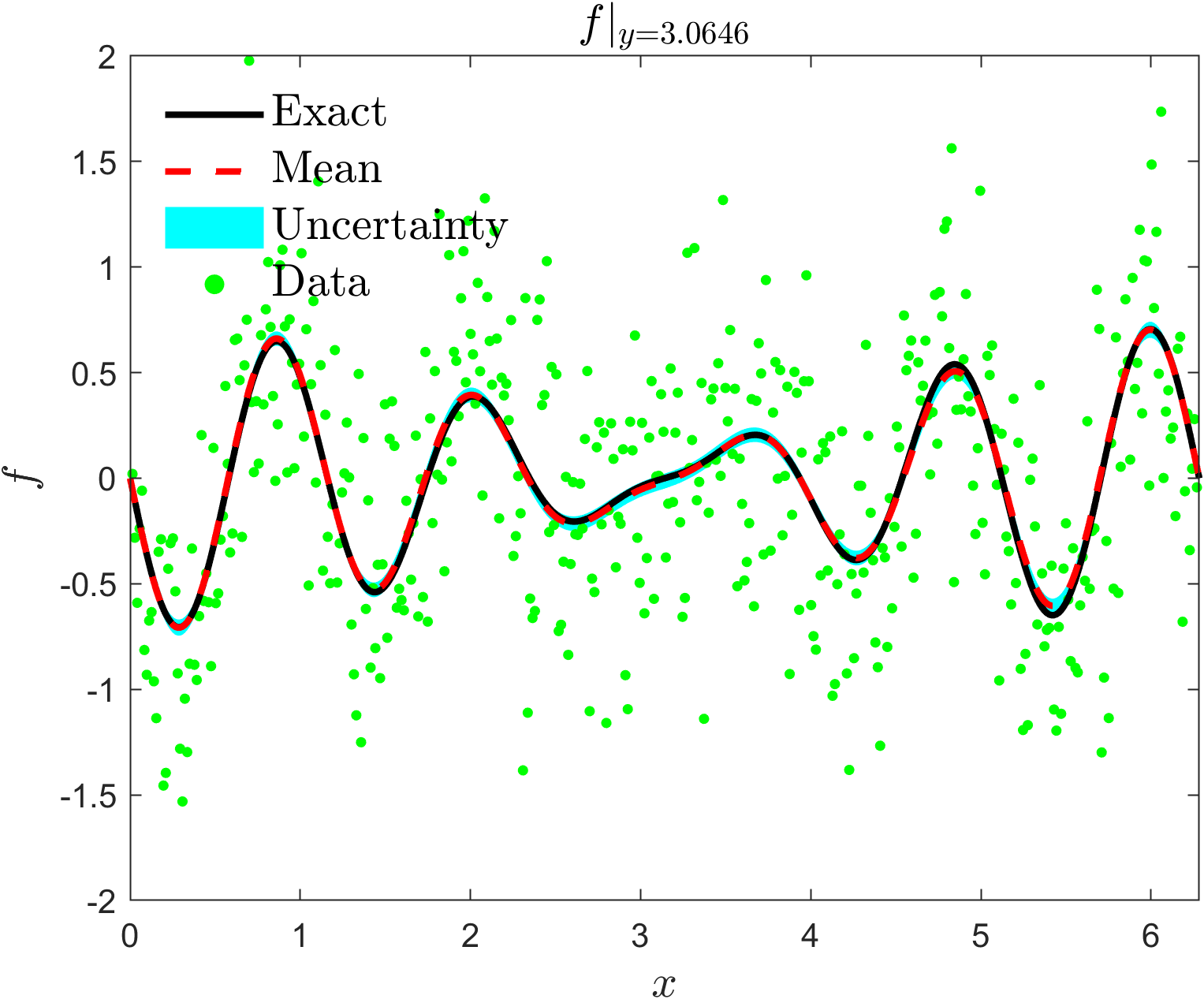}
        \includegraphics[scale=.3]{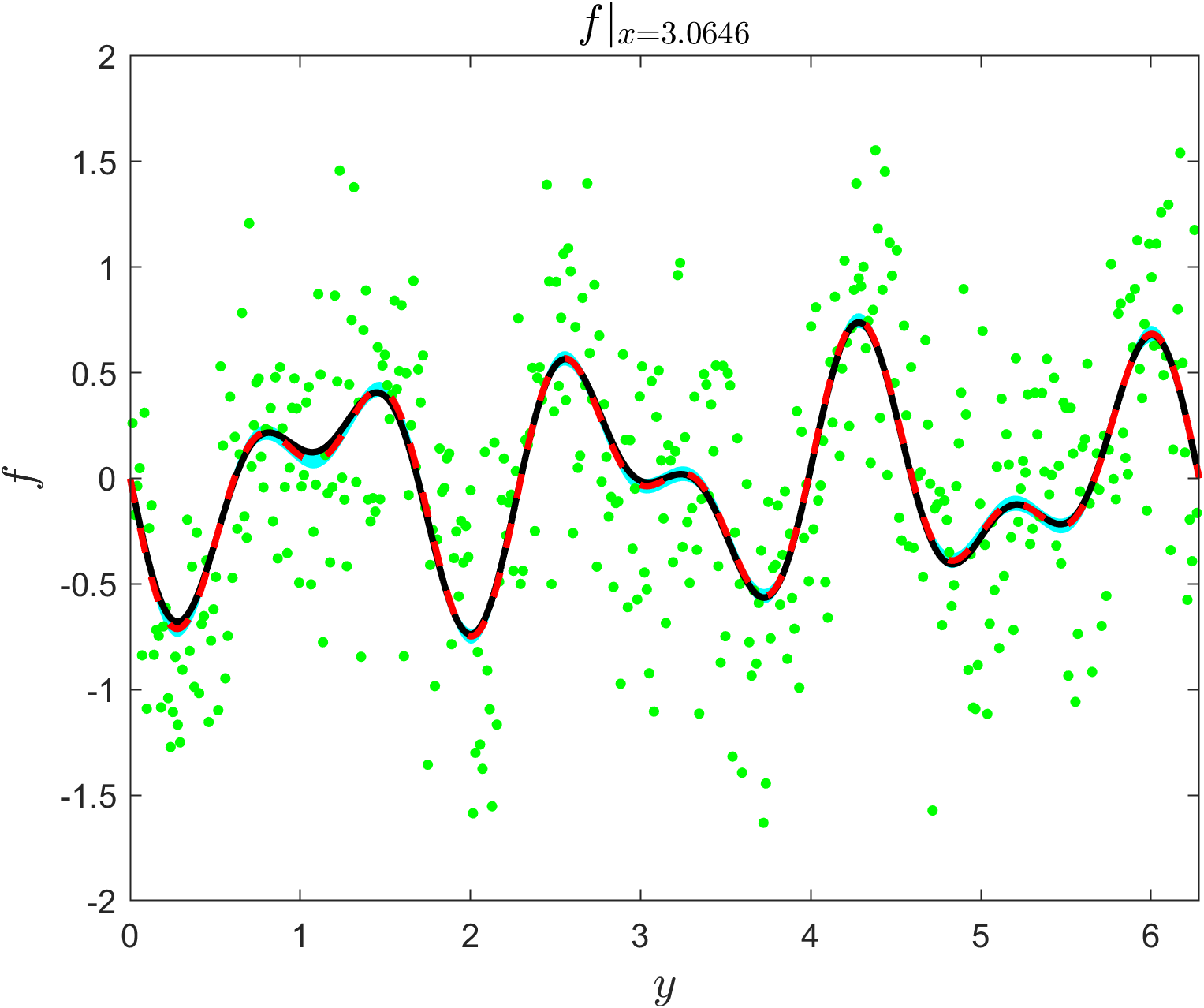}
    }
    \caption{Results of solving \eqref{eq:example_3} with noisy data of $f$ using our Riccati-based approach while traversing the domain sequentially. (a) displays the $7\times7$ uniform domain decomposition and the sequential order of integration of the domain. (b) shows the absolute errors of the predicted means of $u$ and $f$ after all 49 subdomains have been visited. (c) and (d) show 1D slices of the inferences and predicted uncertainties (\textcolor{myCyan}{$\blacksquare$}) of $u$ and $f$, respectively. These results show how our Riccati-based approach can naturally be extended to solve higher-dimensional problems. Namely, by leveraging data streaming, our Riccati-based approach is able to overcome the computational and memory challenges related to the higher dimension of the PDE and subsequently increased dataset size.}
    \label{fig:example_3_1}
\end{figure}

\begin{figure}[ht]
    \centering
    \subfigure[Absolute error of the inference of $u$.]{
       
        \includegraphics[width=.3\textwidth]{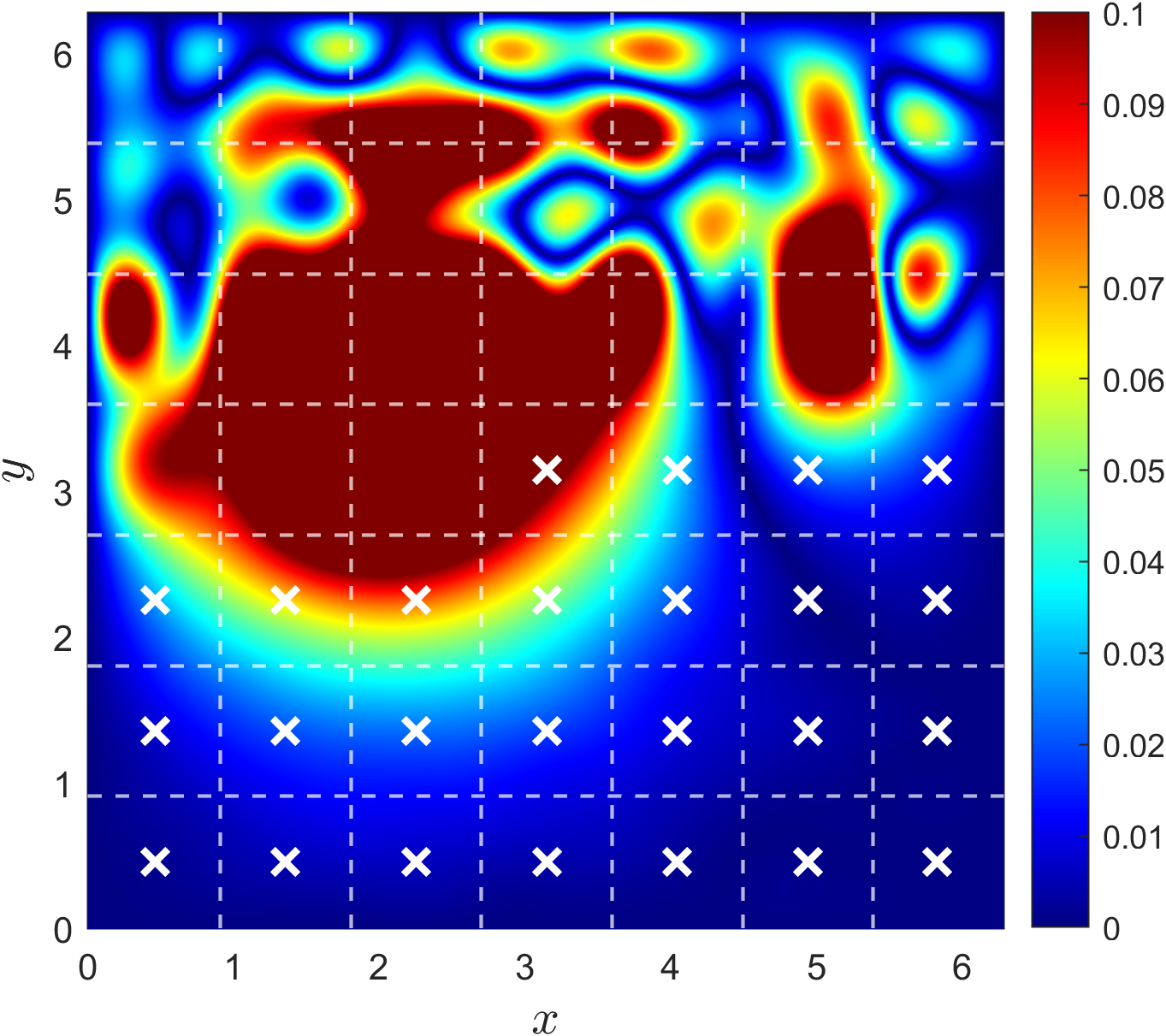}
        \includegraphics[width=.3\textwidth]{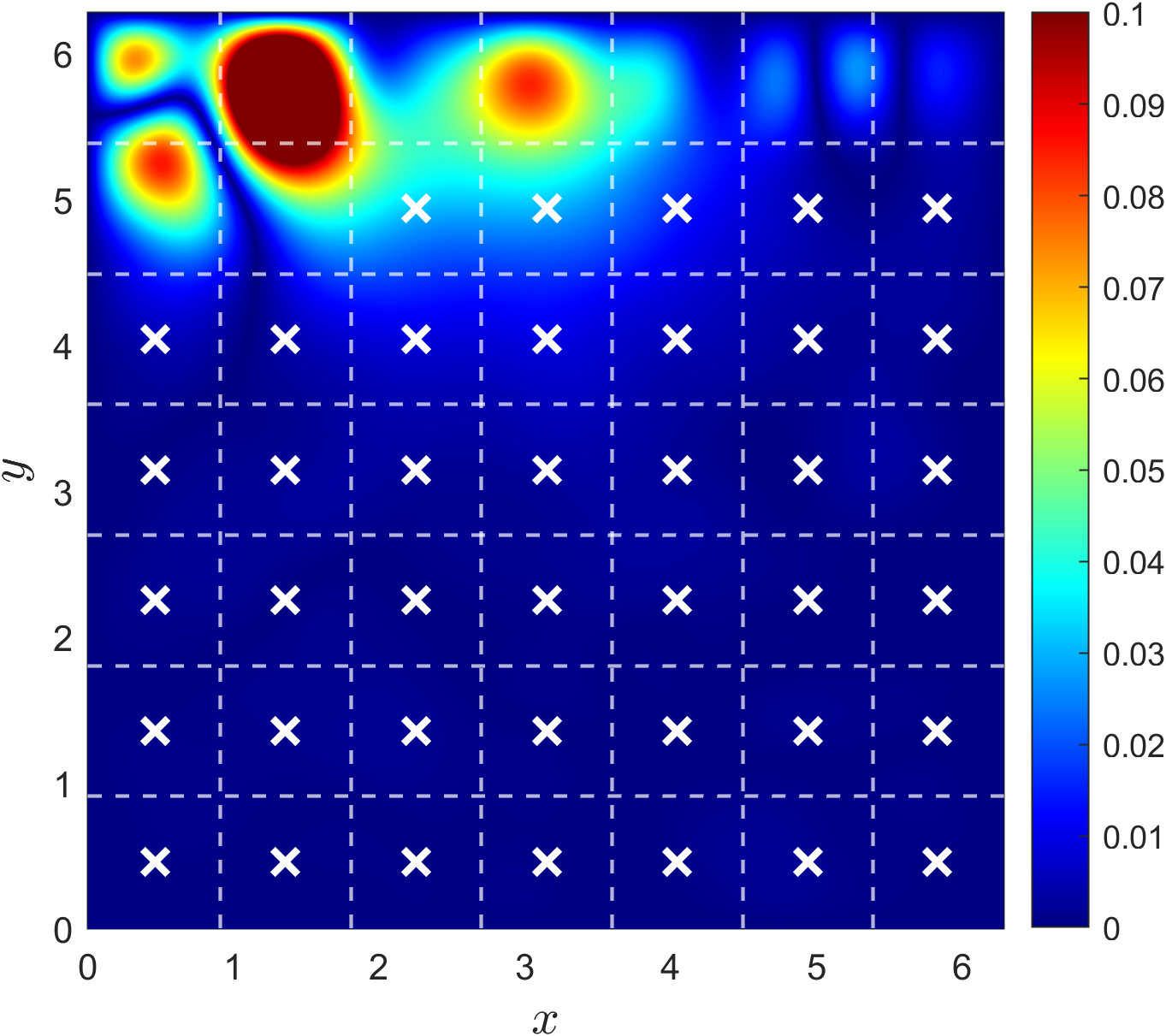}
        \includegraphics[width=.3\textwidth]{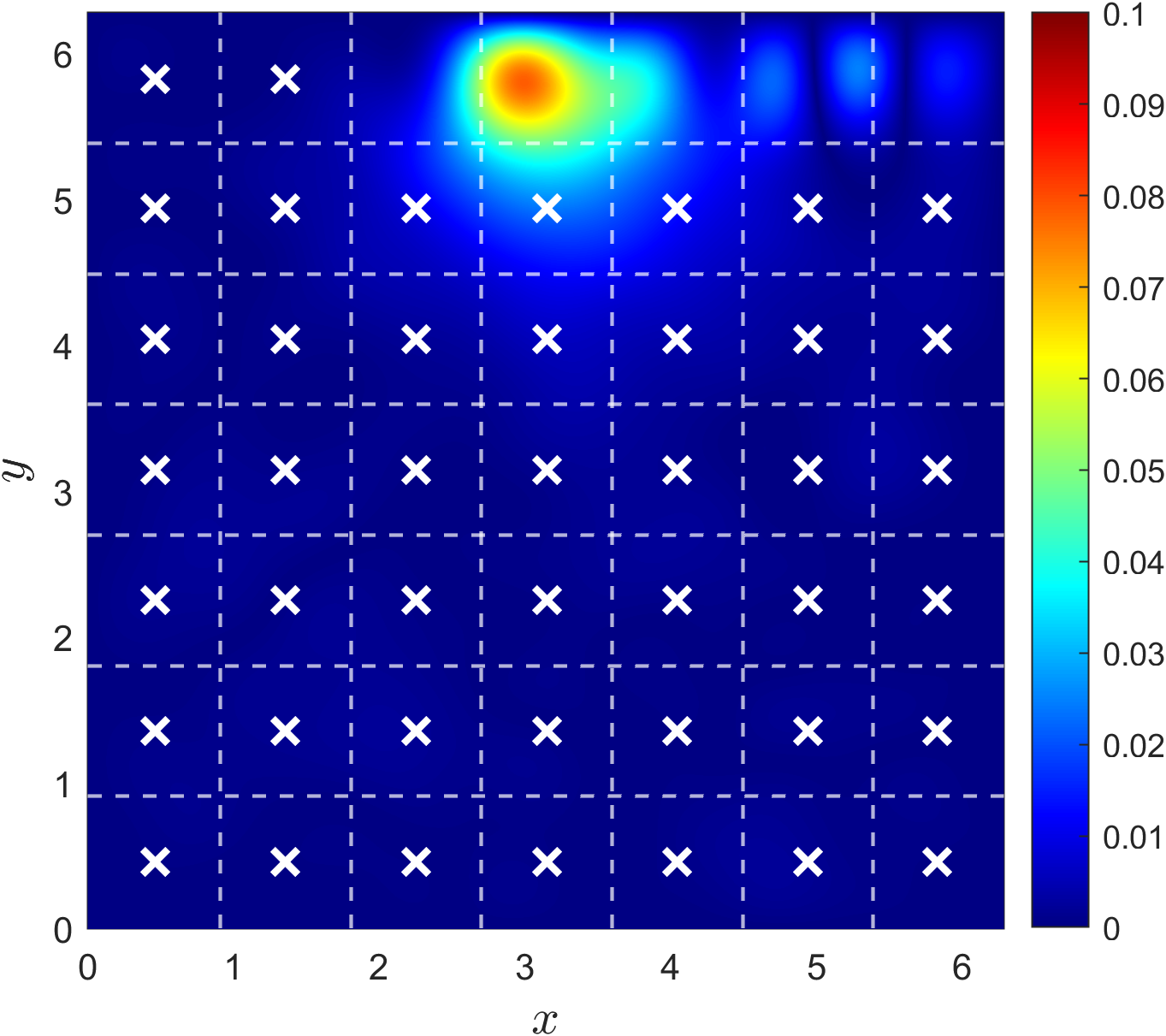}
    }
    \subfigure[Absolute error of the inference/fitting of $f$.]{
        \includegraphics[width=.3\textwidth]{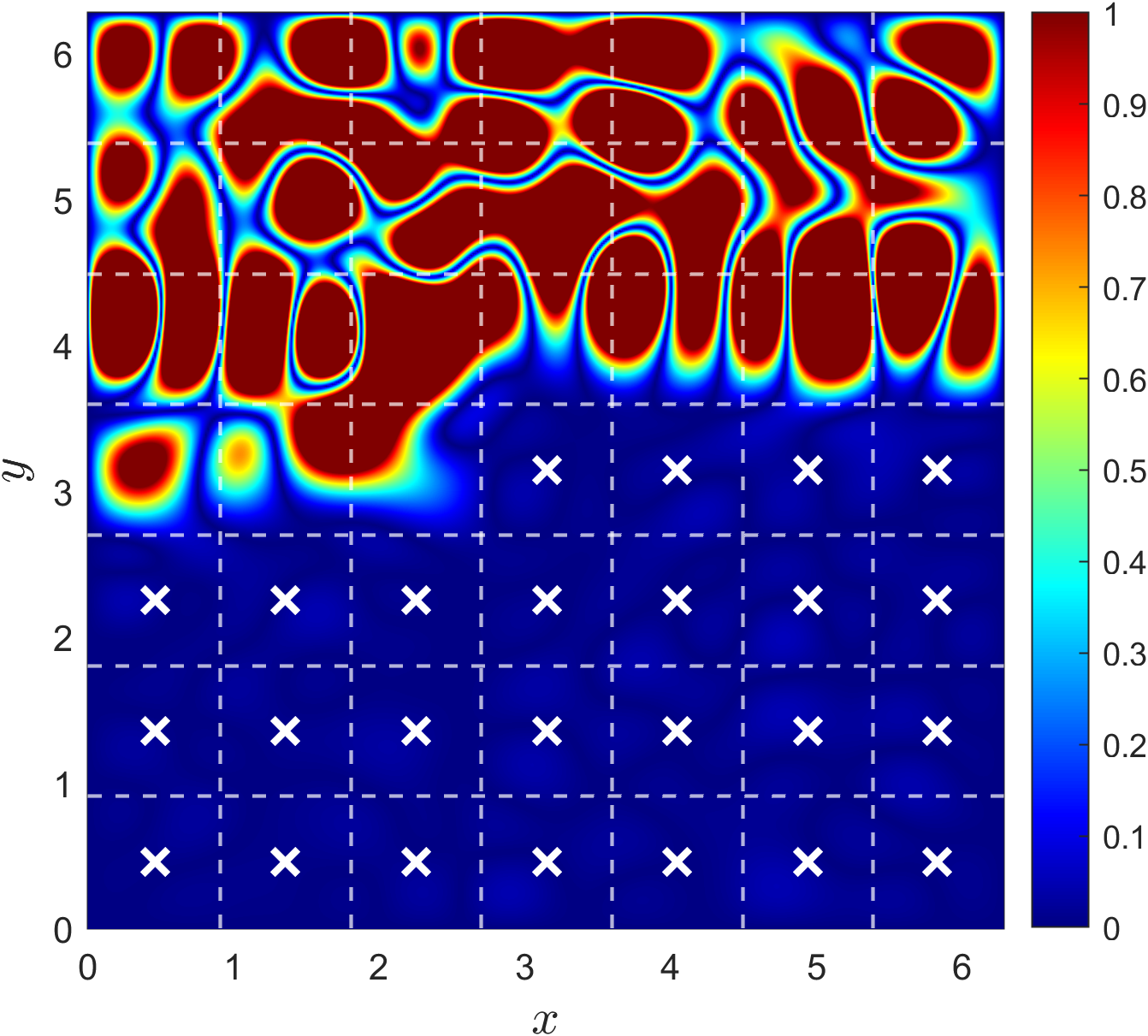}
        \includegraphics[width=.3\textwidth]{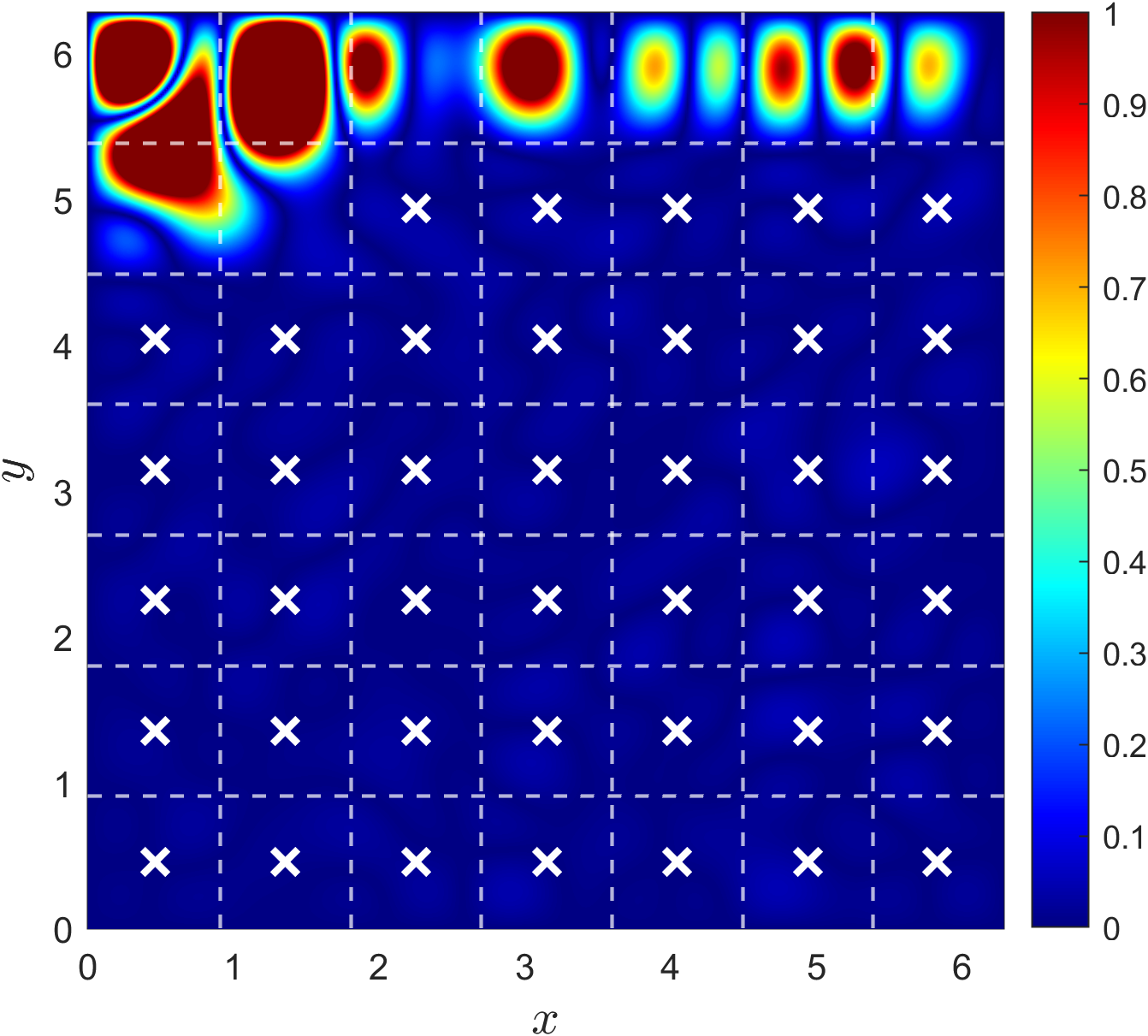}
        \includegraphics[width=.3\textwidth]{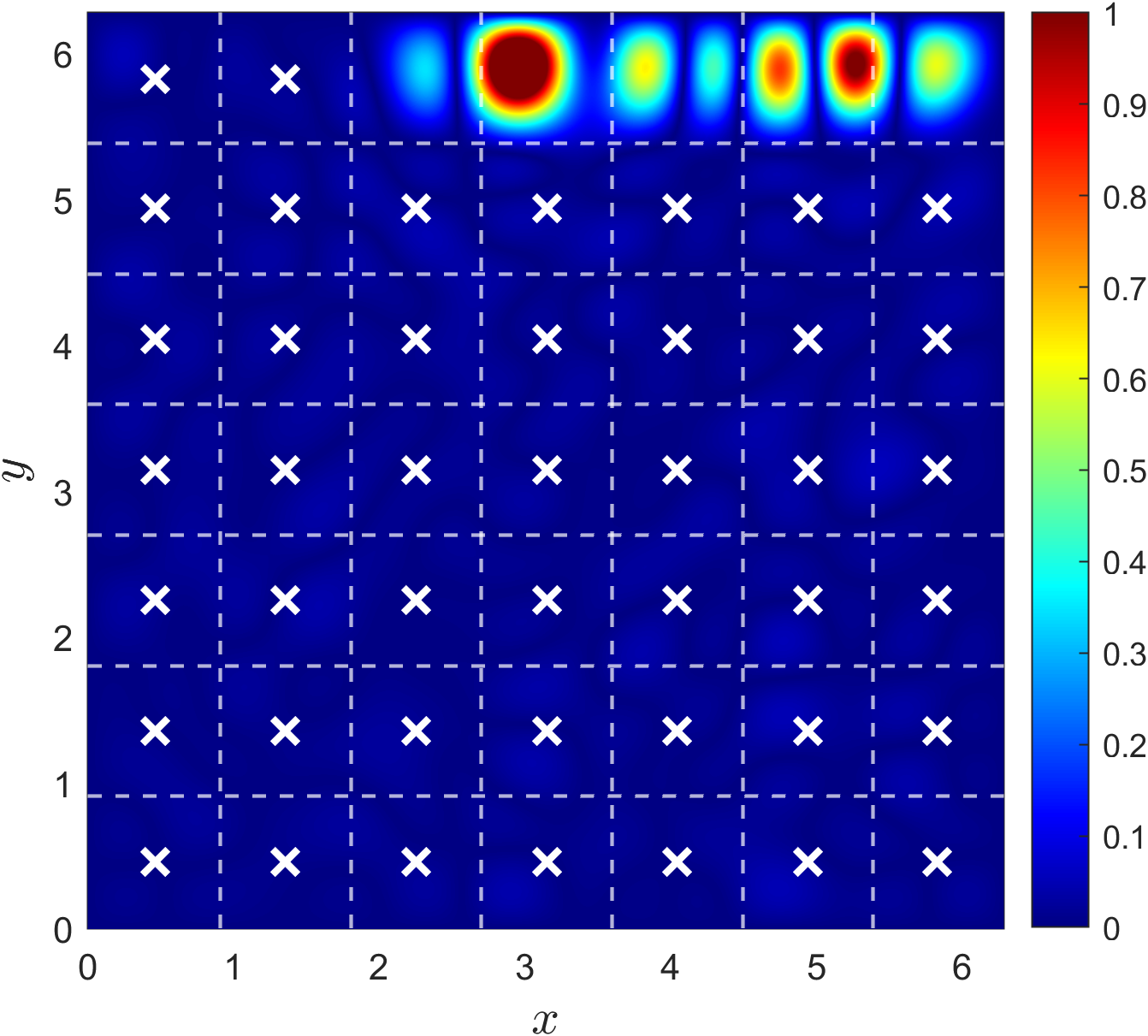}
    }
    \caption{Intermediate results of solving \eqref{eq:example_3} using our Riccati-based approach while traversing the domain sequentially. We collect data from each subdomain sequentially following the pattern in Figure \ref{fig:example_3_1}(a). (a) and (b) shows the absolute errors of the inference of $u$ and $f$, respectively. The white crosses indicate the subdomains that have already been visited. Our Riccati-based approach incrementally updates our predictions as we traverse the domain, which provides computational and memory advantages over having to process or store all of the data from the entire domain at once.}
    \label{fig:example_3_2}
\end{figure}

\begin{table}[ht]
    \footnotesize
    \centering
    \begin{tabular}{c|c|c|c|c}
    \hline
    \hline
    & $25$ subdomains & $40$ subdomains & $44$ subdomains & $49$ subdomains (all)\\
    \hline
        Error of $u$ & $897.75\%$ & $231.56\%$ & $93.19\%$ & $9.73\%$\\
       \hline
       Error of $f$ & $178.51\%$ & $62.74\%$ & $20.38\%$ & $2.32\%$ \\
       \hline
       \hline
    \end{tabular}
    \caption{Relative $L_2$ errors of the predicted means of $u$ and $f$ when solving \eqref{eq:example_3} using our Riccati-based approach and the sequential order of integration shown in Figure \ref{fig:example_3_1}(a). The errors significantly reduce as more subdomains are visited but remain slightly high after all of the data is incorporated due to the high level of noise in the data.}
    \label{tab:example_3}
\end{table}

In this case, we traverse the subdomains sequentially using the snaking pattern shown in Figure \ref{fig:example_3_1}(a). This decomposition pattern naturally extends the sequential data streaming pattern used for the 1D problem in Section~\ref{sec:4_2_1} to the 2D problem considered here. As such, this traversal pattern intuitively coincides with the Riccati-based methodology for adding data points in Section~\ref{sec:3_2_1}.

In Figure \ref{fig:example_3_1}, we see that we learn $u$ and $f$ fairly well, but the accuracy of the predicted means of $u$ and $f$ remains compromised by the large amount of noise in the data. In particular, Figures \ref{fig:example_3_1}(c)-(d) show that the exact values are still within the predicted uncertainty intervals around our predicted means, which indicates that our inferences are still reliable. 
In Figure~\ref{fig:example_3_2} and Table~\ref{tab:example_3}, we present intermediate results for the errors of our inferences after the domain has been partially traversed. In Figure~\ref{fig:example_3_2}, we observe that the absolute error of the inference of $f$ vanishes on each of the visited subdomains, whereas the absolute error of the inference of $u$ is reduced on the visited subdomains but not necessarily eliminated.
This behavior is due to the fact that we do not learn $u$ directly and instead learn $u$ from data of $f$ and information from the PDE.
In Table~\ref{tab:example_3}, we again see that the errors significantly reduce as we traverse domain, but the final error of our inference of $u$ remains somewhat high due to the high noise level of the data.
Since our Riccati-based approach incrementally updates our inferences with one data point at a time, it does not suffer from any computational or memory limitations despite the large dataset size.

\subsubsection{Case B: multi-level domain decomposition}\label{sec:multigrid_domain}

\begin{figure}[ht]
    \centering
    \includegraphics[width=.3\textwidth]{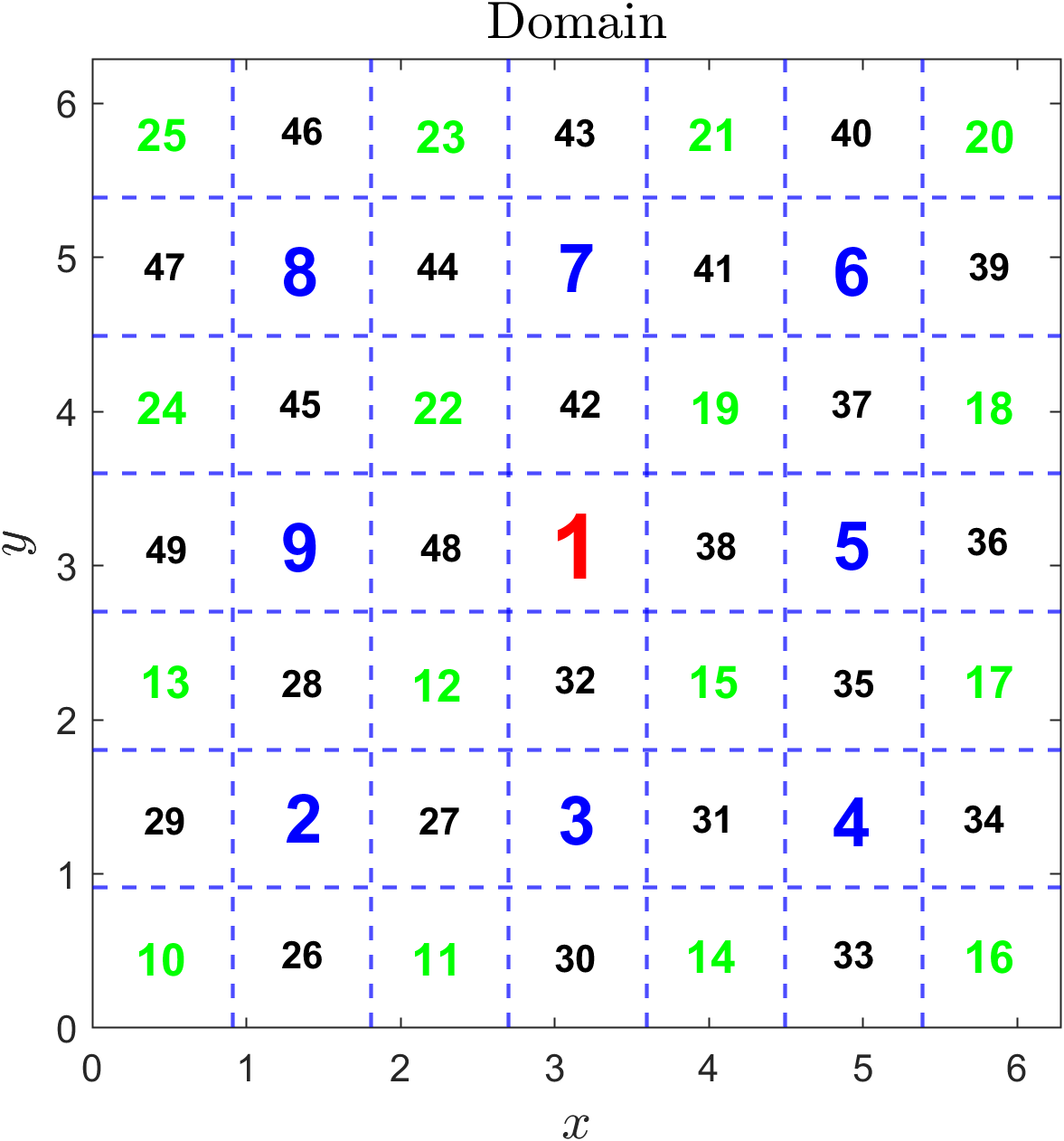}
    \caption{Domain decomposition with multi-level subdomain traversal. The level one subdomain is marked in {\color{red}\textbf{red}}, the level two subdomains are marked in {\color{blue}\textbf{blue}} numbers, the level three subdomains are marked in {\color{green}\textbf{green}} numbers, and the level four subdomains are marked in \textbf{black}.}
    \label{fig:example_3_3}
\end{figure}

\begin{figure}[ht!]
    \centering
    \subfigure[Absolute error of the inference of $u$.]{
        \includegraphics[width=.3\textwidth]{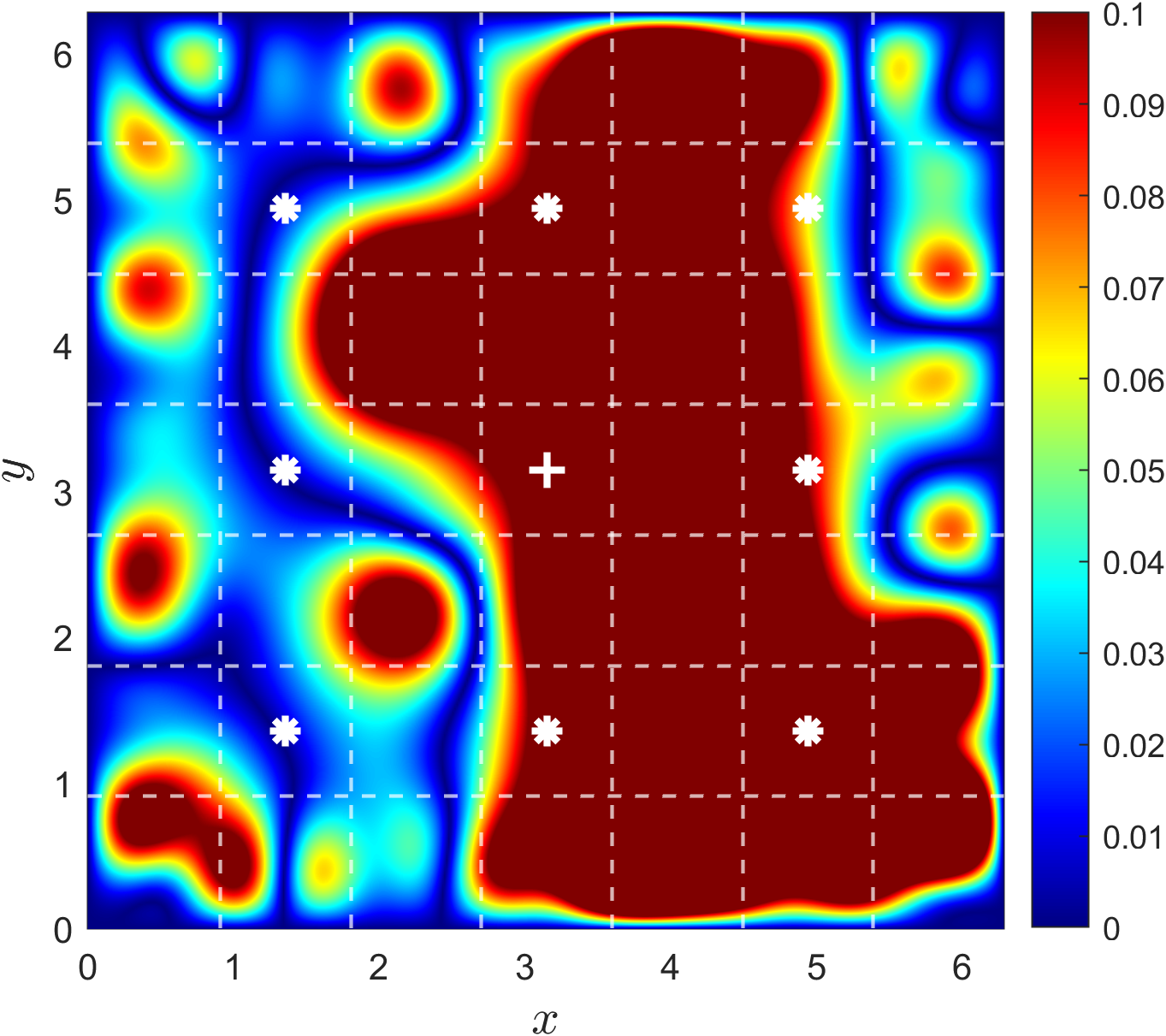}
        \includegraphics[width=.3\textwidth]{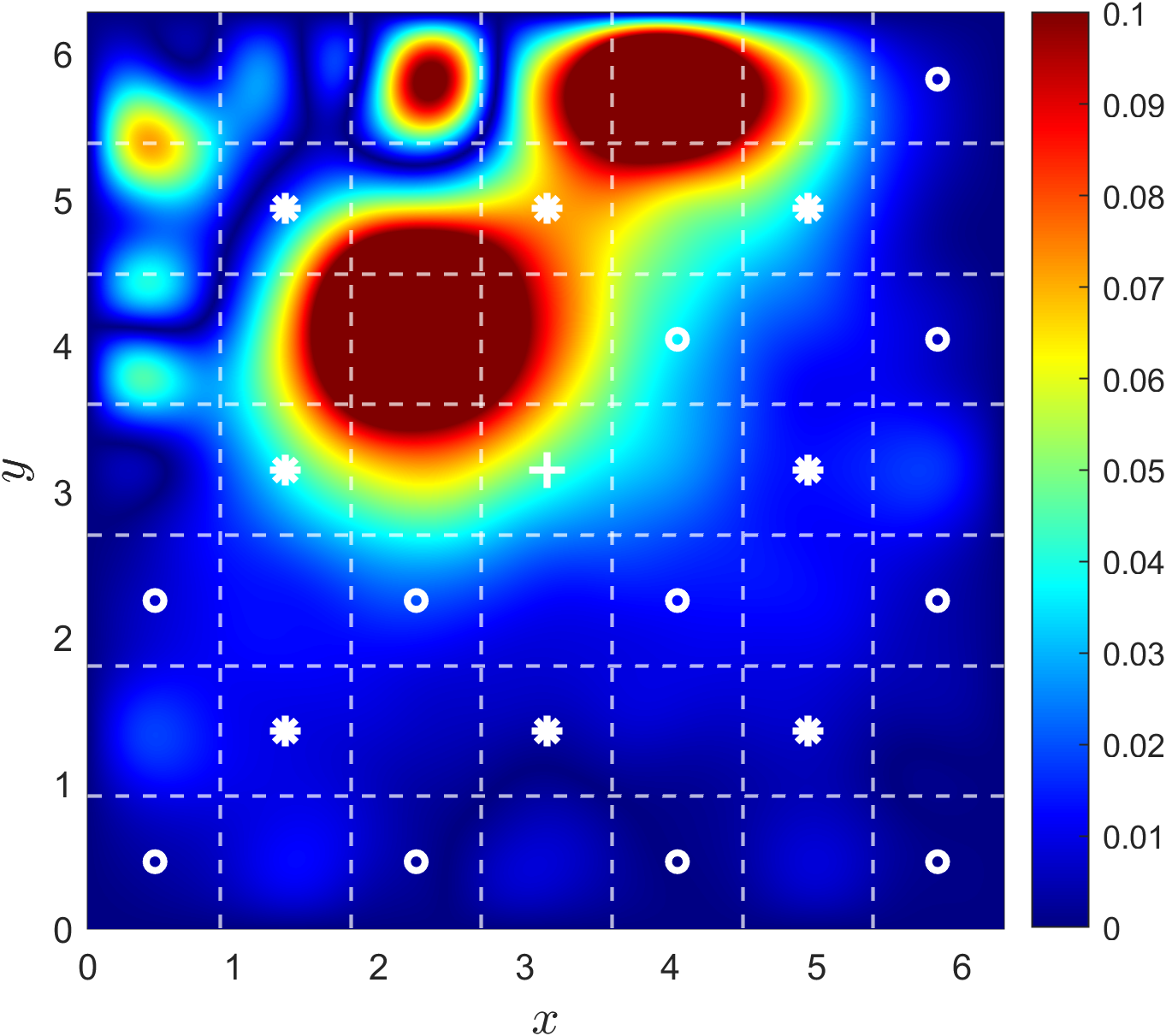}
        \includegraphics[width=.3\textwidth]{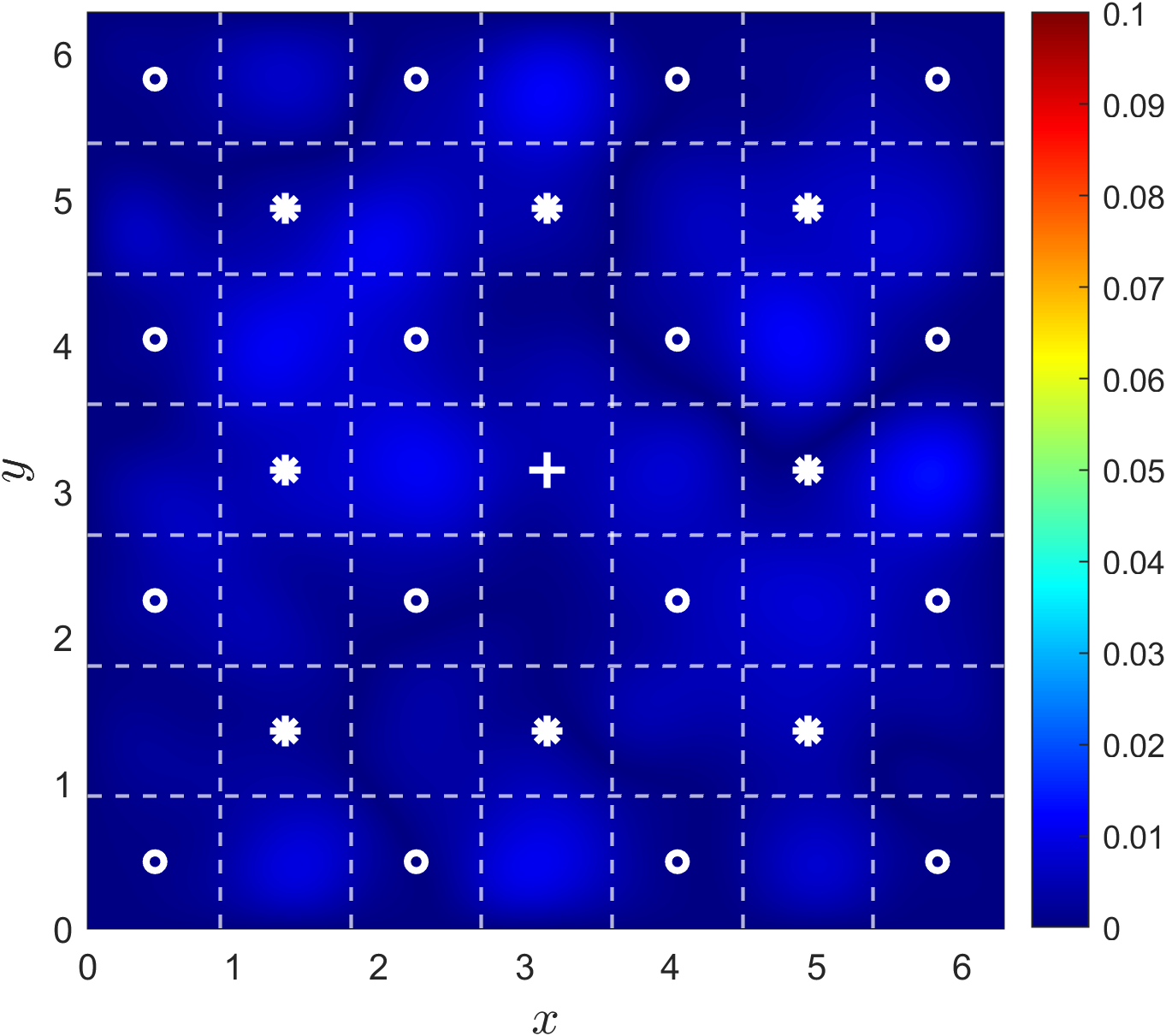}
    }
    \subfigure[Absolute error of the inference/fitting of $f$.]{
        \includegraphics[width=.3\textwidth]{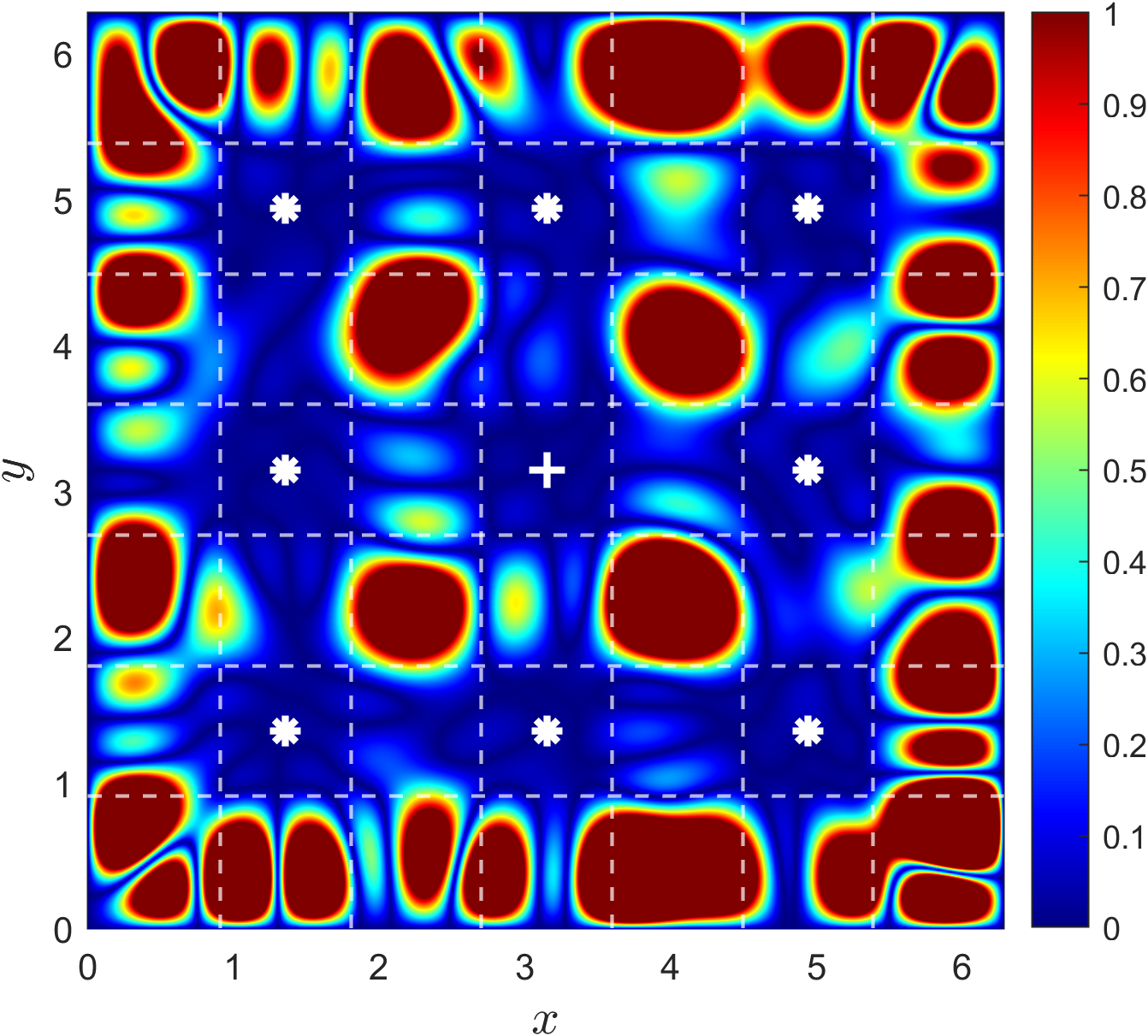}
        \includegraphics[width=.3\textwidth]{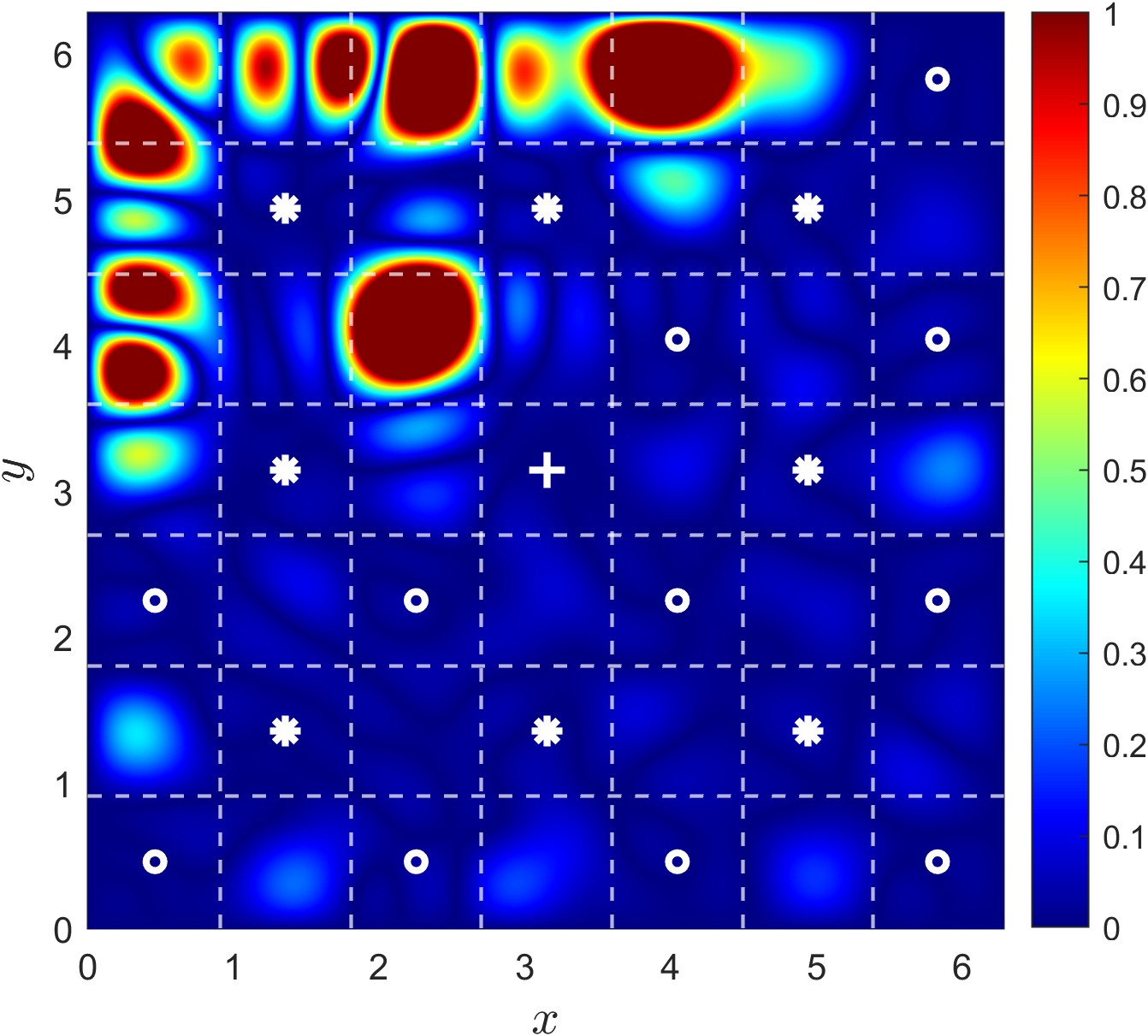}
        \includegraphics[width=.3\textwidth]{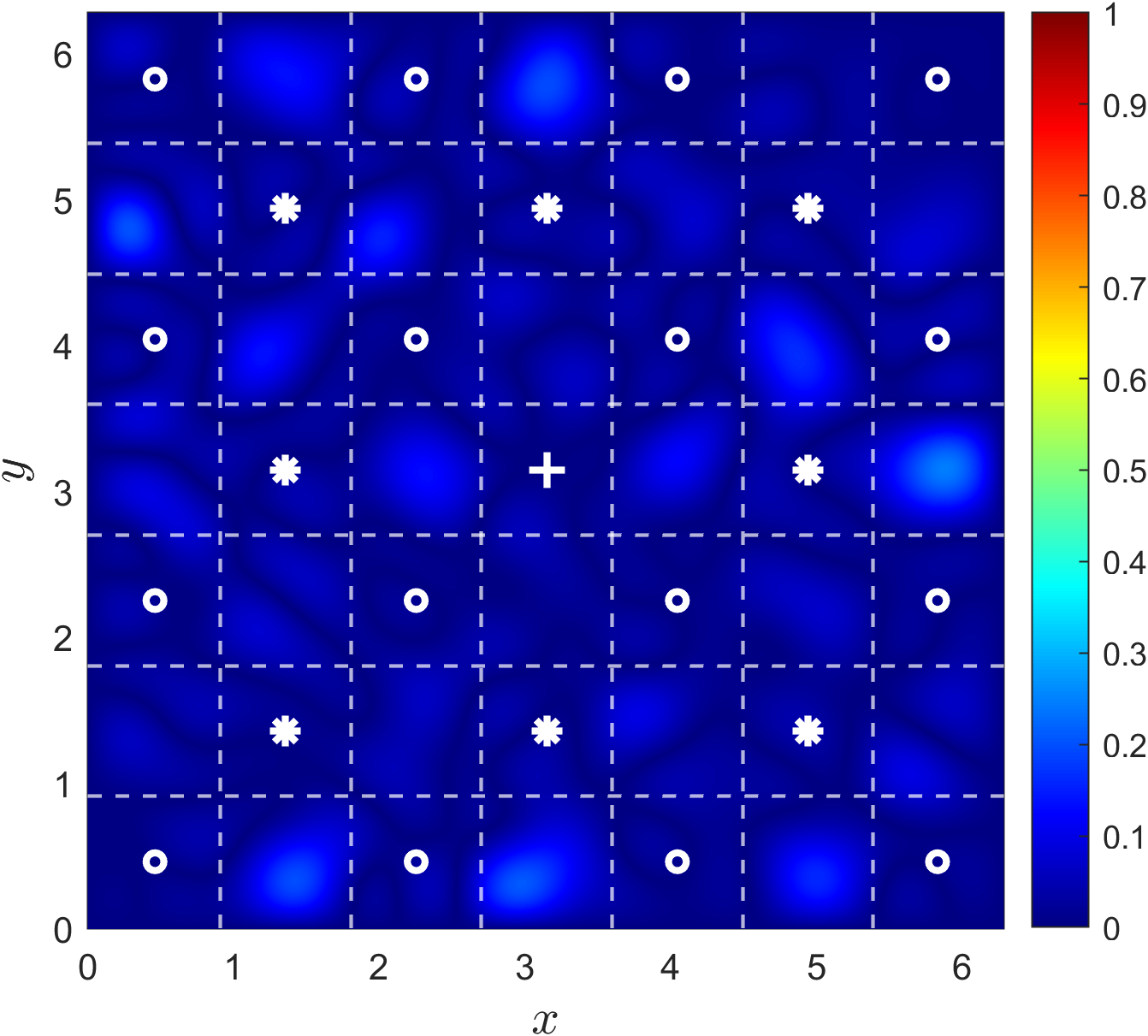}
    }
    \caption{Intermediate results of solving \eqref{eq:example_3} using our Riccati-based approach while traversing the domain using the multi-level order in Figure~\ref{fig:example_3_3}. (a) and (b) show the absolute errors of the inference of $u$ and $f$, respectively. The level one subdomain is marked by the white cross, the level two subdomains are marked by white stars, and the level three subdomains are marked by white circles. Our Riccati-based approach incrementally incorporates data points into the predictions invariant to the order of the data. As such, the end result after all the data is incorporated is identical to Figure~\ref{fig:example_3_1}(b)-(d), despite the intermediate results differing.}
    \label{fig:example_3_4}
\end{figure}

In this section, we follow a multi-level order of subdomain traversal inspired by multigrid methods \cite{hackbusch2013multi, mccormick1987multigrid} to highlight the invariance of our Riccati-based approach to the order of the data. In the previous case, we incorporated the data sequentially. However, this order is not required, as our Riccati-based approach can incrementally incorporate the data into the learned models in any order. Specifically, we consider the multi-level traversal order shown in Figure~\ref{fig:example_3_3}. Level one consists of the center subdomain. Level two consists of the eight subdomains roughly centered around the level one subdomain. Level three consists of 16 subdomains, each corresponding to one corner of each of the level two subdomains. Level four consists of the remaining 24 subdomains. 
Using this type of multi-level order has the potential to allow the learning process to stop earlier (and hence, use less data) because each level captures information across the entire domain. As a result, the error of the inferences is more likely to reduce uniformly across the domain. In contrast, the sequential traversal pattern used in Section~\ref{sec:seq_domain} is more likely to learn accurate predictions on the bottom half of the domain before learning any reasonably reliable prediction on the top half of the domain.

Figure \ref{fig:example_3_3} shows the intermediate results using this multi-level approach. As in the previous case, we see that the errors generally improve in/near the subdomains we have visited. The end result (not shown) after all of the subdomains have been visited is identical to Figure~\ref{fig:example_3_1}(b)-(d) with $9.73\%$ relative $L_2$ error for $u$ and $2.32\%$ relative $L_2$ error for $f$, which demonstrates the invariance of our Riccati-based approach to the order of the data points.

%% file: sec_5.tex
\section{Summary}\label{sec:summary}
In this paper, we established a new theoretical connection between Bayesian inference problems with linear models and Gaussian likelihoods and viscous HJ PDEs with quadratic Hamiltonian. As a first exploration of this connection, we specialized to Gaussian priors and leveraged the result to develop a new Riccati-based methodology for efficiently updating the predicted models. We then demonstrated some potential computational advantages of this Riccati-based approach by applying it to several UQ-based examples from SciML \cite{karniadakis2021physics, psaros2023uncertainty, zou2022neuraluq}. 
In particular, these examples illustrate that this approach naturally coincides with data streaming applications \cite{parisi2019continual, van2019three, settles2009active, ren2021survey}. As such, this Riccati-based approach is amenable to memory-limited hardware architectures, such as FPGAs \cite{KastnerFPGA}, which opens possibilities for real-time applications requiring embedded implementations.

Some other possible future directions for this work are as follows. Currently, our theoretical connection requires a linear model, Gaussian likelihood, and particular assumptions on the prior. To consider a wider range of applications, more research is needed to extend this connection to nonlinear models, non-Gaussian likelihoods, and more general priors. Specifically, the link between Bayesian inference and viscous HJ PDEs no longer holds when the Hamiltonian is non-quadratic. Moreover, the Riccati-based methodology in Section~\ref{sec:riccatimethod} further restricts the prior to be Gaussian. As such, more work is also needed to explore how this connection could be leveraged to create efficient numerical methods for more general learning settings. Another natural extension of this work would be to investigate how this connection could be leveraged to reuse existing efficient ML algorithms to solve high-dimensional HJ PDEs. So far, this work only considers the opposite direction but opens opportunities for new ML-based solvers for fields related to HJ PDEs (e.g., stochastic optimal control~\cite{yong2012stochastic, fabbri2017stochastic}).

%% file: sec_appendix.tex
\section{Details of the hyperparameters in the numerical examples}\label{sec:hyperparameters}

In Section~\ref{sec:examples}, we use RK4 to numerically solve the Riccati ODEs, where the step size $h$ of RK4 is chosen to achieve a balance between high accuracy and stable, efficient computations.
We use Python, the NumPy library \cite{harris2020array}, the JAX library \cite{jax2018github}, and double precision in all numerical examples. An NVIDIA A100 GPU is used to accelerate computations.

In Section~\ref{sec:example_1a}, we use $h=5\times 10^{-5}$. In Section~\ref{sec:example_1b}, the baseline solution using $\sigma=1$ is computed using the method of least squares. Then, $\sigma$ is tuned using the values of $\sigma$ and $h$ shown in Table~\ref{tab:hyperparameter}. 
In Appendix~\ref{sec:example_1c}, the baseline solution using the entire training set is computed using the method of least squares, and we use RK4 with step size $h=1\times 10^{-5}$ to delete each outlier.
In Section~\ref{sec:4_2_1}, we use $h=1\times 10^{-1}$, and in Section~\ref{sec:4_2_2}, we use $h=1\times 10^{-2}$.
In Section~\ref{sec:example_3}, we use $h=2\times10^{-6}$ for both domain decomposition patterns.

\begin{table}[ht]
    \footnotesize
    \centering
    \begin{tabular}{|c|c|c|c|c|c|}
    \hline
    $\sigma$ & $1\rightarrow 0.5$ & $1\rightarrow 2$ & $2\rightarrow 5$ & $5\rightarrow 10$ & $10\rightarrow 20$ \\
    \hline
    $h$ & $10^{-5}$ & $10^{-5}$ & $10^{-6}$ & $10^{-7}$ & $10^{-7}$ \\
    \hline
    \end{tabular}
    \caption{Step size $h$ of RK4 when tuning $\sigma$ in the hyperparameter tuning example in Section~\ref{sec:example_1b}. $\sigma_0 \rightarrow \sigma_1$ denotes the case where $\sigma$ is tuned from $\sigma_0$ to $\sigma_1$.}
    \label{tab:hyperparameter}
\end{table}

\section{Additional results for example 1}\label{sec:example_1c}

\begin{figure}[ht!]
    \centering
    \subfigure[Inference of $u$.]{
        \includegraphics[width=.3\textwidth]{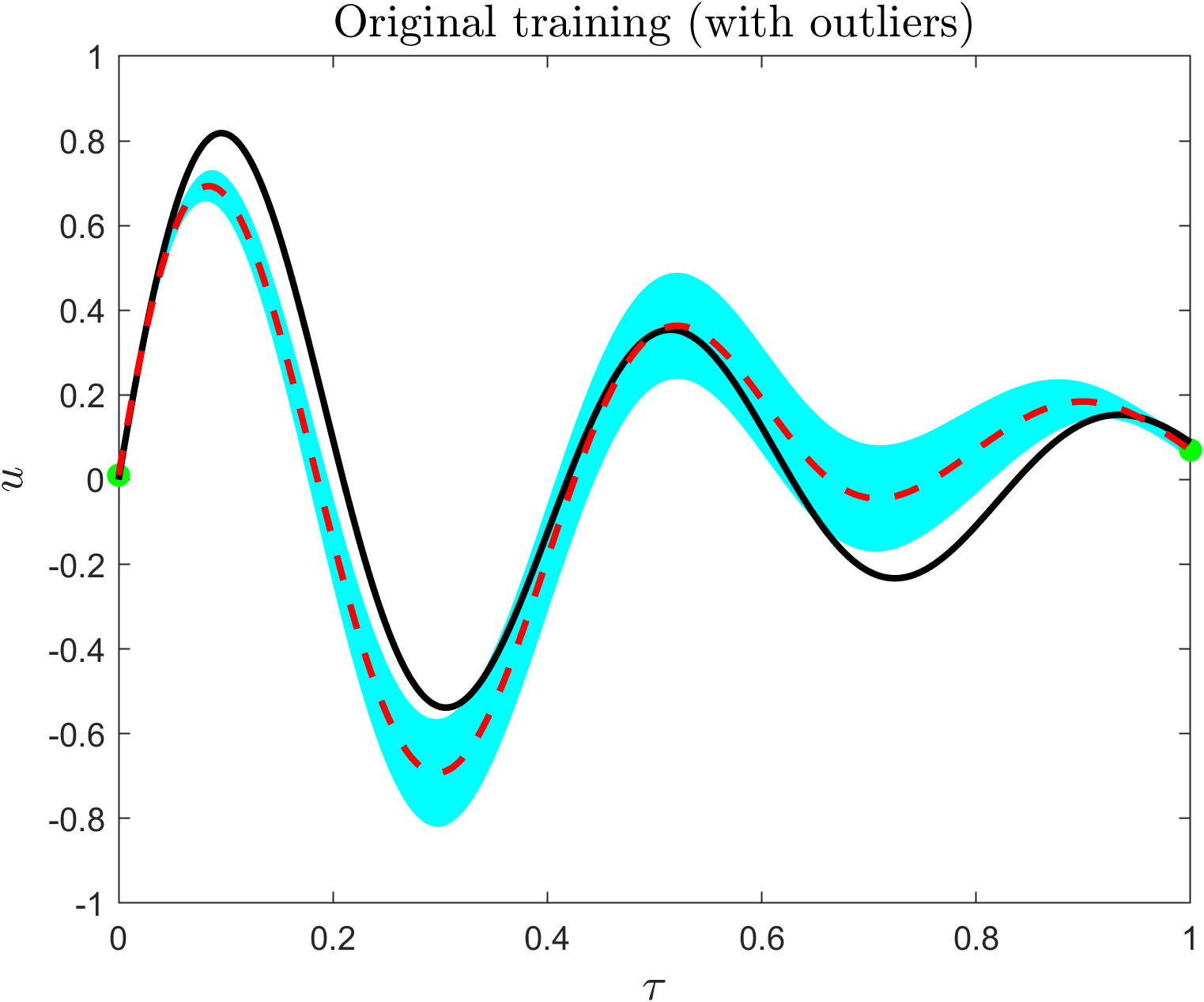}
        \includegraphics[width=.3\textwidth]{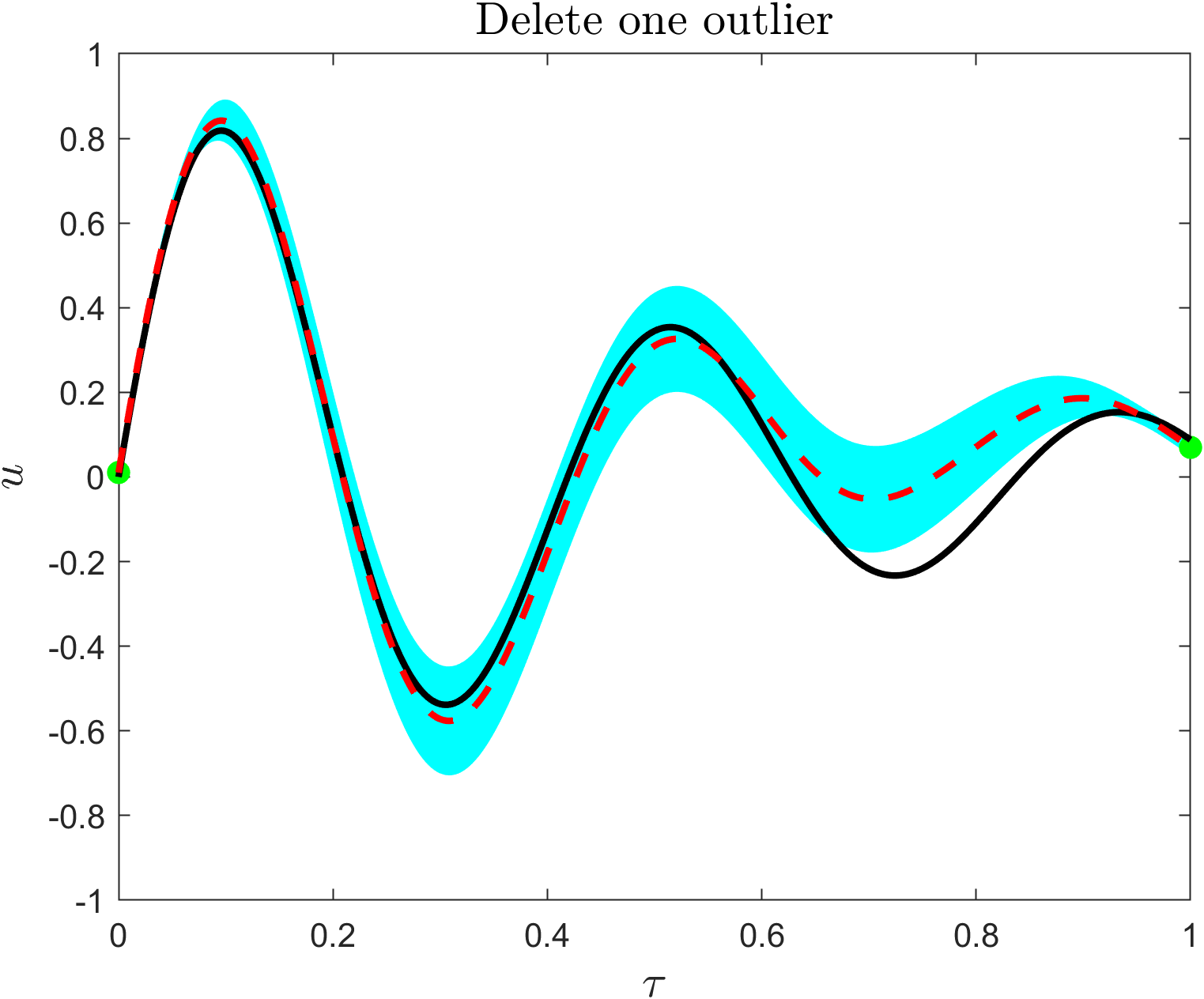}
        \includegraphics[width=.3\textwidth]{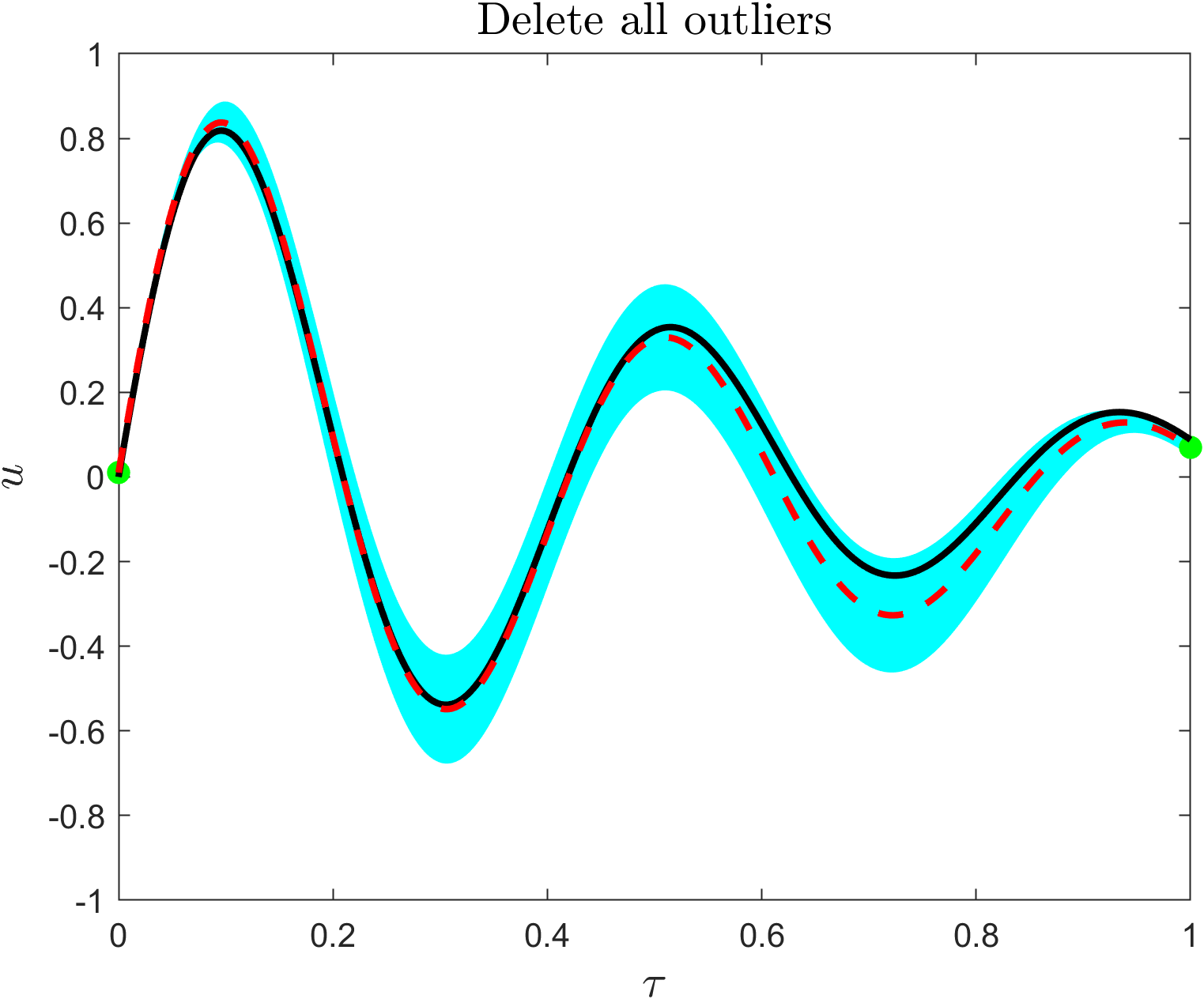}
    }
    \subfigure[Inference/fitting of $f$.]{
        \includegraphics[width=.3\textwidth]{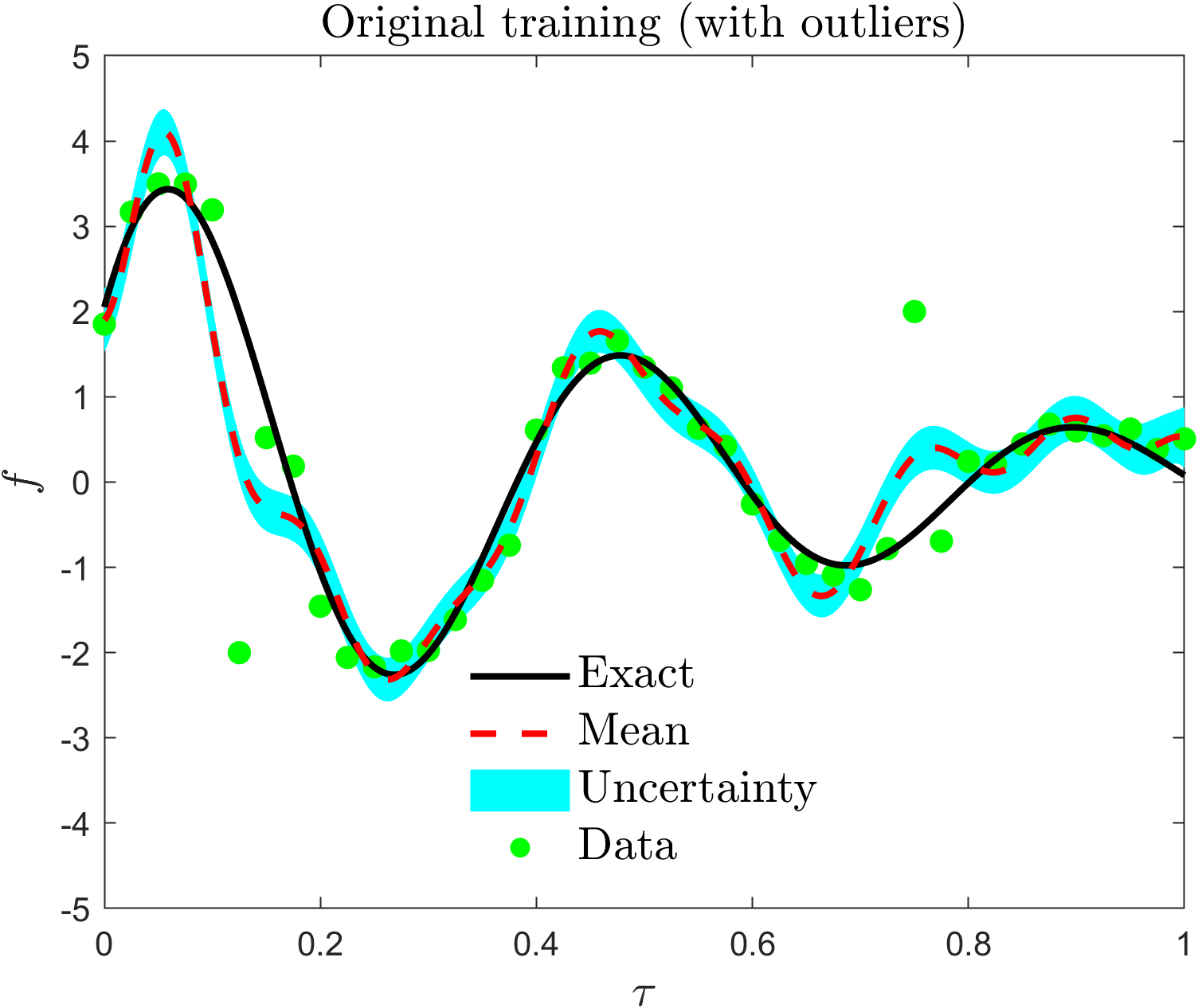}
        \includegraphics[width=.3\textwidth]{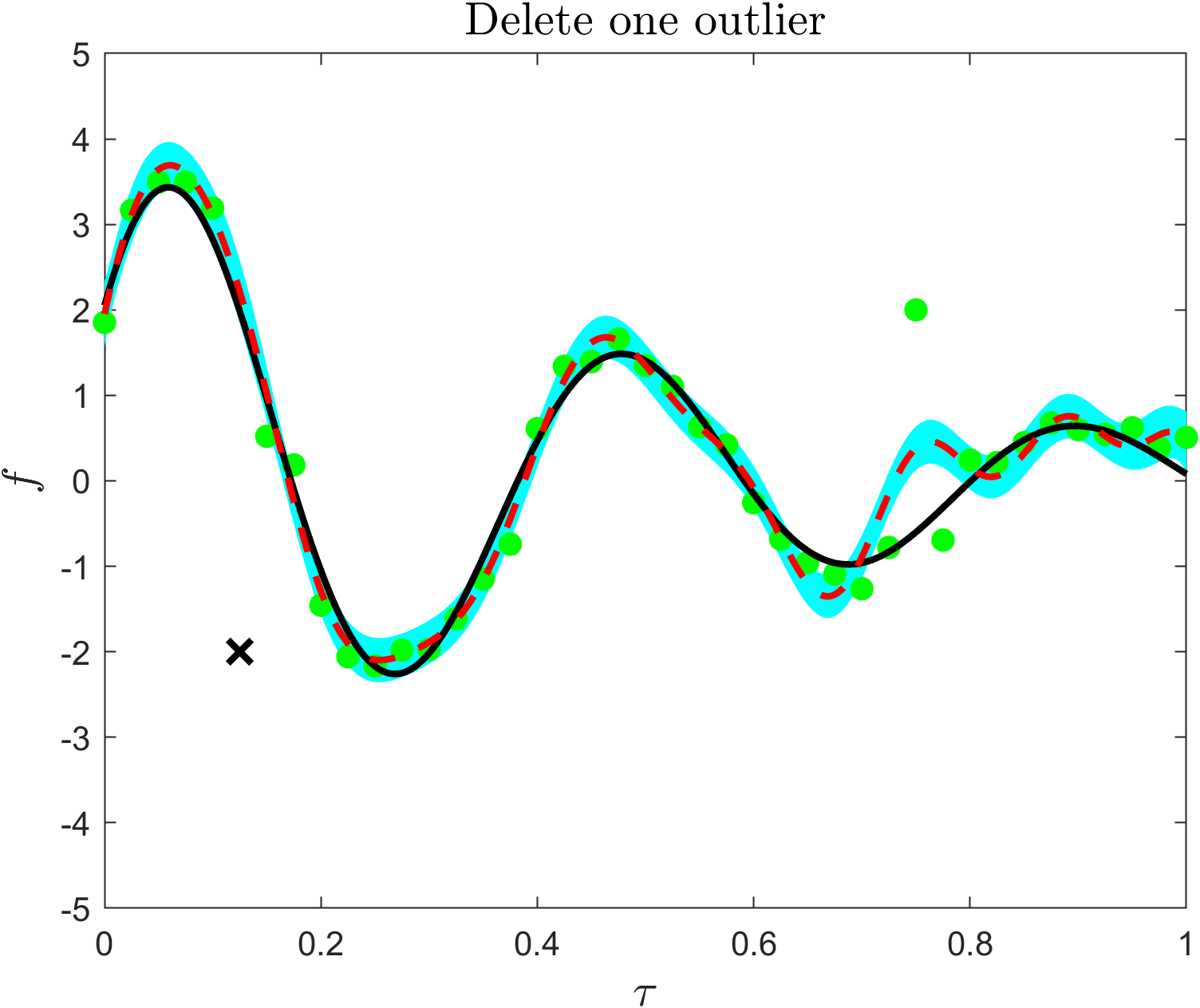}
        \includegraphics[width=.3\textwidth]{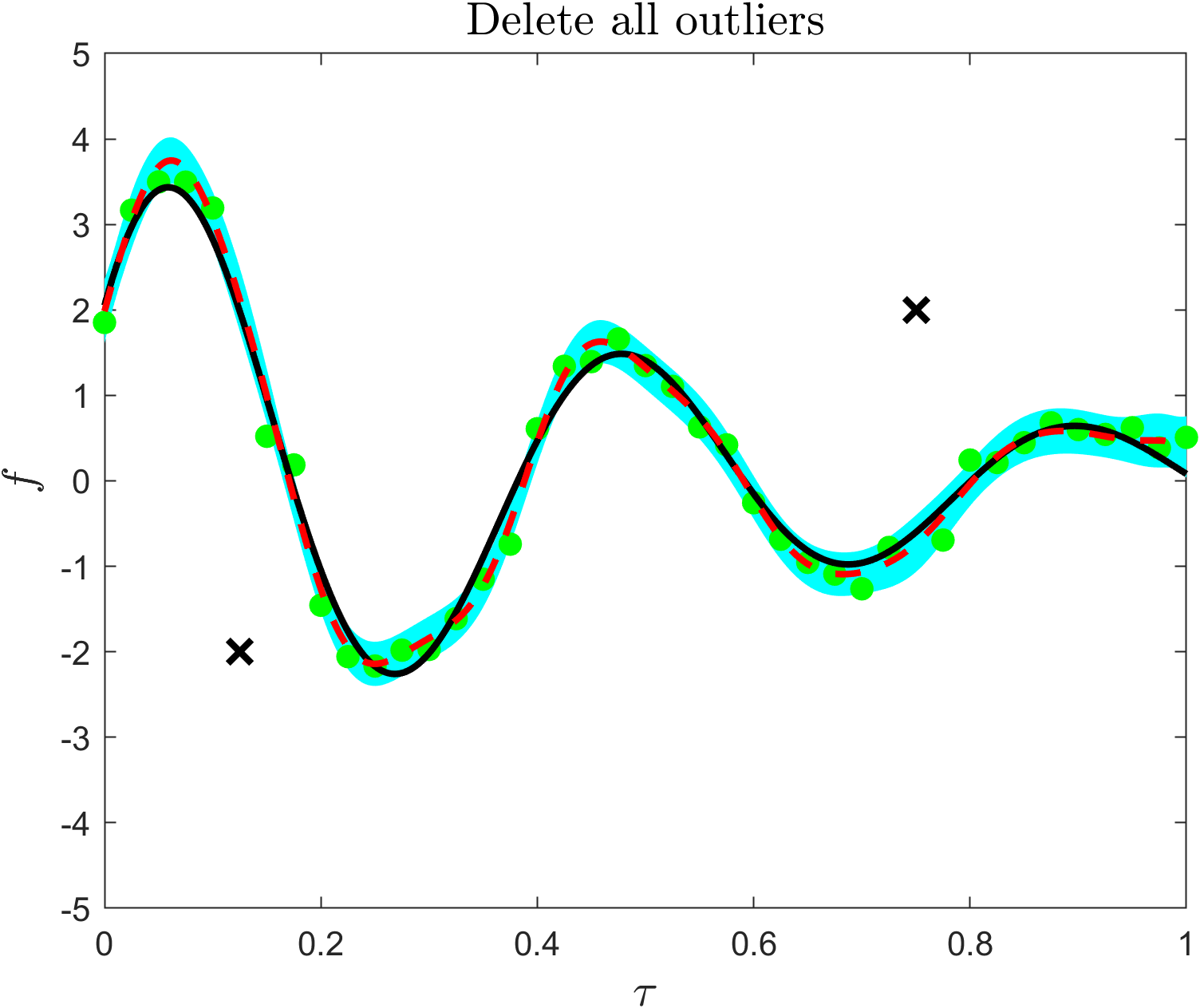}
    }
    \caption{Deleting two outliers ($\boldsymbol{\times}$) sequentially from the trained model using our Riccati-based approach to solve \eqref{eq:example_1}. The \textbf{left} column of (a) and (b) shows the inferences of $u$ and $f$, respectively, after the initial training with all data points. The \textbf{middle} column shows the inferences after the first outlier is deleted, and the \textbf{right} column shows the inferences after two outliers are deleted. The outliers are deleted by solving the associated Riccati ODEs backward in time. This process only uses the results of the previous training step and information of the point to be deleted, which provides potential computational advantages over more standard SciML approaches that would otherwise require retraining on the entire remaining dataset. }
    \label{fig:example_1_3}
\end{figure}

In this section, we present the results of removing outliers from the trained model when solving~\eqref{eq:example_1} using our Riccati-based methodology from Section~\ref{sec:3_2_1}.
Specifically, we first train the model with the entire noisy training set, and then remove the effect of identified outliers from the learned model.
Figure~\ref{fig:example_1_3} shows the results of the initial training and the subsequent sequential removal of two outliers. As expected, removing the outliers improves the model prediction accuracy. After both outliers are removed, the exact values now lie within the uncertainty band around the predicted means, which indicates that our models have become more reliable. Recall that the outliers are removed by solving one step of Riccati ODEs backward in time. As such, we do not require retraining on or access to the entire remaining dataset, which provides potential computational savings.

\section{Additional results for example 2}\label{sec:example_2c}

\begin{figure}[ht]
    \centering
    \subfigure[Inferences of $f$.]{
        \includegraphics[width=.22\textwidth]{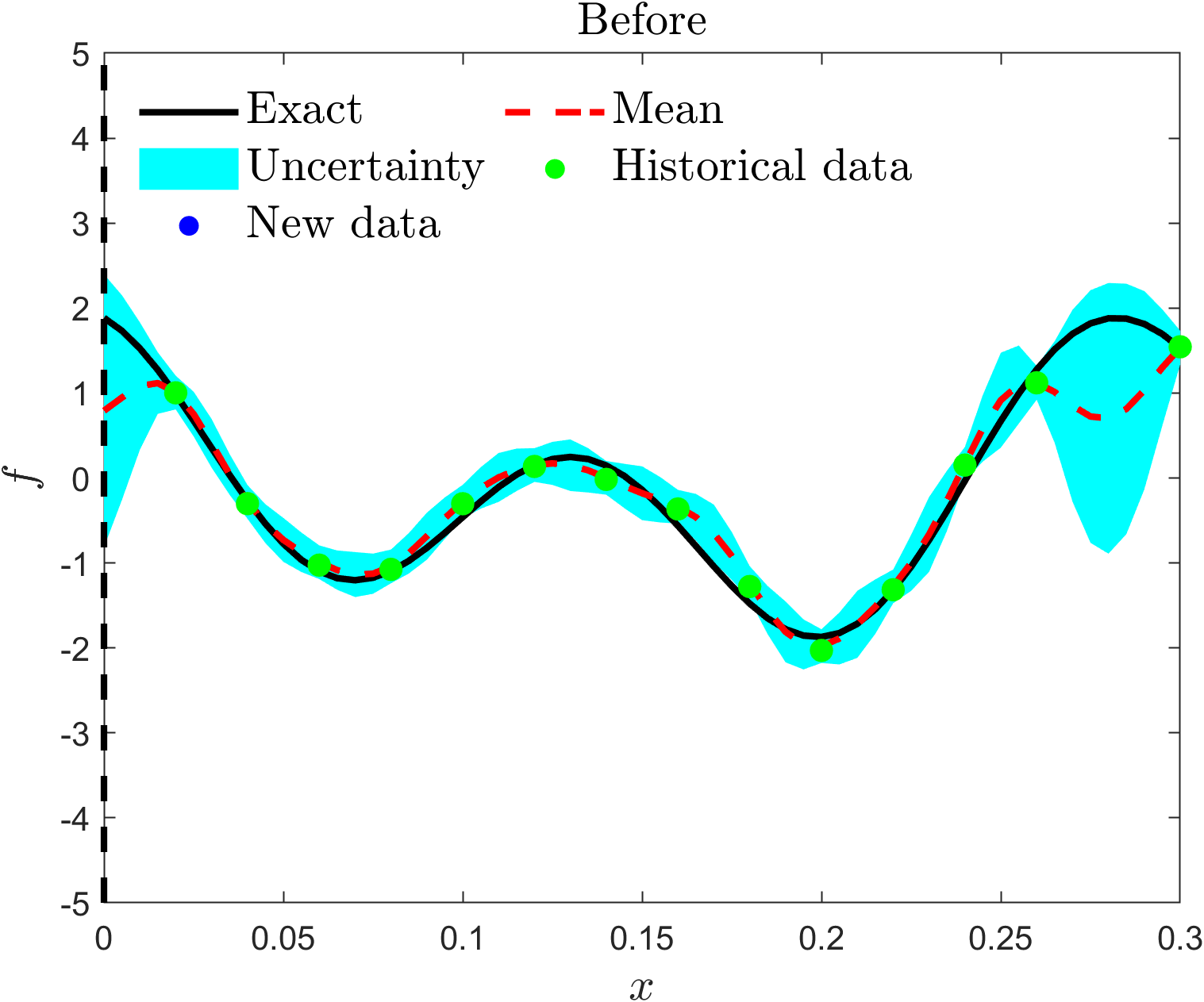}
        \includegraphics[width=.22\textwidth]{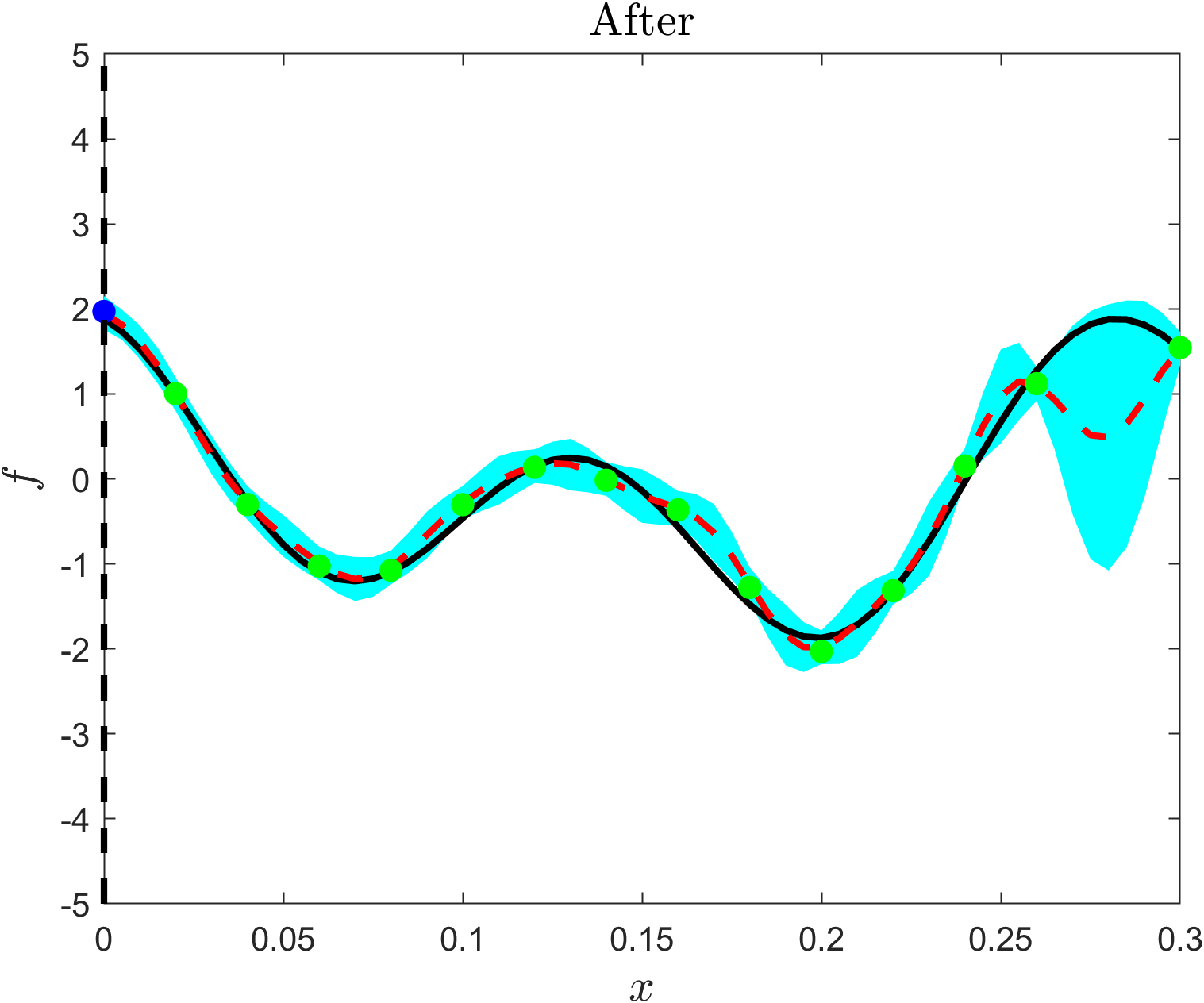}
    }
    \subfigure[Inferences of $u$.]{
        \includegraphics[width=.22\textwidth]{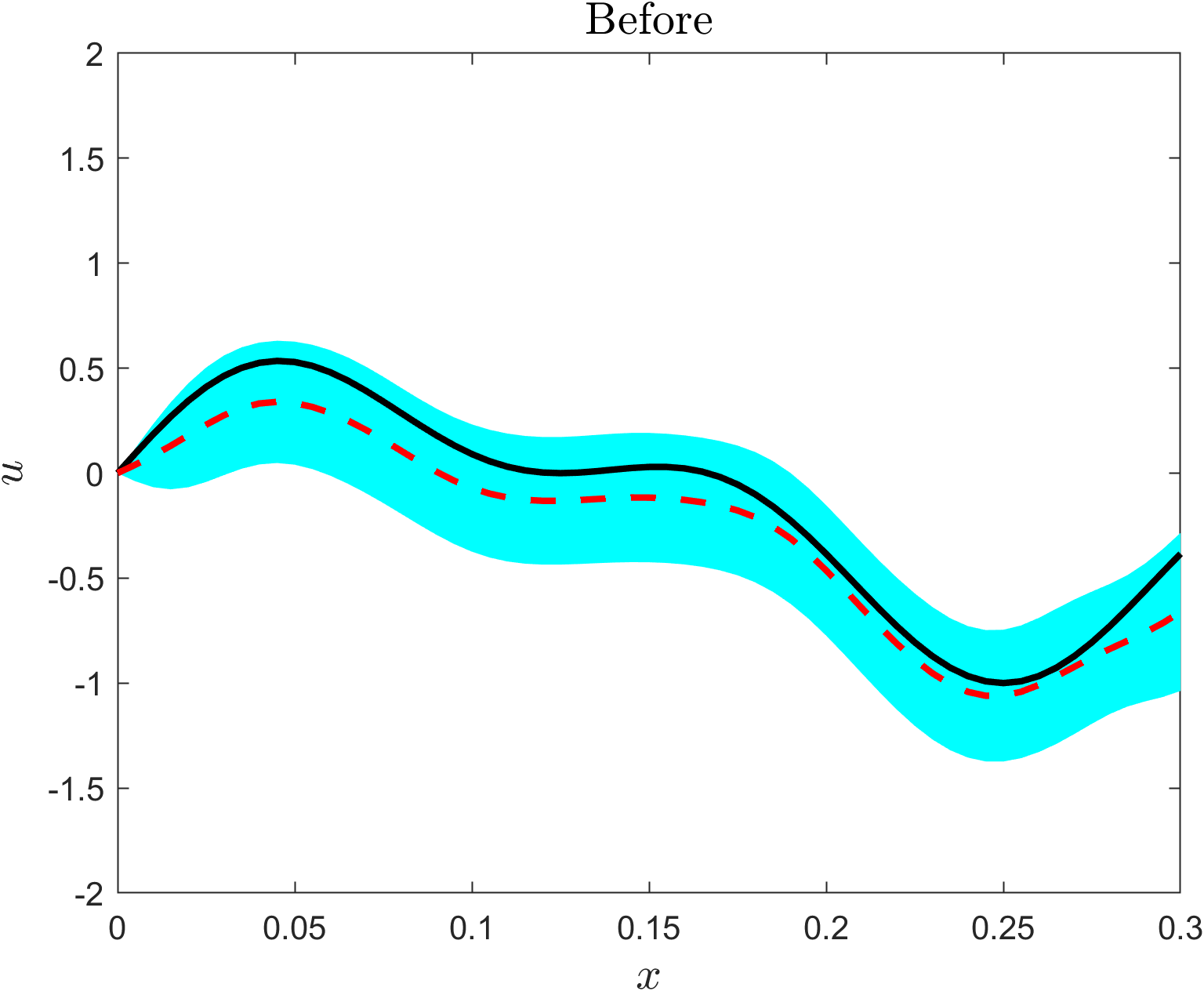}
        \includegraphics[width=.22\textwidth]{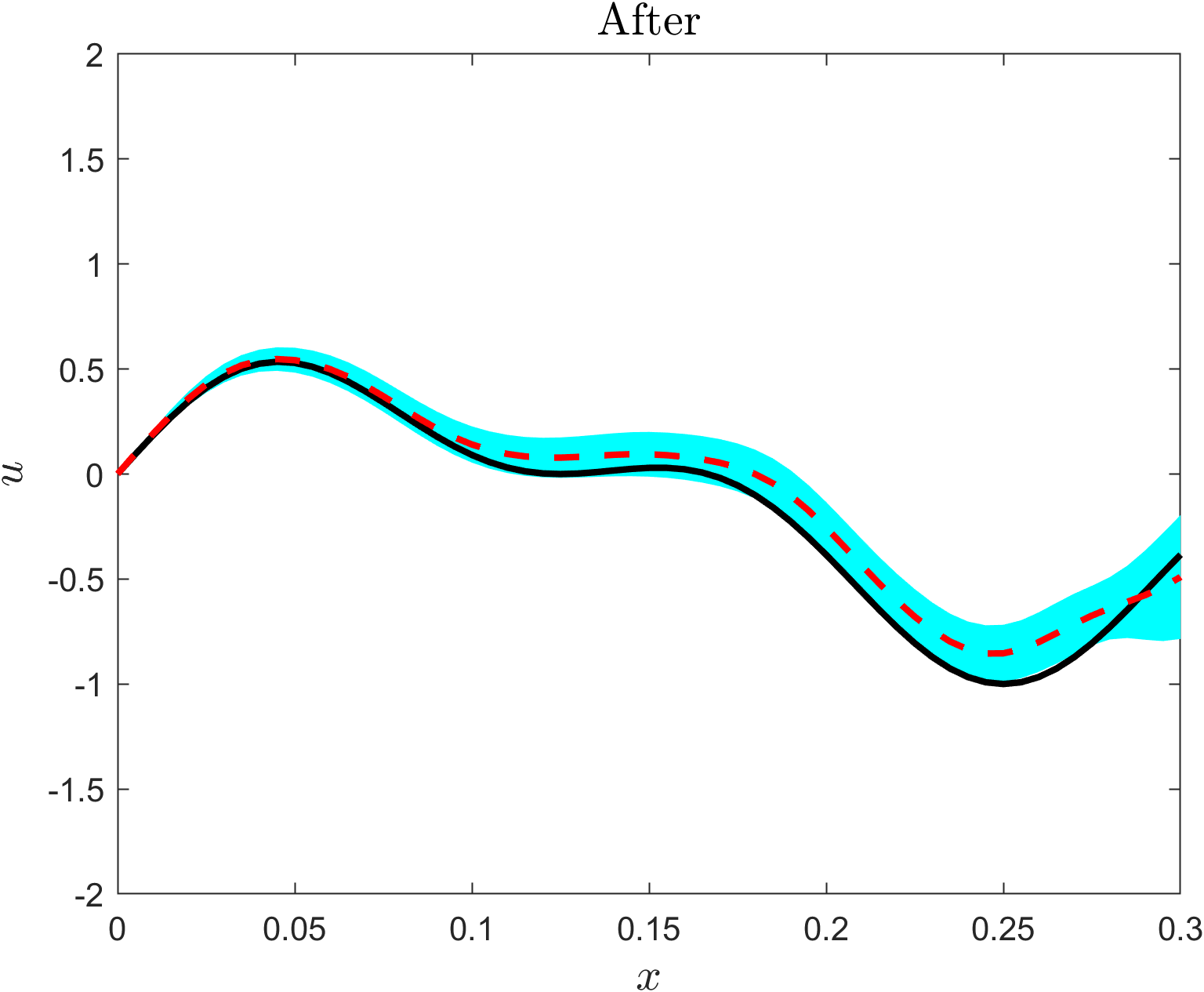}
    }
    \caption{Intermediate result for adding the $47$th data point when using active learning and our Riccati-based approach to solve \eqref{eq:example_2}. 
    The \textbf{left} side of (a) and (b) shows the inferences of $f$ and $u$, respectively, before the placement of the new sensor and incorporation of the $47$th data point. The \textbf{right} side of (a) and (b) shows the updates to $f$ and $u$ after these actions.
    The vertical black dashed line represents the location ($x=0$) where the $47$th sensor will be placed. Note that we zoom in on the region around the new sensor instead of showing the full domain for clarity of comparison. Our Riccati-based approach naturally complements this active learning framework as it can efficiently perform repeated updates of the predicted mean and uncertainty.}
    \label{fig:example_2_4}
\end{figure}

In this section, we present additional results for Case B of Example 2. Recall that in Section~\ref{sec:4_2_2}, we consider the scenario where we have access to high quality data with low-level noise but sensors are expensive to deploy. As such, the location of new data measurements must be chosen carefully.
Following an active learning framework \cite{ren2021survey}, we choose the location of the next sensor to be where the predicted uncertainty is the highest. 
In Figure~\ref{fig:example_2_4}, we show the intermediate results when adding the 47th sensor. In this case, the predicted uncertainty of $u$ more dramatically reduces. This example demonstrates how the potential of our Riccati-based approach for real-time inferences allows it to be seamlessly integrated into active learning applications.